\documentclass[9pt,twocolumn,twoside]{pnas-new}

\templatetype{pnasresearcharticle} 

\usepackage{xcolor}

\def\blankpage{%
      \clearpage%
      \thispagestyle{empty}%
      \addtocounter{page}{-1}%
      \null%
      \clearpage}

\usepackage[final]{pdfpages}

\begin{document}

\title{Hidden Citations Obscure True Impact in Science}

\author[a]{Xiangyi Meng}
\author[a,b]{Onur Varol}
\author[a,c,d,1]{Albert-L{\'a}szl{\'o} Barab{\'a}si}

\affil[a]{Network Science Institute and Department of Physics, Northeastern University, Boston, Massachusetts 02115, USA}
\affil[b]{Faculty of Engineering and Natural Sciences, Sabanci University, Istanbul 34956, Turkey}
\affil[c]{Channing Division of Network Medicine, Department of Medicine, Brigham and Women’s Hospital, Harvard Medical School, Boston, Massachusetts 02115, USA}
\affil[d]{Department of Network and Data Science, Central European University, Budapest 1051, Hungary}

\leadauthor{Meng}

\significancestatement{When a discovery or technique becomes common knowledge, its citations suffer from what Robert Merton called ``obliteration by incorporation.'' This phenomenon leads to the concept of hidden citations, representing unambiguous textual references to a discovery without an explicit citation to the corresponding manuscript(s). Previous attempts to detect hidden citations have been limited to manually identifying in-text mentions. Here, we use machine learning to systematically identify hidden citations, finding that they emerge regardless of publishing venue and discipline, {their frequency being influenced by the level of discussion within manuscript texts.} Hidden citations lead to inevitable credit distortion and capture the ``burden'' of success in science: the more widely a concept is used, the more hidden it is from standard bibliometric analysis.}

\authorcontributions{All authors designed and did the research. O.V. and  A.-L.B. conceived the concept. X.M. developed the methodology. X.M and O.V. collected and analyzed the data. X.M. and A.-L.B. were the lead writers of the manuscript.}
\authordeclaration{A.-L.B. is the scientific founder of Scipher Medicine, Inc., which applies network medicine to biomarker development, of Foodome, Inc., which applies data science to health, and of Datapolis, Inc., which focuses on human mobility.}
\correspondingauthor{\textsuperscript{1}To whom correspondence should be addressed. E-mail: a.barabasi@northeastern.edu}


\begin{abstract}
References, the mechanism scientists rely on to signal previous knowledge, lately have turned into widely used and misused measures of scientific impact. Yet, when a discovery becomes common knowledge, citations suffer from obliteration by incorporation.  This leads to the concept of hidden citation, representing a clear textual credit to a discovery without a reference to the publication embodying it. Here, we rely on unsupervised interpretable machine learning applied to the full text of each paper to systematically identify hidden citations. We find that for influential discoveries hidden citations outnumber citation counts, emerging regardless of publishing venue and discipline. We show that the prevalence of hidden citations is not driven by citation counts, but rather by the degree of the discourse on the topic within the text of the manuscripts, indicating that the more discussed is a discovery, the less visible it is to standard bibliometric analysis. Hidden citations indicate that bibliometric measures offer a limited perspective on quantifying the true impact of a discovery, raising the need to extract knowledge from the full text of the scientific corpus.
\end{abstract}

\dates{This manuscript was compiled on \today}
\doi{\url{www.pnas.org/cgi/doi/10.1073/pnas.XXXXXXXXXX}}

\maketitle
\thispagestyle{firststyle}
\ifthenelse{\boolean{shortarticle}}{\ifthenelse{\boolean{singlecolumn}}{\abscontentformatted}{\abscontent}}{}

\firstpage[13]{2}

\dropcap{``W}e stand on the shoulders of giants,'' the oft-quoted statement acknowledging the cumulative nature of knowledge, has an explicit carrier in the contemporary scientific discourse: the \emph{citation}.
Since the 1960s, references---which serve primarily as a mechanism to signal prior knowledge, enhance credibility, and protect against plagiarism---have taken on a secondary role of \emph{allocating scientific credit}, turning into an often used and misused measure of scientific impact~\cite{cit_g79,online-publ_e08,atypical_umsj13}.
Yet, when a discovery or technique becomes common knowledge to such a degree that it does not warrant citation any longer, citations suffer from 
what Robert Merton in 1968 called ``obliteration by incorporation (OBI)~\cite{soc-theor-soc-struct,obliter-inc_g75}.''
{For example, concepts like general relativity or black hole evaporation today are so embedded into scientific literacy, that only rarely do manuscripts focusing on the topics cite Einstein's 1915 work~\cite{gr_e15} or Unruh's 1976 paper~\cite{unruh-eff_u76}.}
As a consequence, foundational ideas of science are undercited, without being underused. 
This phenomenon leads to {hidden citations}, representing unambiguous allusions to a body of knowledge without an explicit citation to the manuscript(s) that introduced it.
Hidden citations,
also known as implicit, indirect~\cite{obliter-inc_t92} or informal citations~\cite{obliter-inc_mc09},
can also be induced by restrictions imposed by publishing venues on the number of references, prompting authors to cite reviews and books to signal a wider body of knowledge, rather than crediting the original discoveries.
While Merton considered such hidden citations the highest level of acknowledgement---a badge of honor, rather than a negative effect~\cite{soc-theor-soc-struct}, such credit is no longer accessible to traditional bibliometric measures.
In the era when citations are widely used as measures of impact~\cite{sci-sci-cit-dist_rfc08,long-term-impact_wsb13,sci-sci-netw_zszwfws17,sci-sci_fbbehmprsuvwwb18,sci-sci-dyn_ghbgbe18,sci-sci-prize_mu18,sci-sci-team-size_wwe19,sci-sci-fresh-team_zfdwh21,sci-sci-neural-netw_pkbra21}, hidden citations remain \emph{hidden}---not to human beings---but to the quantitative and statistical tools frequently used to quantify scientific credit. This leads to
systematic distortion of the credit landscape, diminishing the quantifiable impact of the very discoveries that define scientific progress.

Previous attempts to detect hidden citations have been limited to  manually searching and identifying in-text mentions such as ``Southern blot~\cite{obliter-inc_t92},''
``density functional theory~\cite{obliter-inc_mc09},'' ``Nash equilibrium~\cite{obliter-inc_m11},'' ``evolutionarily stable strategy~\cite{obliter-inc_m12},'' and ``bounded rationality~\cite{obliter-inc_m14,obliter-inc_m15}.'' Yet, the lack of automated methodology for determining in-text allusions [including eponyms~\cite{eponym-extract_c14,eponym_sgs22}, relating a person to a discovery] and their corresponding primordial references~\cite{prim-ref_c18} has limited our ability to understand the prevalence of hidden citations to a narrow corpus of manually inspected papers, raising the need for ``a more comprehensive estimate of uncitedness~\cite{obliter-inc_t92}.''
Here, we fill this gap by using machine learning to automatically detect \emph{catchphrases}~\cite{obliter-inc_m11}, representing in-text allusions to specific discoveries, matching them with the appropriate primordial references called \emph{foundational papers}. The method allows us to systematically identify hidden citations across the whole scientific literature, and to {trace} the factors responsible for credit distortion.
As roughly $90\%$ of obliteration by incorporation  happens in the main text of a manuscript~\cite{obliter-inc_m14}, we apply machine learning to the full text, helping us better capturing the accumulation and {distortion} of credit in science.

\begin{figure*}[t!]
	\centering
	\includegraphics[width=11.4cm]{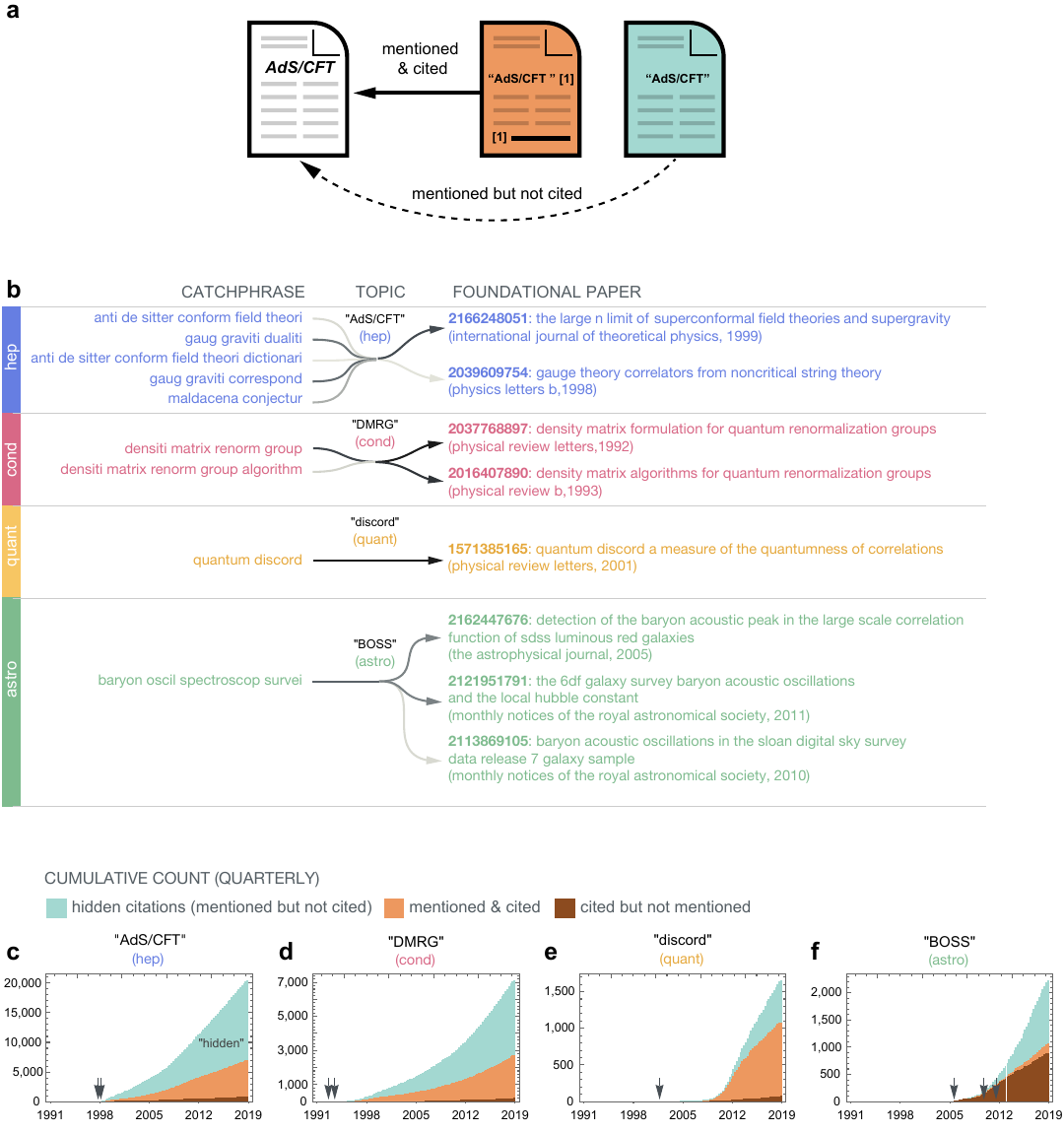}
	\caption{\label{fig_diagram}
    \textbf{Hidden citations.}
		\textbf{(a)} A foundational paper is a manuscript that introduces a new concept that subsequently defines a topic of inquiry by the scientific community, such as the topic ``anti-de Sitter/conformal field theory,'' also known as ``AdS/CFT~\cite{ads-cft_m99}.''
		Papers focusing on the topic mention the catchphrase ``AdS/CFT'' or ``anti-de Sitter/conformal field theory,'' followed by a citation to one of the foundational papers. Often, however, the catchphrases are present without explicit citations, resulting in hidden citations.
		\textbf{(b)} Exemplary topics selected from high energy physics (hep), condensed matter physics (cond), quantum physics (quant), and astrophysics (astro), together with their corresponding catchphrase(s) (lemmatized as word stems) and foundational paper(s) (Microsoft Academic Graph id). Darker arrows denote the algorithm's higher statistical confidence for the respective foundational paper.
		\textbf{(c-f)} Time evolution of citations and hidden citations for the topics listed in (b). 
		The arrows denote the publication date(s) of the foundational paper(s) for each topic.    
	   }
\end{figure*}

\section*{Results}

Each scientific discovery builds on a body of knowledge embodied by latent topics that are topically named within a manuscript and accompanied by citations to the foundational papers. 
For example, papers focusing on anti-de Sitter/conformal field theory (AdS/CFT), exploring the correspondence between general relativity and quantum field theory, cite the 1999 paper that introduced the concept [Fig.~\ref{fig_diagram}(a)]. 
Yet, many papers on AdS/CFT use language 
that for experts unambiguously defines the paper's topic, without citing the foundational work. To identify such hidden citations, we use the Latent Dirichlet Allocation (LDA) model~\cite{lda_bnj03,lda_jwyfjlz19} to detect topics in the text of a publication, inferring latent topical structures from a corpus of full-text citation contexts based on symbolic natural language processing and Bayesian inference. In contrast with neural-network-based 
Word2Vec~\cite{word2vec_msccd13} or BERT~\cite{bert_dclt18}, the LDA model is an unsupervised machine learning approach  that is interpretable,  allowing us to associate the outcomes of LDA with confidence levels through transparent probabilistic logic (see Methods).
	
We identified $343$ topics in physics that accumulate hidden citations, each with at least one catchphrase {and} at least one {foundational paper} (see Methods).
Shown as examples are four topics uncovered by the algorithm [Fig.~\ref{fig_diagram}(b)], as well as the {followers} of each topic [Fig.~\ref{fig_diagram}(c-f)], defined as papers that either cited the foundational papers of the given topic or mentioned the  corresponding topic-specific catchphrases, or both (see Methods). For example,
the orange regions denote the temporal evolution of the number of papers that simultaneously cite the foundational papers and carry the respective catchphrases. 
The top green region captures hidden citations, counting papers that make an unambiguous textual reference to the topic but fail to cite any of the foundational papers. For example, less than half of the papers that use the catchphrases of  ``AdS/CFT'' cited any of the two foundational papers: 1999 paper by Maldacena and 1998 paper by Gubser \textit{et al.} 
[Fig.~\ref{fig_diagram}(c)]. 
Taken together, we find that for the four topics featured in Fig.~\ref{fig_diagram}(b), hidden citations correspond to
$65.8\%$,
$61.7\%$, 
$34.6\%$, 
and $52.3\%$ 
of all detectable credit since the publication of the respective topic's first foundational paper, overcoming the bibliometrically quantifiable and tabulated citations.

\begin{figure*}[t!]
	\centering
	\includegraphics[width=11.4cm]{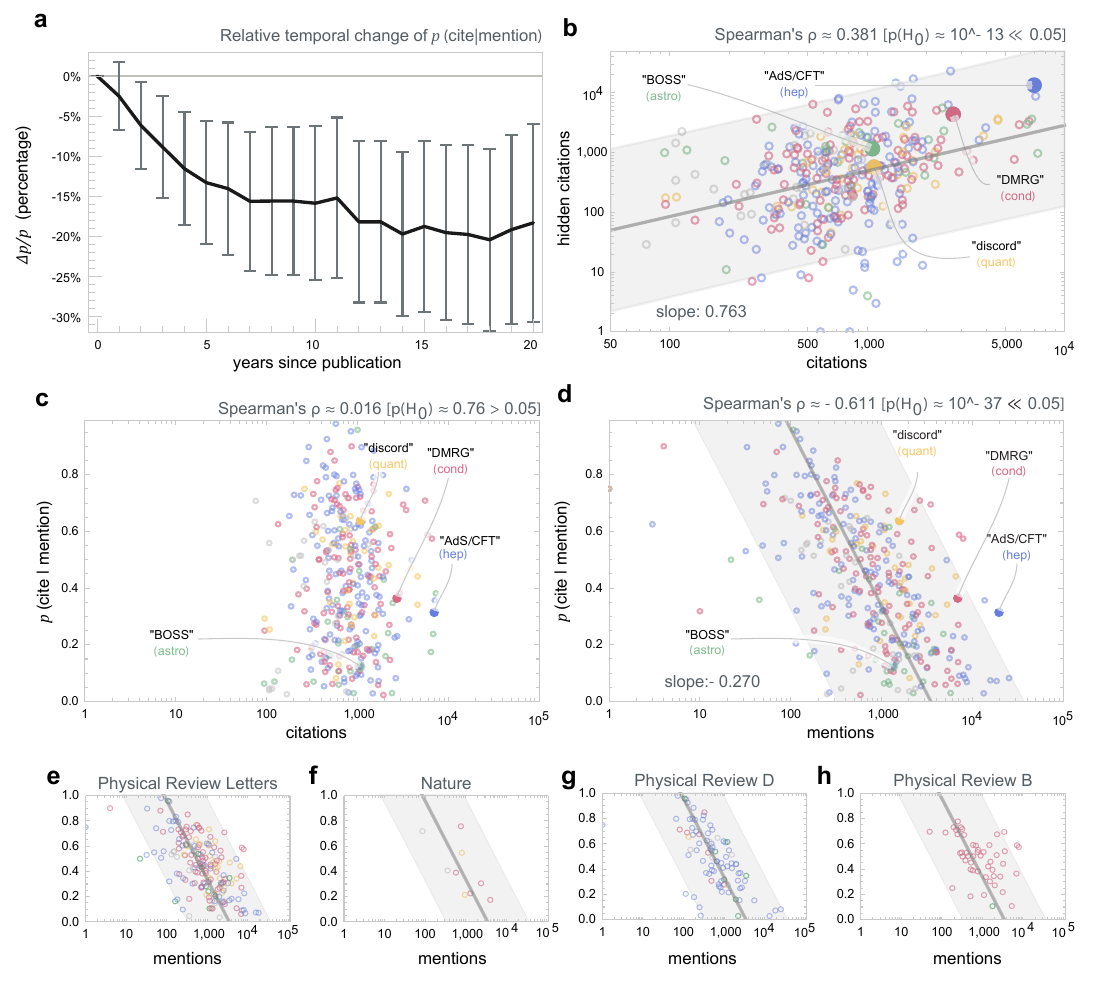}
	\caption{\label{fig_temporal} 
	    \textbf{Factors that drive hidden citations.}
		\textbf{(a)} The temporal change of $p(\text{cite}|\text{mention})$, the probability that a paper mentioning the topic-specific catchphrases will also cite the foundational paper, as a function of time (years since publication). On average, $p(\text{cite}|\text{mention})$ per topic drops by approximately $20\%$ after $20$ years of publication of the first foundational paper.
        Error bars represent $95\%$ confidence intervals.
		\textbf{(b)} Topics with more citations ($c$) tend to have more hidden citations ($h$) (with Spearman's rank correlation $\rho\approx 0.381$ and null hypothesis $H_0$ rejected). 
		Most topics fall into the $95\%$ single-observation confidence bands with a log-log slope $0.763{\pm0.208}$, indicating that $h\sim c^{0.763}$.
		\textbf{(c)} $p(\text{cite}|\text{mention})$ as a function of citations per topic ($\rho\approx 0.016$, $H_0$ not rejected), indicating that the probability of a textual reference becoming a hidden citation is not driven by the number of citations to the topic.
		\textbf{(d)} $p(\text{cite}|\text{mention})$ as a function of mentions per topic ($\rho\approx -0.611$, $H_0$ rejected). The strong negative correlation indicates that hidden citations are driven by the number of textual mentions of the topic.
		Most topics fall into the $95\%$ confidence bands with a log-linear slope $-0.27{\pm0.04}$. The pattern holds for four distinct publication venues \textbf{(e-h)}.
		}
\end{figure*}
	
The high proportion of hidden citations prompts us to calculate the temporal changes in the conditional probability that a paper that mentions the topic-specific catchphrases cites the foundational papers, $p(\text{cite}|\text{mention})$ (SI, Section~9).
We find that the probability that the foundational papers are cited drops by approximately $20\%$ after $20$ years [Fig.~\ref{fig_temporal}(a)], indicating that the reliance on hidden citations, hence OBI, strengthens over time.

\begin{figure*}[t!]
	\centering
		\includegraphics[width=11.4cm]{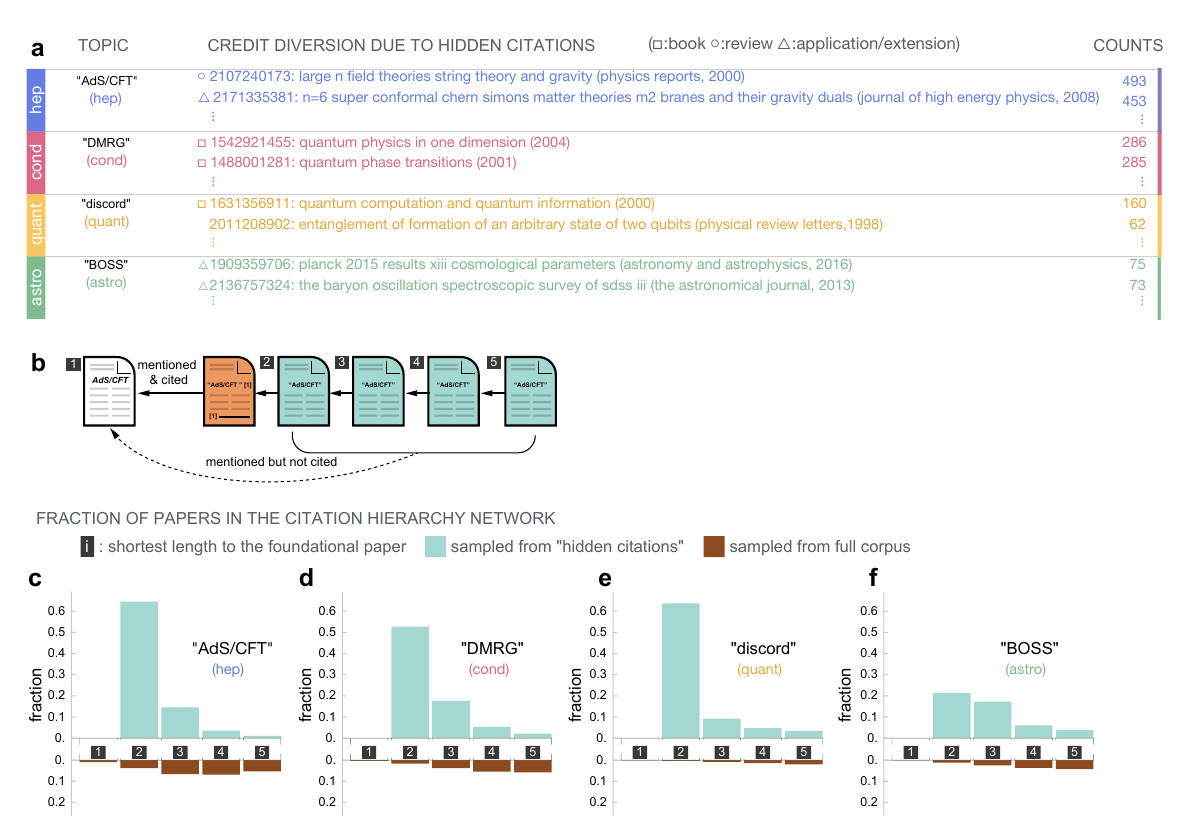}
		\caption{\label{fig_topic} \textbf{Credit redirected.} \textbf{(a)} The most cited alternatives for four topics that acquire hidden citations, primarily indicating that credit is often diverted to books, reviews or applications/extensions of the foundational papers. \textbf{(b)} Most alternatives to hidden citations are related to the foundational paper, detectable by tracking the citation path between the alternative and the foundational paper. \textbf{(c-f)} Fraction of hidden citations ranked by their citation hierarchy to the foundational papers. For each topic (except ``BOSS''), around $60\%$ of hidden citations (green, top) cited other arXiv papers that explicitly cited the foundational papers. For a randomly sampled reference from the full arXiv, this fraction is negligible (brown, bottom).}
\end{figure*}
	
Do hidden citations correspond to pure untracked credit [a.k.a.~implicit citations~\cite{obliter-inc_t92}], or is credit diverted to other works [a.k.a.~indirect citations~\cite{obliter-inc_t92}]? To distinguish these two mechanisms, we identified the most frequently cocited publications accompanying a hidden citation.
We find that for ``AdS/CFT'' the most cited alternative is a review coauthored by the authors of the two foundational papers, and for ``DMRG''
the most cited alternatives are two books [Fig.~\ref{fig_topic}(a)]. Credit is also diverted to applications of the topic, such as the application of AdS/CFT to topological quantum field theory, or to extensions on the topic, like in the expanded ``BOSS'' datasets [Fig.~\ref{fig_topic}(a)]. 
Overall, we find that the works that collect the credit from hidden citations tend to cite the foundational papers, or cite papers that in turn cite the foundational papers [Fig.~\ref{fig_topic}(b)].
Indeed, around $60\%$ of hidden citations have a citation path length of $2$ to the foundational papers [Fig.~\ref{fig_topic}(c-f)].
indicating that 
hidden citations do cite and give credit to papers whose topics closely relate to the {foundational} papers.
To determine whether the previously observed increase in reliance on hidden citations over time [Fig.~\ref{fig_temporal}(a)] is dominated by implicit citations or indirect citations, we recalculated the temporal changes [Fig.~\ref{fig_temporal}(a)], this time also including indirect citations, i.e.,~the hidden citations that have a citation path length of at most $2$. We find 
now that $p\left[(\text{cite}+\text{indirectly cite})|\text{mention}\right]$ increases with time (SI, Section~10), indicating that the increasing reliance on hidden citations is accompanied by an increasing tendency to divert credit to other works.

As Fig.~\ref{fig_temporal}(b) shows, topics with more citations ($c$) tend to accumulate more hidden citations ($h$), {a trend} approximated by a sublinear dependence $h\sim c^{0.763}$, indicating that on average a topic with $5,000$ citations accumulates approximately $1,000$ hidden citations.
While this scaling suggests that citations are the main driving force of hidden citations, our measurements indicate otherwise. 
Indeed, we find a negligible correlation ($\rho\approx0.016$) between $p(\text{cite}|\text{mention})$ (the probability of being cited if mentioned) and the number of citations for the respective topic [Fig.~\ref{fig_temporal}(c)] (SI, Section~8). 
We find, however, a strong negative correlation ($\rho\approx-0.611$) between $p(\text{cite}|\text{mention})$ and the number of mentions per topic [Fig.~\ref{fig_temporal}(d)].
In other words, the more discussed is a discovery in the textual context of a paper, the less likely scientists feel the need to explicitly cite it, a {``burden'' of success} that is independent of the publication venue [Fig.~\ref{fig_temporal}(e-h)].

\begin{figure*}[t!]
	\centering
	\includegraphics[width=11.4cm]{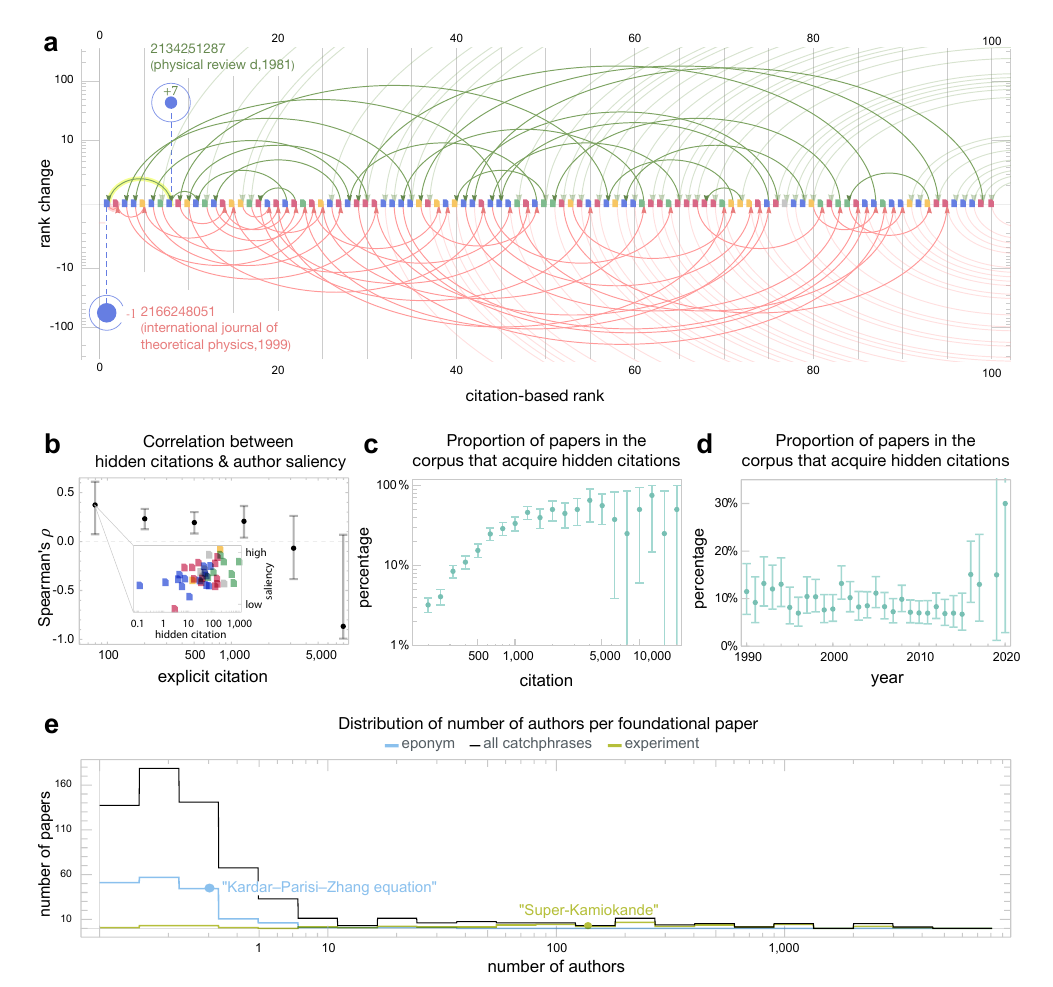}
		\caption{\label{fig_paper}
			\textbf{Foundational papers.}
			\textbf{(a)} Changes in the citation-based ranks of the top-ranked foundational papers after taking hidden citations into account, shown by arrows from the old explicit-citation-based rank to the new explicit-plus-hidden-citation-based rank (green: rank rise; red: rank drop). After accounting for hidden citations, the ``cosmological inflation theory'' paper (2134251287), 
            ranked \#8 based on explicit citation counts, takes the top spot.
            \textbf{(b)} For foundational papers with similar numbers of explicit citations, the paper with more hidden citations tends to result in higher average author saliency (inset).
			The proportion of papers in the corpus that can acquire hidden citations increases with \textbf{(c)} the explicit citations but not with \textbf{(d)} the publication year of the papers. Error bars represent $95\%$ confidence intervals.
			\textbf{(e)} Distribution of foundational papers by the number of authors per foundational paper, shown for all catchphrases (black) and for eponym-related (blue) and experiment-related catchphrases (green).
		}
\end{figure*}

To explore the impact of hidden citations on bibliometric measures, 
{it is tempting to calculate the hidden citations of \emph{individual} foundational papers (SI, Section~11). We must approach this with caution: since our methodology operates at the topic level, transferring citation counts from topics down to individual papers is inherently imprecise and cannot be guaranteed to be accurate. Consequently, the paper-specific observations we offer here should be viewed as provisional insights rather than definitive conclusions.}

{Our first observation is that the ratio of hidden to explicit citations is, on average, $0.98$:$1$, indicating that papers tend to acquire hidden citations at the same rate as they acquire explicit citations.
Yet, we do observe considerable variability in this ratio. Hence, for some foundational papers hidden citations can dominate over explicit citations.}
Examples include
the paper introducing the cosmological inflation theory in 1981 
that acquired $8.8$ times more hidden citations than explicit citations, or
the 1974 work that merged the electromagnetic, weak, and strong forces into a single force, which accumulated $6.6$ times more hidden citations than explicit citations.
This prompted us to calculate the changes in citation-based ranks between foundational papers {(SI, Section~11).}
As Fig.~\ref{fig_paper}(a) indicates, most papers in the top $100$ list suffer rank loss (green lines), thanks to a few publications that accumulate an exceptional number of hidden citations, and gain significantly in rank (red lines).
For example, the most cited paper of arXiv, the 1999 paper which started the formal theory of AdS/CFT, 
loses its top ranking once we take hidden citations into account, to the 1981 paper previously ranked \#8, which started the phenomenological study of the cosmological inflation theory 
[Fig.~\ref{fig_paper}(a)].
Hidden citations could potentially have an impact on authors as well. To see if this is the case, we adopted the Microsoft Academic Graph's {``author saliency''} metric, that relies on the heterougeneous network structure of the connectivity of articles, authors, and journals, designed to be less susceptible to raw citation counts and temporal bias~\cite{mag_wshwdk20}.
We find that, when we compare two foundational papers with similar numbers of explicit citations, authors with more hidden citations have {higher} average saliency, a positive correlation notable for papers with less than $3,000$ citations [Fig.~\ref{fig_paper}(b)].
This suggests that OBI {tends to correlate} with a \emph{positive} impact on authors whose papers' full citation impact has not yet developed.
Interestingly, the effect disappears for well-recognized papers, for which missing citations {does not appear to affect} their authors' reputation.

Papers that became foundational papers and acquired hidden citations tend to be highly cited, accumulating on average $434{\pm34}$ explicit citations, in contrast with $1.4$ explicit citations for all physics papers in the corpus.
Yet, not only highly cited papers acquire hidden citations.
We find that even among papers with citations $\lesssim500$, a nonnegligible fraction ($>10\%$) of papers {may} acquire hidden citations [Fig.~\ref{fig_paper}(c)].
Since our approach to identifying hidden citations (see Methods) is conservative, designed to reduce false positive errors, the actual fraction of papers that acquire hidden citations is likely higher.
We also find that hidden citations are not limited to older papers, but they accompany recent publications as well [Fig.~\ref{fig_paper}(d)], such as the discovery of gravitational wave (2016) or exclusion of dark matter particles in the Large Underground Xenon experiment (2017).

Finally, we investigated the sociodemographic characteristics (gender, country of origin, and the prestige of institution) of the authors of foundational papers  (SI, Section~12). {We find that hidden citations capture a universal phenomenon that can emerge in any institution, regardless of their level of prestige, 
and we observe no statistically significant bias based on gender or country of origin in hidden citations apart from the overall biases in explicit citations that have long been observed~\cite{divers_k04,divers_l06,divers_ad18,sci-sci-gender_hgsb20}.
Hence, although the methodology presented in the paper can account for additional hidden citations, it simply reflects the existing biases.}

\begin{figure*}[t!]
	\centering
	\includegraphics[width=11.4cm]{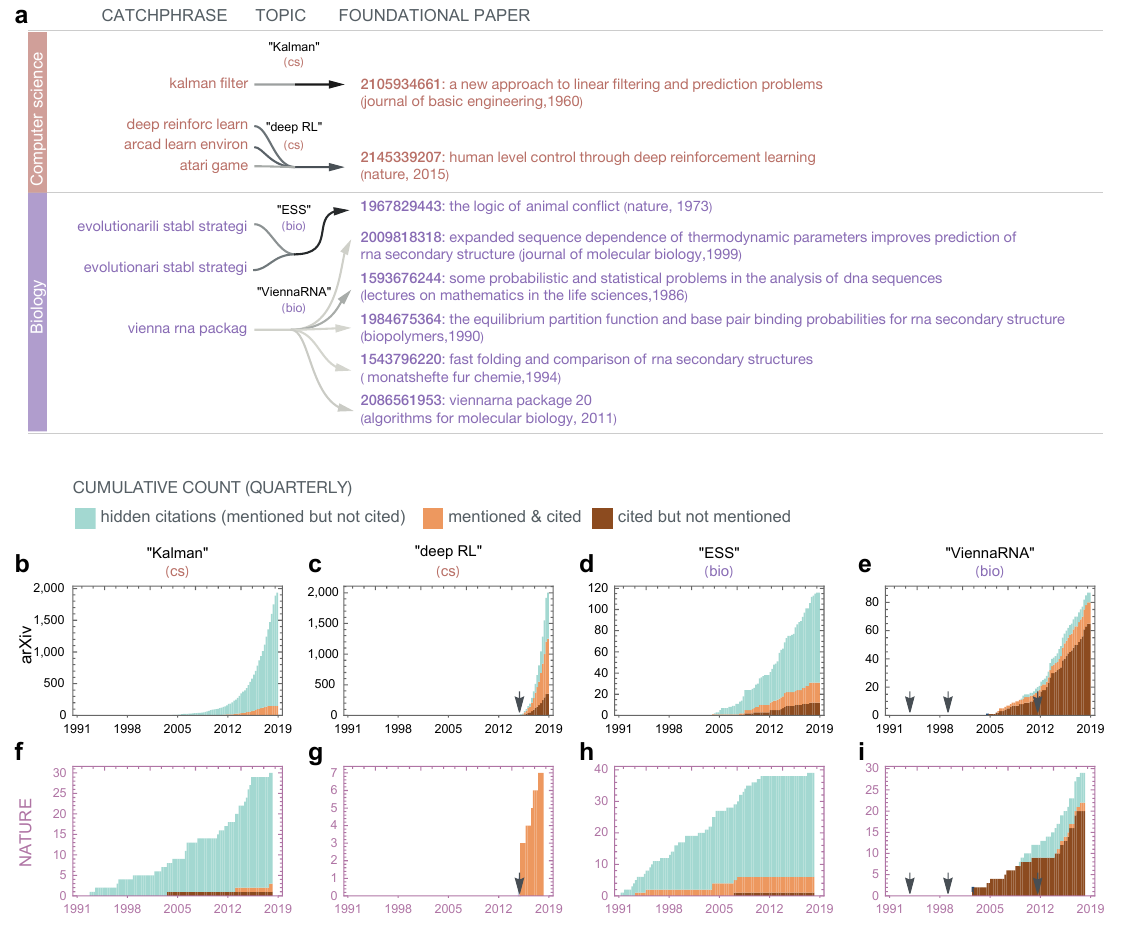}
	\caption{\label{fig_extra} 
		\textbf{Hidden citations across disciplines and venues.}
		\textbf{(a)} Four topics selected from computer science (cs) and biology (bio) [cf.~Fig.~\ref{fig_diagram}(b)].
		\textbf{(b-i)} Time evolution of citations and hidden citations [cf.~Fig.~\ref{fig_diagram}(c-f)] for the four topics shown in \textbf{(a)}, identified from arXiv \textbf{(b-e)} and \emph{Nature} \textbf{(f-i)}.
		}
\end{figure*}

{Depending on the specific assumptions made in our methodology to redistribute credit from topics to individual papers, the observations [Figs.~\ref{fig_paper}(a-d)] will likely look different. This serves as a reminder of the significant role hidden citations---and how they are calculated---play in the allocation of scientific credit.}

{These observations highlight the potential risks associated with ignoring or misidentifying hidden citations in credit allocation. Such risks} raise an important question: what determines the emergence of hidden citations?
Our analysis suggests that there is a prerequisite for a paper to acquire hidden citations: it must develop exclusive catchphrases that are synonymous to the paper itself,
becoming a ``{conceptual symbol}''~\cite{concept-symb_s78,norm-cit-theor_s04} within the field.
For example, whenever ``quantum discord'' is mentioned, an expert in the field will immediately link it to the 2001 foundational paper, 
and vice versa, seeing a citation to that particular reference, an expert thinks of ``quantum discord.''
To quantify this dual correspondence, we first measured the degree of non-exclusivity of linking a given $n$-gram $w$ to a paper $d$ by calculating the specific conditional entropy $\mathcal{S}(d|w)$ (SI, Section~13), finding that $\mathcal{S}(d|w)$ is considerably lower for catchphrases than for non-catchphrases. For example, $\mathcal{S}(d|\text{``quantum discord''})\approx4.07$ in contrast with $\mathcal{S}(d|\text{``quantum mechanics''})\approx6.73$, indicating that ``quantum discord'' is a catchphrase pointing to a well defined foundational paper, while ``quantum mechanics'' is too general to be exclusively assigned to one or a few foundational papers.
Inversely, we measured the specific conditional entropy $\mathcal{S}(w|d)$ of seeing paper $d$ and linking it to an $n$-gram $w$, finding again that $\mathcal{S}(w|\text{1571385165})\approx5.97$ for the 2001 foundational paper 
with catchphrase ``quantum discord'' is lower than the highly cited 1999 paper also focusing on quantum information processing, 
$\mathcal{S}(w|\text{2097039598})\approx7.72$, but not categorized by our algorithm as a foundational paper. These results confirm that to develop hidden citations, a (catchphrase)--(foundational paper) pair must acquire \emph{mutual} exclusivity: a paper does not accumulate hidden citations if its conceptual significance does not lead to an unambiguous catchphrase, or if that catchphrase is not exclusive enough for the community to unambiguously link it back to the original paper. 

Asking where such exclusive catchphrases originate,
we find that for $78.8\%$ of the $880$ foundational papers 
the corresponding catchphrases do not appear in the titles or the abstracts of them (SI, Section~14).
This indicates that catchphrases are typically not proposed by the authors of the foundational papers, but are assigned later by the community~\cite{norm-cit-theor_s04}.
We also find that $26.0\%$ of all foundational papers have catchphrases that correspond to eponyms (e.g.,~``Kardar–Parisi–Zhang equation,'' that governs surface growth) 
and another $7.1\%$ acquire the names of experimental projects (e.g.,~``Super-Kamiokande,'' the discovery of neutrino oscillation) 
(SI, Section~14).
Eponym-related catchphrases emerge mainly for papers with short author lists---indeed, foundational papers with eponyms as catchphrases have $2.50{\pm0.36}$ authors on average, in contrast with $72.4{\pm26.5}$ authors for non-eponym-related catchphrases, and  $405{\pm164}$ for papers with experiment-related catchphrases [Fig.~\ref{fig_paper}(e)].

Identifying hidden citations in all areas of science requires a large and unbiased corpus of full-text citation contexts. While such a corpus is so far unavailable for all science, we have access to $818,311$ computer science and $140,865$ biology full-text manuscripts, allowing us to identify catchphrases and foundational papers in these fields as well [Fig.~\ref{fig_extra}(a)]. 
The patterns governing hidden citations are largely indistinguishable from those documented for physics: we observe a significant number of hidden citations for established research topics like ``Kalman'' (refining estimates from new measurements) 
and ``ESS'' (evolutionary strategies in natural selection)  
and even for newer topics like ``deep RL'' (deep neural networks and artificial-intelligence) 
and ``ViennaRNA'' (analysis of RNA structures)  
[Fig.~\ref{fig_extra}(b-e)]. 
We also analyzed a corpus of $88,637$ full-text \emph{Nature} articles~\cite{nature-reach_gkvb19}, which cover multiple disciplines, finding evidence of hidden citations in highly selective peer-reviewed venues as well [Fig.~\ref{fig_extra}(f-i)]. 
These results indicate that hidden citations are a universal phenomenon, emerging in all areas of science and publishing venues, disciplines, and research topics.

\section*{Discussion}
	
Acknowledging discoveries on which new research builds on is an integral part of the scientific discourse. 
Yet, with the exponential growth of science and limits on the number of allowed references, a paper’s ability to credit all sources of inspiration is limited. 
Such limitations lead to inevitable credit distortion, manifest in situations where the textual context indicates that credit is due, but it is not accompanied by explicit citations to the pertinent work. 
Hidden citations capture the ``burden'' of success in science: the more widely a concept is used by the scientific community, the more likely that it will accrue hidden citations. 

Systematically tabulating hidden citations, together with explicit citations can help us more accurately identify emerging topics and evaluate their true impact~\cite{obliter-inc_t92,obliter-inc_mc09}.  
That being said, both explicit and hidden citations represent unequal ``atoms of peer recognition~\cite{matthew-eff_m88},'' offering different degrees of credit per citation.
Indeed, when citations point to highly cited papers (driven by the authors' fame, race, gender, etc.) without contributing to the paper's topic, they offer less credit~\cite{matthew-eff_m68}. Negative citations~\cite{negat-cit_clo15} should also offer less credit than positive citations; yet, we find that the prevalence of negative or positive texts is low in both the explicit and hidden citations of foundational papers (SI, Section~6). This can be attributed to the fact that in order for a paper to become a foundational paper and acquire hidden citations, the discovery or technique it presents should have already been accepted as common knowledge, leaving limited room for debates.
There is also a difference in crediting conceptual versus methodological advances: while papers that cite the foundational papers but fail to give appropriate textual references are extremely rare [Fig.~\ref{fig_diagram}(c-e), Fig.~\ref{fig_extra}(b-d, f-h)], for papers introducing databases [Fig.~\ref{fig_diagram}(f)] or tools [Fig.~\ref{fig_extra}(e,i)] textual references are less frequent.
One explanation is that these foundational papers are not just cited for their dataset or methodological efforts; they are also frequently cited for supporting the corresponding general concept, namely, ``baryon oscillation'' or ``RNA structure.'' In the latter case, authors often fail to mention the words ``survey'' or ``package'' when citing these papers. This textual bias suggests that database or methodological advances are often less acknowledged, 
in line with earlier findings~\cite{data-cit_bcddhpsw20}.
This is because when playing the supportive role of a general concept, the papers lost their merits as foundational papers.
The community working on the general concept often benefits from the database or methodological efforts without textually referencing and explicitly acknowledging the effort that went into creating it (SI, Section~7).

It is, therefore, important to go beyond {simple counts of} explicit and hidden citations, and develop new metrics that can also differentiate the degree of credit carried by each citation, a process to which a complete corpus of both explicit and hidden citations is a prerequisite.

{While our unsupervised methodology allows us to tabulate hidden citations at scale, the current methodology is designed to be conservative and to minimize false positive errors (see Methods). Thus, currently it may overlook hidden citations, limiting the completeness of topics. Note that the missing hidden citations can be recovered by lowering the identification thresholds, at the expense of increasing false positive errors. Another limitation is that some topics may not be present in the arXiv corpus, either because they have not been studied or discussed in a sufficient number of arXiv papers or because they are too narrow or outdated.} They could be recovered if we apply our algorithm to a more extensive full-text database that spans multiple disciplines and time periods. 
However, there is a major barrier to achieving this: the lack of systematic access to full-text papers.
Indeed, while citation counts and other metadata are now freely and easily available for research purposes, access to the full text of all research papers is restricted by commercial interests, limiting the deployment of tools capable of accurately tabulating hidden citations
and their role in the scientific discourse.

\matmethods{

{
Traditionally, the LDA model is used to uncover latent topics within a collection of documents. Each document is assumed to be a mixture of multiple topics, and each topic is characterized by a distribution over phrases.
Here, instead of exploring latent topical structures, we focus on the explicit textual observables, aiming to reveal the correspondence between phrases and documents.
}

{The input of the LDA model is a list of $2$-tuples $\{w,d\}$ between an \emph{$n$-gram} $w$  (a phrase of $n$ words, where the value of $n$ can freely vary to accommodate long phrases) and an accompanied 
text-based \emph{document} $d$ [denoted by a unique code, e.g.,~the Microsoft Academic Graph (MAG) id~\cite{mag_sssmehw15}]. The document does not contain the full text of the MAG paper $d$. Instead, it comprises the citation contexts of $d$, which represent the textual discussions by the community when citing $d$. Each $2$-tuple accounts for an exact occurrence of an $n$-gram $w$ in document $d$. 
For example, 
\begin{equation*}
    \text{input}=\begin{pmatrix}
        \text{\{``string theory'', 2166248051\}}\\ \text{\{``gauge-gravity duality'', 2166248051\}}\\ \text{\{``quantum discord'', 1571385165\}}\\
        \cdots
    \end{pmatrix}.
\end{equation*}
The output of the LDA model is a list (of the same length of the input) of $3$-tuples $\{w,z,d\}$, where each input $2$-tuple $\{w,d\}$ acquires a new latent variable $z$ that corresponds to a specific topic, such as
\begin{equation*}
    \text{output}=\begin{pmatrix}
        \text{\{``string theory'', {topic} $1$, 2166248051\}}\\ \text{\{``gauge-gravity duality'', {topic} $1$, 2166248051\}}\\ \text{\{``quantum discord'', {topic} $2$, 1571385165\}}\\
        \cdots
    \end{pmatrix}.
\end{equation*}
The output indicates that ``string theory'' and ``gauge-gravity duality'' belong to the same topic $1$, while ``quantum discord'' belongs to a different topic $2$.
The joint probability $P(w,z,d)$ of the concurrent occurrence of the $3$-tuple $\{w,z,d\}$, which can be estimated from the output, enables us to define and calculate two key terms: 
}
	 
(1) An $n$-gram $w$ with $P(z|w)>P_\text{th}^\text{catch}$ is a \emph{catchphrase} of topic $z$, implying that whenever the $n$-gram $w$ is seen in a document, we are confident that topic $z$ is referred to.
For example, if there are $1,919$ occurrences of $\{w=\text{``quantum discord''},z,d\}$ in the output, among which $1,916$ $3$-tuples also have $z=3$, then $P(z|w)\approx0.998\pm0.002$, representing the conditional probability of referring to topic $z=3$ given occurrence of $w=$``quantum discord''. If $P(z|w)$ is larger than $P_\text{th}^\text{catch}$, then we have statistical confidence that ``quantum discord'' is a catchphrase of the topic $z=3$.
	 
(2) A {document} $d$ with $P(d|z)>P_\text{th}^\text{found}$ is a \emph{foundational paper} of topic $z$, implying that whenever topic $z$ is referred to, we expect a citation to the MAG paper $d$,
indicating that the foundational paper is sufficiently disruptive~\cite{sci-sci-team-size_wwe19} to serve as a representative of the topic.
For example, if there are $9,742$ occurrences of $\{w,z=3,d\}$ in total in the output, among which $2,091$ $3$-tuples also {include the document} $d=\text{1571385165}$, then $P(d|z)\approx0.215\pm0.008$ which, if larger than $P_\text{th}^\text{found}$, makes $d=\text{1571385165}$ a foundational paper of topic $z=3$.

We rely on a strict criterion to choose catchphrases ($P_\text{th}^\text{catch}=0.95$) but a loose criterion at including foundational papers ($P_\text{th}^\text{found}=0.05$), reducing the false positive rate of incorrectly assigning a too general $n$-gram as a catchphrase, or concentrating too much hidden citations for only one or two papers, hence remaining conservative at identifying hidden citations per foundational paper.
{This unavoidably results in the exclusion of some topics for which the catchphrases and foundational papers are less exclusively defined. Therefore, our results are not based on a complete collection but a sampled aggregation of topics.}

After a latent topic $z$ is inferred by LDA, we identify all papers that follow and explore the topic $z$ [Fig.~\ref{fig_diagram}(a)], including all the papers that explicitly cite the foundational paper(s) of topic $z$, as well as papers that only mention the topic-specific catchphrase(s) but lack citations to the foundational paper(s) (SI, Section~4). {The latter corresponds to hidden citations, as they explicitly build on the {catchphrase(s)} associated with topic $z$.}
For example, a hidden citation is detected when a paper mentions the catchphrase ``quantum discord'' but lacks a citation to the foundational paper $d=\text{1571385165}$. 
	
We trained the LDA classifier using the \emph{unarXive} dataset~\cite{unarxive_sf20} that offers full-text coverage for $1,043,126$ publications, annotated with
{citation contexts}, obtained after merging the entire arXiv~\cite{arxiv_g09} with MAG~\cite{mag_sssmehw15} (SI, Section~1).
Established in 1991 as the first preprint archive, arXiv offers a fairly unbiased coverage of physical sciences. 
We identified from the citation contexts (from arXiv) all $n$-grams $w$ and each paper $d$ (in MAG) they refer to  (SI, Section~2), initially filtering out books and reviews. 
Following the arXiv taxonomy, the results are categorized into five categories (SI, Section~5): high energy physics (``hep''), condensed matter physics (``cond''), quantum physics (``quant''), astrophysics (``astro''), and the rest (``other'').
For example, the LDA model predicts that each time the catchphrase ``anti-de Sitter conform field theory'' is mentioned, it should be accompanied by either a reference to the 1999 paper (2166248051) by Maldacena~\cite{ads-cft_m99}, 
or to (2039609754) by 
Gubser, Klebanov, and Polyakov, 
both within the ``hep'' topic ``AdS/CFT'' [Fig.~\ref{fig_diagram}(b)]. 
Similarly, for the ``density matrix renormalization group'' catchphrase, the LDA model expects references to two papers (2037768897 and 2016407890) 
by White, 
within the ``cond'' topic ``DMRG'' 
that focuses on many-body ground-state wave functions [Fig.~\ref{fig_diagram}(b)].
{To validate the accuracy of LDA, we have consulted specialists in ``hep'' and ``quant'' to manually check ten randomly selected topics each (SI, Section~5). We find that $9$ out of $10$ of the specialists' choices of the foundational papers are identified by the algorithm, resulting in a $90\%$ effectiveness of our automated approach.}
}

\showmatmethods{} 

\acknow{We would like to thank Alice~Grishchenko for help with data visualization. X.M. is indebted to Yanchen~Liu, Rodrigo~Dorantes-Gilardi, Bingsheng~Chen, Alexander~J.~Gates, Louis~M.~Shekhtman, and Jing~Ma for fruitful discussions. This manuscript was posted on arXiv:2310.16181.

}

\showacknow{} 

\section*{Funding}
This research was funded by the National Science Foundation (SES-2219575), the Eric and Wendy Schmidt Fund for Strategic Innovation (G-22-63228), John Templeton Foundation (\#62452), and Air Force Office of Scientific Research (FA9550-19-1-0354).
X.M. was supported by the NetSeed: Seedling Research Award of Northeastern University. A.-L.B. is supported by the European Union's Horizon 2020 research and innovation programme under grant agreement No 810115 – DYNASNET.

\section*{Data Availability}
The referenced dataset \emph{unarXive} is available at \href{https://doi.org/10.5281/zenodo.4313164}{https://doi.org/10.5281/zenodo.4313164}. The referenced dataset of the full texts of \emph{Nature} articles is provided by \emph{Nature}.
The textual dataset generated during the study is provided in the Supplementary Data.
The code used for this manuscript is available at \href{https://github.com/Barabasi-Lab/hidden-citation}{https://github.com/Barabasi-Lab/hidden-citation}.

\bibsplit[23]

\bibliography{HiddenCitation}


\blankpage


\section*{Supplementary material}

**Supporting Information Text**

Table S1. Statistics of joint arXiv-MAG citation networks in different disciplines.

|                  | Citing papers | Cited papers | Total     | Papers without a MAG id |
|------------------|---------------|--------------|-----------|-------------------------|
| Physics          | 708, 484      | 1, 314, 660  | 1, 650, 515 | 66, 283                 |
| Computer science | 120, 072      | 752, 603     | 818, 311  | 3, 512                  |
| Biology          | 10, 031       | 134, 352     | 140, 865  | 601                     |

## 1. Bibliographic Databases

**1.1 UnarXive: joint arXiv-MAG citation network for topic modeling**
The *unarXive* (1) assembles 994, 351 publications' full texts, annotated in-text citations, and links to metadata, built by merging the arXiv database (2) (Aug. 23, 1991–Jan. 31, 2019) with 2, 746, 288 records of the Microsoft academic graph (MAG) (3, 4). This allows us to construct a large-scale joint arXiv-MAG citation network for different fields:

*Physics.*—We selected full-text papers from arXiv categories with taxonomic names that start with "hep," "astro-ph," "cond-mat," "gr-qc," "math-ph," "phys," and "quant-ph," corresponding to specific disciplines of physics. We identified 708, 484 matched arXiv papers that are contained in the unarXive dataset, of which the full texts are readily parsed (with formulas deleted and reference anchors added) by unarXive (1). These arXiv papers cited 1, 314, 660 papers in total, which are identifiable by either a MAG id or an arXiv id, or both. Indeed, many of these papers have a MAG id but not an arXiv id, as they were not shared on arXiv. The overall joint arXiv-MAG citation network consists of 1, 650, 515 identifiable papers (citing papers + cited papers), among which 66, 283 papers ($\approx 4\%$) do not have a MAG id, possibly because these papers were not published or because of matching errors in unarXive. In practice, our model (Section 3) did not identify any foundational paper that does not have a MAG id, suggesting that these 66, 283 papers have limited influence. A summary of the statistics of the joint arXiv-MAG citation network in physics is given in Table S1.

*Computer science and biology.*—The arXiv categories start with "cs" and "q-bio," respectively. The analyses are the same as for physics (see Table S1).

We used the joint arXiv-MAG citation networks for topic modeling, extracting catchphrases and foundational papers by latent Dirichlet allocation (LDA) (5) (Section 3). Considering the similarity in the observed patterns for different disciplines, in the following we focus on physics unless otherwise noted.

**1.2 Nature: full-text corpus**
We also used a corpus of 88, 637 full-text papers (6) published in *Nature* (Nov. 4, 1869–Apr. 5, 2018). We used these papers, in parallel with the full-text arXiv corpus (2), to determine and compare the hidden citations of each topic from highly selective peer-reviewed vs. preprint venues, respectively (Section 4).

## 2. Citation Contexts Preprocessing

Citation contexts are already provided in the unarXive dataset (1). Still, it is necessary to preprocess the data before using them as the input of LDA. For example, there are 8,252 citation contexts in total for Maldacena's paper (2166248051) (7) in unarXive. An example of a citation context looks like:

*In this way, one is led to conjecture a remarkable duality that relates conventional (non-gravitational) quantum field theory on the boundary to string theories or M-theory on the bulk. This gauge theory/string theory duality is often referred to as anti-de Sitter/conformal field theory (AdS/CFT) correspondence, which was first spelled out in a seminal paper by Maldacena in 1997 MAINCIT . (A detailed review on AdS/CFT correspondence was given in Ref.)*

Note that "MAINCIT" is an anchor to indicate where Maldacena's paper was cited.

The preprocessing procedure consists of 3 steps:

1. Identify and expand abbreviations. We noticed that, in the main text of most papers, the first appearance of an abbreviation is often included in a pair of parentheses, and the full phrase is often given before the parentheses. These appearances have been collected in the citation contexts of the unarXive dataset (see example above). Thus, for each paper, the full version of a phrase and its abbreviation can be located by matching the special characters of "(" and ")" as well as the initials of the phrase and the corresponding letters in the abbreviation. A map between phrases and abbreviations can be generated for each paper. We also added a handful of common abbreviations which have well-documented meanings in physics fields:

   PAMELA→payload for antimatter matter exploration and light-nuclei astrophysics,
   SQUID→superconducting quantum interference device,
   BCFT→boundary conformal field theory,
   DMRG→density matrix renormalization group,
   FPGA→field-programmable gate array,
   MEMS→micro-electromechanical system,
   OTOC→out-of-time-order correlator,
   POVM→positive operator-valued measure,
   QLVM→quasi-localized vibrational modes,
   SUSY→supersymmetry,
   WIMP→weakly interacting massive particle,
   AdS→anti-de Sitter,
   BEC→Bose-Einstein condensation,
   CFT→conformal field theory,
   CMB→cosmic microwave background,
   EFT→effective field theory,
   GRB→gamma-ray burst,
   KdV→Korteweg-de Vries,
   LRO→long-range order,
   MFT→mean-field theory,
   OQS→open quantum system,
   PBG→photonic band gap,
   PMT→photomultiplier tube,
   PWA→partial-wave amplitude,
   QCD→quantum chromodynamics,
   QCP→quantum critical point,
   QED→quantum electrodynamics,
   QFT→quantum field theory,
   QND→quantum nondemolition,
   RMT→random matrix theory,
   RWA→rotating wave approximation,
   SEM→scanning electron microscope,
   SMD→silicon microstrip detector,
   SRO→short-range order,
   SSM→standard solar model,
   SYM→super Yang Mills,
   UED→universal extra dimensional,
   VEV→vacuum expectation value,
   WZW→Wess-Zumino-Witten,
   BH→black hole,
   GR→general relativity,
   IR→infrared,
   MC→Monte-Carlo,
   RG→renormalization group,
   SM→standard model,
   UV→ultraviolet

   For each paper, we used the generated map between phrases and abbreviations (plus the common abbreviations) to locate the abbreviation strings from the citation contexts and replace them by the full-version phrases.

2. Delete stop words and build $n$-grams. We used the "DeleteStopwords" function of Mathematica to delete the stop words (such as "a," "the," etc.). We also counted academic writing abbreviations (like "Eq.," "Fig.," "Ref.," etc.) as well as punctuation as stop words. We left a delimiter at the original position of each stop word. Each citation context was then cut into consecutive $n$-grams separated by these delimiters.

3. Convert to word stems and transliterate. We used the "WordStem" function of Mathematica to convert every word in $n$-grams to the word stem form (such as "approximation"→"approxim," etc.). We then used the "Transliterate" function of Mathematica to transliterate and remove possible diacritics (such as "naïve"→"naive," etc.) of characters.

After preprocessing, our example of the citation context will look like:

{wai}, {led}, {conjectur}, {remark, dualiti}, {relat, convent}, {non, gravit}, {quantum, field, theori}, {boundari}, {string, theori}, {m, theori}, {bulk}, {gaug, theori, string, theori, dualiti}, {refer}, {anti, de, sitter, conform, field, theori}, {anti, de, sitter, conform, field, theori}, {correspond}, {spell}, {semin, paper}, {maldacena}, {1997}, {MAINCIT}, {detail, review}, {anti, de, sitter, conform, field, theori, correspond}, {given}.

We see that {anti, de, sitter, conform, field, theori} appears as a 6-gram; while {m, theori} is a 2-gram.

## 3. Latent Dirichlet Allocation (LDA) Training

|  GRAM | TFIDF |
|---:|---:|
| bose einstein conden | 23.179 |
| larg hadron collid | 21.870 |
| anti de sitter conform field theori correspond | 18.911 |
| minim supersymmetr standard model | 18.162 |
| black hole | 16.992 |
| dark matter | 16.843 |
| standard model | 16.793 |
| cosmic microwav background | 15.531 |
| conform field theori | 13.420 |
| weakli interact massiv particl | 12.708 |
| densiti function theori | 11.846 |
| gener rel | 11.691 |
| dynam mean field theori | 11.365 |
| quantum chromodynam | 11.124 |
| parton distribut function | 11.089 |
| densiti matrix renorm group | 11.063 |
| spin orbit coupl | 10.916 |
| entangl entropi | 10.600 |
| dark energi | 10.484 |
| string theori | 10.291 |
| ⋮ | ⋮ |

**Fig. S1. List of $n$-grams with the highest tf-idf scores extracted from the citation contexts.**

*Preparing training set.*—To prepare the training set, we first built the *document corpus* and the *dictionary* for LDA. To this end, we picked the top 10,000 cited papers from the joint arXiv-MAG citation network in physics. We noticed that the "most" cited paper was possibly an artifact of the unarXive dataset. That paper was named "hep-th" (8) and had been wrongly identified as being "cited" by all arXiv papers from the "hep-th" category. We deleted the paper and used the remaining 9,999 papers as the document corpus, where the most cited is Nielsen and Chuang's textbook (1631356911) on quantum information (9).

After preprocessing the citation contexts of the 9,999 papers, we collected all $n$-grams and deleted those that are (1) too short or too long, i.e., having a total string length smaller than 5 or larger than 50, and (2) 1-grams, i.e., single words. We deleted single words because many of them either are too common to carry useful information or are people's names that is not representative enough to be an eponym of a discovery.

The collection of all $n$-grams were further sorted by their term frequency–inverse document frequency (tf-idf) scores (10, 11). The tf-idf is a statistical measure that is widely used to evaluate the importance of a phrase to the document corpus. The tf-idf score of a specific $n$-gram in a specific document increases proportionally to the number of times it appears in the document, offset by the number of documents in the corpus that contain the $n$-gram, which helps to adjust for the fact that some general $n$-grams appear more frequently. We calculated the overall tf-idf score of an $n$-gram by taking the sum of its tf-idf scores across all documents in the corpus. We also multiplied the tf-idf score by $n$, compensating the rarity of long phrases. In Fig. S1, we list the $n$-grams with the highest tf-idf scores which were calculated using the "FeatureExtract" function of Mathematica. We took the top 100,000 $n$-grams, capturing the most important terminologies in the citation contexts, and built the dictionary from them.

Given the document corpus and the dictionary, we generated the LDA input, i.e., a list of "2-tuples" $\{w, d\}$, by (1) shrinking the citation contexts to only three $n$-grams ahead of the citation anchor (MAINCIT), and (2) deleting all $w$ not included in the dictionary. The first step was adopted because we found that using three $n$-grams was the most computationally efficient way to extract the topical patterns without losing too much information. Then, for each paper $d$ from the document corpus and for each $w$ from the dictionary, we add a "2-tuple" $\{w, d\}$ to the input list whenever there is an occurrence of $w$ in the citation contexts of $d$. The result was a list of $3,600,853$ $w$-$d$ pairs as the input to LDA.

*Training.*—The hyperparameters were chosen as $\alpha = 0.01$ and $\beta = 0.001$, as commonly used (12), and the number of latent topics $z$ was set to be $2,000$. We adopted a synchronized parallel version of the *collapsed Gibbs sampling* method for LDA training (12), rewrote it in *Mathematica*, and ran it on an 8-core i7 processor for 1200 iterations of sampling, which took 28 hours approximately. During the iterations, shuffling between processor cores was added to avoid any possible statistical bias of parallelization.

*Result.*—The output consisted of a total of $3,600,853$ "3-tuples" $\{w, z, d\}$. Since the LDA model is not a discriminative model but a *generative* model, we can calculate not only the *forward* conditional probabilities, $P(z|d)$ and $P(w|z)$, which are most usually considered (5), but also the *backward* ones, namely, $P(z|w)$ and $P(d|z)$. Here, we show a few examples of $P(z|w)$ (Fig. S2) and $P(d|z)$ (Fig. S3) for different $n$-gram $w$, topic $z$, and paper $d$. The error bars for each $P(z|w)$ and $P(d|z)$ are calculated by Bayesian estimation given the priors $\alpha$ and $\beta$, respectively. By definition, the catchphrases and the foundational papers are the $n$-grams $w$ and the papers $d$ that have $P(z|w)$ and $P(d|z)$ larger than the corresponding thresholds, $P_{\text{th}}^{\text{catch}}$

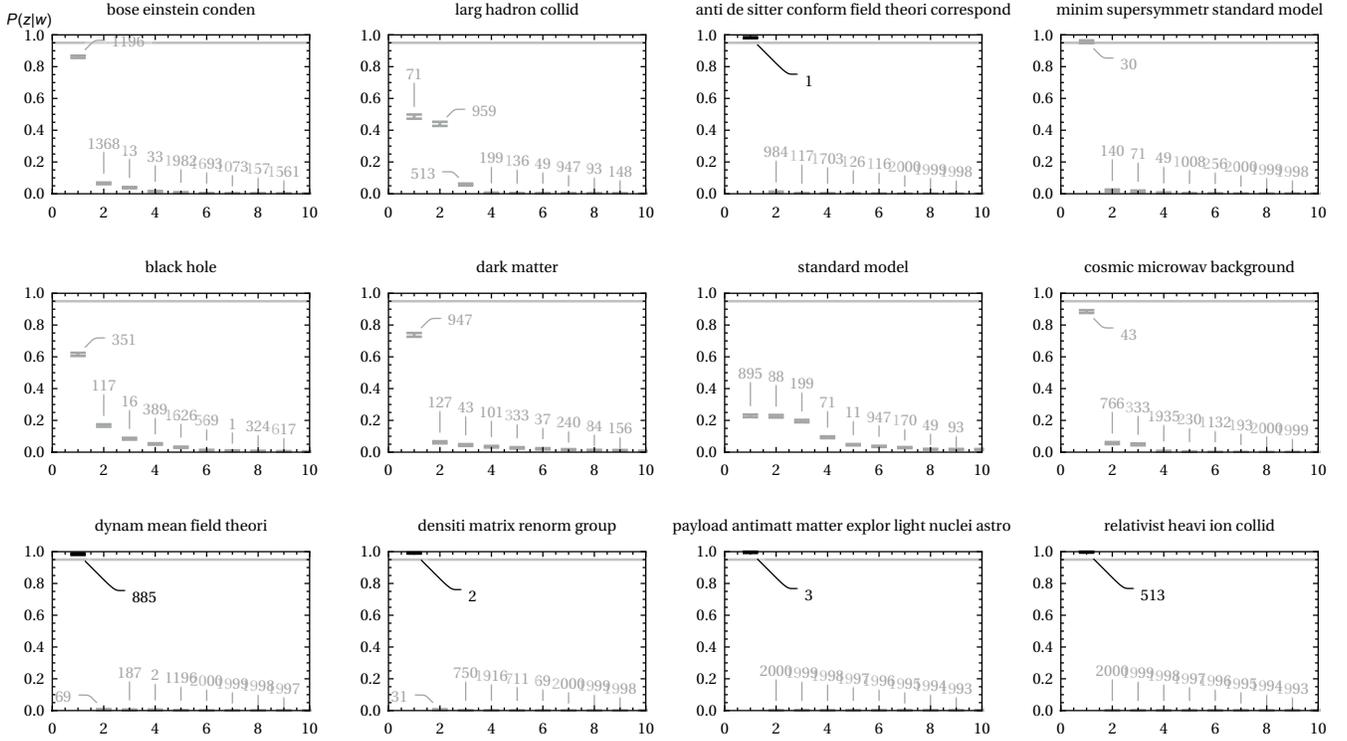

**Fig. S2. Conditional probability $P(z|w)$ for $n$-gram $w$.** Each latent topic $z$ is labeled by its topic id. If a $\{w, z\}$ pair satisfies $P(z|w) \geq P_{\text{th}}^{\text{catch}} = 0.95$ (light gray horizontal line), then the $n$-gram $w$ is considered as a catchphrase of the topic $z$. Error bars represent 95% confidence intervals.

and $P_{\text{th}}^{\text{found}}$, respectively. In practice, we compare the *lower error bars* (calculated by the Beta distribution with the Jeffreys prior, see Section 8) of $P(z|w)$ and $P(d|z)$ with the thresholds $P_{\text{th}}^{\text{catch}}$ and $P_{\text{th}}^{\text{found}}$. This way, we remain conservative in picking the catchphrases and foundational papers, especially when the counting of $w$ and $d$ is not sufficient to provide a confident estimation of $P(z|w)$ and $P(d|z)$. In other words, catchphrases and foundational papers of smaller topics are less likely to be identified. Moreover, we choose $P_{\text{th}}^{\text{catch}} = 0.95$ and $P_{\text{th}}^{\text{found}} = 0.05$, i.e., we rely on a strict criteria to choose catchphrases (high $P_{\text{th}}^{\text{catch}}$) but a loose criteria at including foundational papers (low $P_{\text{th}}^{\text{found}}$), reducing the *false positive* rate for identifying hidden citations, so that conservatively speaking, we would not see any "hidden citation" if credits to foundational papers had been properly allocated in existing citation patterns.

For example, "densiti matrix renorm group" is a catchphrase of topic 2, and White's paper (2037768897) (13) is a foundational paper of topic 2. This is because $P(z = 2|w = \text{"densiti matrix renorm group"}) \approx 0.991^{+0.002}_{-0.003}$, the lower error bar (derived by Bayesian estimation, see Section 8) of which is larger than 0.95; and $P(d = 2037768897|z = 2) \approx 0.464^{+0.010}_{-0.010}$, the lower error bar of which is larger than 0.05 (Figs. S2 and S3).

One exception we considered was when the foundational paper is a *book* or a *review paper*, in which case its loss of citations need not be concerned. Citations to a book or a review paper usually only serve for referencing a knowledge base, rather than pointing to specific scientific discoveries. Thus, we ignored foundational papers that are either identified by MAG as books or published in common review-paper venues including "Reviews of Modern Physics," "Physics Reports," "Reports on Progress in Physics," and "Physics Today." The rest, which include mostly journal and conference articles, were considered as true foundational papers.

Under our definition, some latent topics may not have corresponding catchphrases or foundational papers. The reason is that the latent topic is not specific enough, which may be attributed to either that our LDA model was not sufficiently trained, or that the topic itself has suffered so severe loss of citations that the correspondence between catchphrases and foundational papers can no longer be identified. For example, we cannot produce a catchphrase "schrodinger equat" ("Schrödinger equation") using LDA, because even though we know that the Schrödinger equation was first introduced by Schrödinger in 1926 (14), the citation contexts consisting of the 2-gram "schrodinger equat" have already been extremely diluted, giving preference to more recent papers. While such *false negative* results are inevitable, we can reduce the false negative rate by decreasing $P_{\text{th}}^{\text{catch}}$.

We picked from the 2,000 latent topics only the topics that have at least one catchphrase and one foundational paper and then collected these catchphrases and foundational papers. The final result consists of 497 catchphrases, 353 topics, and 880 foundational papers, covering most well-acknowledged modern topics in physics. A list of these topics can be found in the

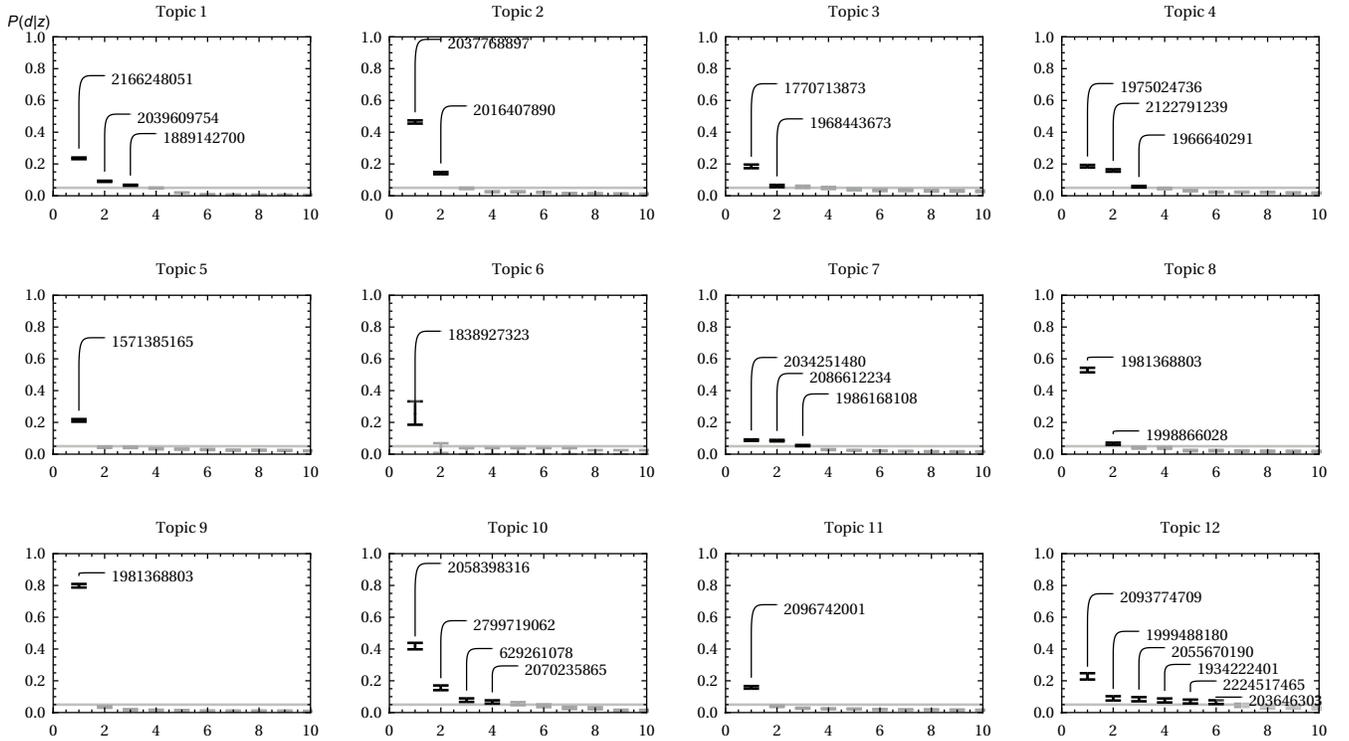

**Fig. S3. Conditional probability $P(d|z)$ for topic $z$.** Each paper $d$ is labeled by its MAG id. If a $\{z, d\}$ pair satisfies $P(d|z) \geq P_{\text{th}}^{\text{found}} = 0.05$ (light gray horizontal line), then the paper $d$ is considered as a foundational paper of the topic $z$. Error bars represent 95% confidence intervals.

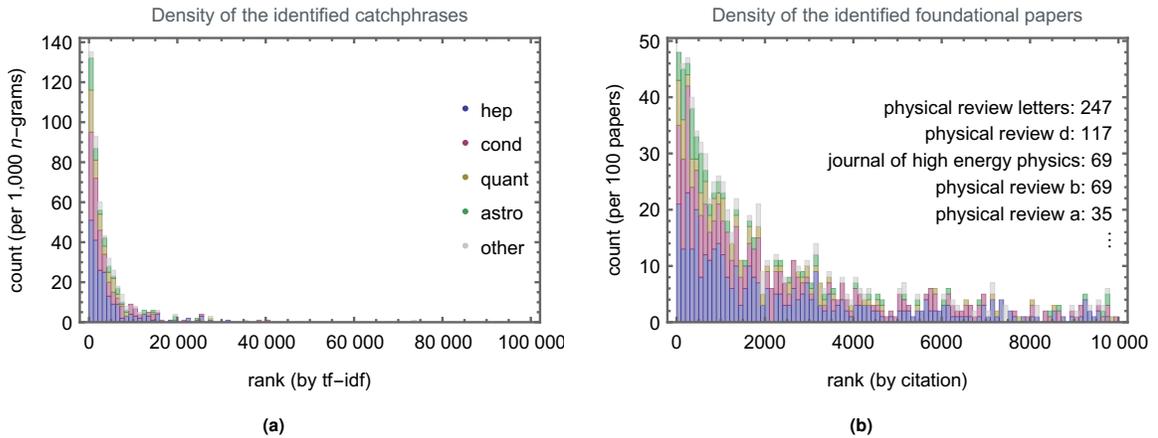

**Fig. S4. Densities of catchphrases and foundational papers extracted by LDA in physics.** **(a)** Catchphrases (per $1,000$ $n$-grams). The dictionary used by LDA contains $100,000$ $n$-grams, ranked by their tf-idf scores, among which $497$ catchphrases are identified. **(b)** Foundational papers (per $100$ papers). The document corpus used by LDA includes $9,999$ most cited papers, among which $880$ foundational papers are identified. Inset: top 5 journals that published the most foundational papers.

attached spreadsheet (Supplementary Data 1). We also calculated the densities of the 497 catchphrases and 880 foundational papers among the dictionary of 100,000 $n$-grams and the document corpus of 9,999 papers, as shown in Fig. S4. We see that the dictionary size and the document corpus size we chose for LDA training are large enough to potentially cover all useful catchphrases and foundational papers for identifying hidden citations: beyond the 100,000 $n$-grams and 9,999 papers, the densities of catchphrases and foundational papers become negligible. We also find that these foundational papers were mostly published in *Physical Review Letters* and other prestigious publication venues in physics.

Note that our method of choosing catchphrases and foundational papers using two thresholds $P_{\text{th}}^{\text{catch}}$ and $P_{\text{th}}^{\text{found}}$ is heuristic and, if needed, can be improved by a more elaborate algorithm. For example, instead of using a threshold of absolute value, we

may set a threshold on the relative change between the probabilities for the $i$th and $(i+1)$th catchphrases (or foundational papers). A catchphrase (foundational paper) becomes manifest whenever there is a huge gap in $P(z|w)$ $[P(d|z)]$ between it and its following $n$-grams (papers).

*Computer science and biology.*—The LDA training process for topics in computer science and biology was very similar to the training process for physics. However, considering the smaller size of these topics in arXiv, we used different parameters:

1. *Computer science:*
   Document corpus: citation contexts of top $5,000$ cited papers;
   Dictionary: top $50,000$ $n$-grams;
   Number of latent topics: $1,000$.

2. *Biology:*
   Document corpus: citation contexts of top $1,000$ cited papers;
   Dictionary: top $10,000$ $n$-grams;
   Number of latent topics: $200$.

The other parameters remained the same. The final results consist of 277 catchphrases, 206 topics, and 432 foundational papers for computer science, and 177 catchphrases, 94 topics, and 177 foundational papers for biology, respectively. In addition to the topics that are considered in the main text of our paper, below are a few more examples (catchphrases):

1. *Computer science:* gener adversari net, non orthogon multipl access, dijkstra algorithm, indian buffet process, polar code, ...

2. *Biology:* neuron avalanch, protein data bank, phylogenet invari, protein synthesi, speci tree, human connectom project, ...

Finally, we note that there may be room for further methodological improvements over LDA. For example, a hierarchical LDA model (15) could be used for identifing hierarchical subtopics, or one could implement a stochastic block model (16) to exploit different prior distributions other than the Dirichlet distribution. The self-consistency of the obtained results did not warrant further improvements on the tools.

## 4. Determine the Hidden Citations of Each Topic

To determine the hidden citations of each topic, we examined not only papers that explicitly cited the foundational papers ("citations"), but also papers that mentioned the corresponding catchphrases ("mentions"). The total number of citations and mentions represents the expected credits of the topic. The unarXive dataset provides full citation records from arXiv papers to MAG papers (arXiv "citing" MAG), as well as full texts of all considered arXiv papers (1). Hence, we can directly count the citations and mentions for each topic separately:

*Citations.*—For each topic, we took the MAG id's of all foundational papers and identified every citation to them in unarXive, among all 708, 484 arXiv papers in physics (Table S1). The collection of all papers that cit at least one of the foundational papers are considered as the citations to the topic.

*Mentions.*—For each topic, we search-and-matched the strings of the catchphrases in the full texts of all 708, 484 arXiv papers in physics (Table S1). The full texts were preprocessed in advance, following the same procedure for citation contexts described in Section 2. The collection of all papers that mention at least one of the catchphrases are considered as the mentions to the topic.

The arXiv papers that belong to "mentions" but not "citations" are considered as hidden citations.

*Computer science and biology.*—For topics in computer science and biology, in addition to arXiv, we also took advantage of a corpus of 88, 637 full-text Nature papers (6). The counting of the citations and mentions for each topic is the same as for arXiv. Note, however, that the Nature corpus does not include links to MAG. Hence, we manually searched the citation records of the foundational papers using Google Scholar, selected the citations from papers that published in Nature, matched them to our Nature corpus, and hence determined the citations of each topic within the scope of Nature.

## 5. Statistics of Topics

*Physics.*—We categorized the 353 topics into 5 categories that follow the arXiv taxonomy: high energy physics ("hep"), condensed matter physics ("cond"), quantum physics ("quant"), astrophysics ("astro"), and the rest ("other"), summarized in Table S2. A description of the arXiv taxonomy can be found at https://arxiv.org/category_taxonomy.

**Table S2. Topical categories by arXiv taxonomy.**

| Category | Field | ArXiv taxonomy | # of topics |
|---|---|---|---|
| "hep" | high energy physics | hep-ex, hep-lat, hep-... | 141 |
| "cond" | condensed matter physics | cond-mat.dis-nn, cond-mat.mes-hall, cond-mat... | 112 |
| "quant" | quantum physics | quant-ph | 38 |
| "astro" | astrophysics | astro-ph.CO, astro-ph.EP, astro-ph... | 31 |
| "other" | math physics, general physics | math-ph, gr-qc, physics.acc-ph, physics... | 31 |

We find that most topics are multidisciplinary, as their mentions often come from a variety of categories. To determine the main category of the topic, we choose the category with the largest number of mentions.

In the main paper, we chose a topic for each of the four categories ("hep," "cond," "quant," and "astro") as examples: "AdS/CFT (7, 17)," "DMRG (13, 18)," "discord (19)," and "BOSS (20–22)."

To examine the precision of the topic identification, we sought the expertise of two specialists in "hep" and "quantum" and conducted a manual review of ten randomly selected topics each (Table S3). In this review, we presented a list of catchphrases and asked the specialists to identify the paper they believed to be the most fundamental for the catchphrases. We conclude that most of the specialists' choices of the foundational papers are identified by the algorithm, demonstrating the effectiveness of our automated approach.

*Computer science and biology.*—We did not categorize topics into different categories due to the incomplete coverage of arXiv. In the main paper, we chose two topics as examples for computer science ("cs") and biology ("bio"), respectively: "Kalman (23)" ("cs"), "deep RL (24)" ("cs"), "ESS (25)" ("bio"), and "ViennaRNA (26–30)" ("bio").

**Table S3. Manual review of LDA results by specialists in "hep" and "quant."**

| Catchphrases ("hep") | Specialist's choice (paper title) | Identified by LDA |
|---|---|---|
| discret light cone quantiz | Solving field theory in one space and one time dimension | Y |
| minim composit higg model | The minimal composite Higgs model | N |
| bagger lambert gustavsson model, bagger lambert gustavsson | Gauge symmetry and supersymmetry of multiple M2 branes | Y |
| entangl wedg | The gravity dual of a density matrix | Y |
| larg volum scenario | Systematics of moduli stabilisation in Calabi–Yau flux compactifications | Y |
| new massiv graviti | Massive gravity in three dimensions | Y |
| color glass conden, balitski kovchegov, mclerran venugopalan model | Computing quark and gluon distribution functions for very large nuclei | Y |
| kerr conform field theori correspond | The Kerr/CFT correspondence | Y |
| balitski fadin kuraev lipatov | The Pomeranchuk singularity in nonabelian gauge theories | Y |
| giant graviton | Invasion of the giant gravitons from anti-de Sitter space | Y |

| Catchphrases ("quant") | Specialist's choice (paper title) | Identified by LDA |
|---|---|---|
| sideband cool | Sideband cooling of micromechanical motion to the quantum ground state | Y |
| continu spontan local | Combining stochastic dynamical state vector reduction with spontaneous localization | Y |
| einstein podolski rosen, einstein podolski rosen paradox | Can quantum mechanical description of physical reality be considered complete? | Y |
| calderbank shor stean code, calderbank shor stean | Multiple particle interference and quantum error correction | Y |
| entangl sudden death | Finite time disentanglement via spontaneous emission | Y |
| univ set | Elementary gates for quantum computation | Y |
| stochast local oper classic commun, bound entangl | Mixed state entanglement and distillation: Is there a "bound" entanglement in nature? | Y |
| non determinist polynomi problem complet | The complexity of theorem proving procedures | N |
| clauser horn shimoni holt, bell clauser horn shimoni holt inequ | Proposed experiment to test local hidden variable theories | Y |
| data process inequ, asymptot equipartit properti | A mathematical theory of communication | Y |

## 6. Sentiment Analysis of Topics

It is possible to go beyond simple citation counts and *differentiate* the credits that each explicit citation or hidden citation represents. For this, we ran a sentiment analysis on the sentences where the catchphrases were mentioned, noting that negative sentences should offer less credit than positive sentences. We used a natural language processing model, VADER (31), to examine the sentiments of both explicit and hidden citations. We find that a majority of citations are neutral sentiments, regardless of being explicit or hidden. For example, for the papers that explicitly cited the two foundational papers on "AdS/CFT," we extracted $27,289$ in-line sentences where the catchphrase "AdS/CFT" was mentioned, finding that $70.1\% \pm 0.5\%$ of the sentences are neutral. We also examined hidden citations, i.e., papers that mentioned the catchphrase "AdS/CFT" but did not cite the foundational papers, finding that $68.8\% \pm 0.5\%$ of the $32,052$ sentences are neutral. In other words, we find that the prevalence of sentimental texts is low in both explicit and hidden citations of foundational papers. This can be attributed to the fact that in order for a paper to become a foundational paper (and be able to acquire hidden citations), the discovery or technique it presents should have already been accepted as common knowledge, leaving little room for debates.

Among the sentences that are sentimental ($\sim 30\%$), we find that $89.7\% \pm 0.6\%$ express positive sentiments (for example: "...AdS/CFT duality is a beautiful relation among degrees of freedom in string theory...") and only $10.3\% \pm 0.6\%$ are negative sentiments (for example: "...Even though there are many important examples of AdS/CFT, a general condition that a given CFT should have its AdS dual has not been known completely..."). This ratio between positive/negative sentiments is once again similar for explicit and hidden citations. Our results suggest that hidden citations are indistinct from explicit citations—they have the same prevalence of sentiments and the same ratio between positive/negative sentiments. In other words, hidden citations are just an additional form of reference that has been hidden from traditional bibliometric analysis.

## 7. Topics on Dataset or Methodological Efforts

Topics like "BOSS" or "ViennaRNA" are rather complex. For example, two different topics are indeed closely related to the astrophysical terminology *baryon oscillation* (Table S4). The first topic corresponds to the general concept of the Baryon (Acoustic) Oscillation (BO), studying the acoustic density wave of matters caused by gravitational effects in the early universe. This concept was first introduced in a theoretical paper (Eisenstein, D. J., Hu, W. and Tegmark, M. ApJ, 1998), according to Bassett and Hlozek (32). The second topic, Baryon Oscillation Spectroscopic Survey (BOSS), corresponds to the observation of BO—measuring the standard length scale (the Statistical Standard Ruler) of the universe by collecting signals of acoustic waves from massive objects. Although the BO topic represents a theoretical discovery, the BOSS topic mainly represents experts' efforts towards building a database. Indeed, the catchphrase "BOSS" explicitly includes the term "survey" (a.k.a. dataset). Note that BOSS does not denote a single survey but multiple surveys that have been collected since 2000. The three foundational papers (2005, 2010, 2011) our algorithm identified corresponds to different subsequent surveys (SDSS-I, Early Data Release, that includes 14 million detected objects; SDSS-II, Data Release 7, that includes 230 million detected objects; and the 6dF Galaxy Survey, that includes 9,000 galaxies), each offering a more complete dataset (hence improving the accuracy of BOSS), not introducing general concepts or discoveries.

**Table S4. Comparison between BOSS and BO.**

| Catchphrase | Topic | Foundational paper |
|---|---|---|
| baryon oscil spectroscop survei | BOSS | detection of the baryon acoustic peak in the large scale correlation function of sdss luminous red galaxies (the astrophysical journal, 2005) |
| | | baryon acoustic oscillations in the sloan digital sky survey data release 7 galaxy sample (monthly notices of the royal astronomical society, 2010) |
| | | the 6df galaxy survey baryon acoustic oscillations and the local hubble constant (monthly notices of the royal astronomical society, 2011) |
| baryon oscil/baryon acoust oscil | BO | Cosmic Complementarity: H_0 and Omega_m from Combining CMB Experiments and Redshift Surveys (the astrophysical journal, 1998) |

While the three papers (2005, 2010, 2011) our algorithm identified are foundational for BOSS, they were frequently cited in the studies of the general concept of BO as well. When citing these papers in the context of BO, authors often fail to textually add the word "survey." This explains why there were more than half of the papers that did not mention the full catchphrase of BOSS (because "survey" is missing) but cited the three papers [dark red histogram in Fig. 1(f) in the main text]. The same pattern was observed for the "Vienna RNA package" topic [Fig. 4(a) in the main text], representing a programming package for analyzing RNA structures.

This textual bias of missing the keyword "survey" (or "package") supports our finding that database (or methodological) efforts are less acknowledged. This is because when cited in the BO context, the three papers (2005, 2010, 2011) are not considered as foundational for the general concept of BO but only as supportive for BO. As a result, the papers lost their merits as the foundational papers of BOSS and lost the acknowledgement of their efforts in building up the surveys. The BO community benefited from the accessible database without explicitly acknowledging the effort that went into creating it.

## 8. Bayesian Estimation of $p(\text{cite}|\text{mention})$

We denote the number of arXiv papers that mentioned the catchphrase(s) by $N_m$ (mentions), and among which the number of papers that cited the foundational paper(s) by $N_{c \cap m}$ (citations ∩ mentions). Naïvely, the probability for papers that mentioned the catchphrases to cite the foundational paper(s), $p(\text{cite}|\text{mention})$, can be estimated by the ratio of frequencies, $N_{c \cap m}/N_m$. However, such a frequentist estimation assumes $p(\text{cite}|\text{mention})$ is a determined number and ignores the probabilistic distribution of $p(\text{cite}|\text{mention})$, namely, a "probability of probability." This ignorance may produce a false estimation of the compound probability (e.g., sum of probabilities) of $p(\text{cite}|\text{mention})$.

The more accurate method is to use the Bayesian estimation, so that the probabilistic distribution of $p(\text{cite}|\text{mention})$ is given by a Beta distribution,

$$p(\text{cite}|\text{mention}) \sim \mathcal{B}(N_{c \cap m} + 1/2, N_m - N_{c \cap m} + 1/2), \qquad [1]$$

where the two 1/2's in Eq. (1) are the Jeffreys prior used for our Bayesian estimation. The expectation of Eq. (1) is given by

$$\langle p(\text{cite}|\text{mention}) \rangle = (N_{c \cap m} + 1/2)/(N_m + 1), \qquad [2]$$

which is different from the frequentist estimation, especially when $N_m$ and $N_{c \cap m}$ are small.

This Bayesian estimation with the Jeffreys prior is used throughout our paper, for example, when calculating conditional probabilities $P(z|w)$ and $P(d|z)$ in Section 3. All ± signs and error bars in our main text and SI denote two-sided 95% confidence interval in the estimation of the mean, which can be calculated given the Beta distribution [Eq. (1)].

## 9. Bayesian Estimation of the Temporal Change of $p(\text{cite}|\text{mention})$

We aim to calculate the temporal change of $p(\text{cite}|\text{mention})$ for each individual topic. To this end, for each topic, we divide its timespan into a series of nonoverlapping windows by calendar years (from the first calendar year when a catchphrase was mentioned to the year of the most recent record); for each time window, we count only $N_{c\cap m}$ and $N_m$ that occurred within the window and use them to calculate the Beta distribution of $p(\text{cite}|\text{mention})$ [Eq. (1)]. Therefore, for each window we have an independent distribution of $p(\text{cite}|\text{mention})$.

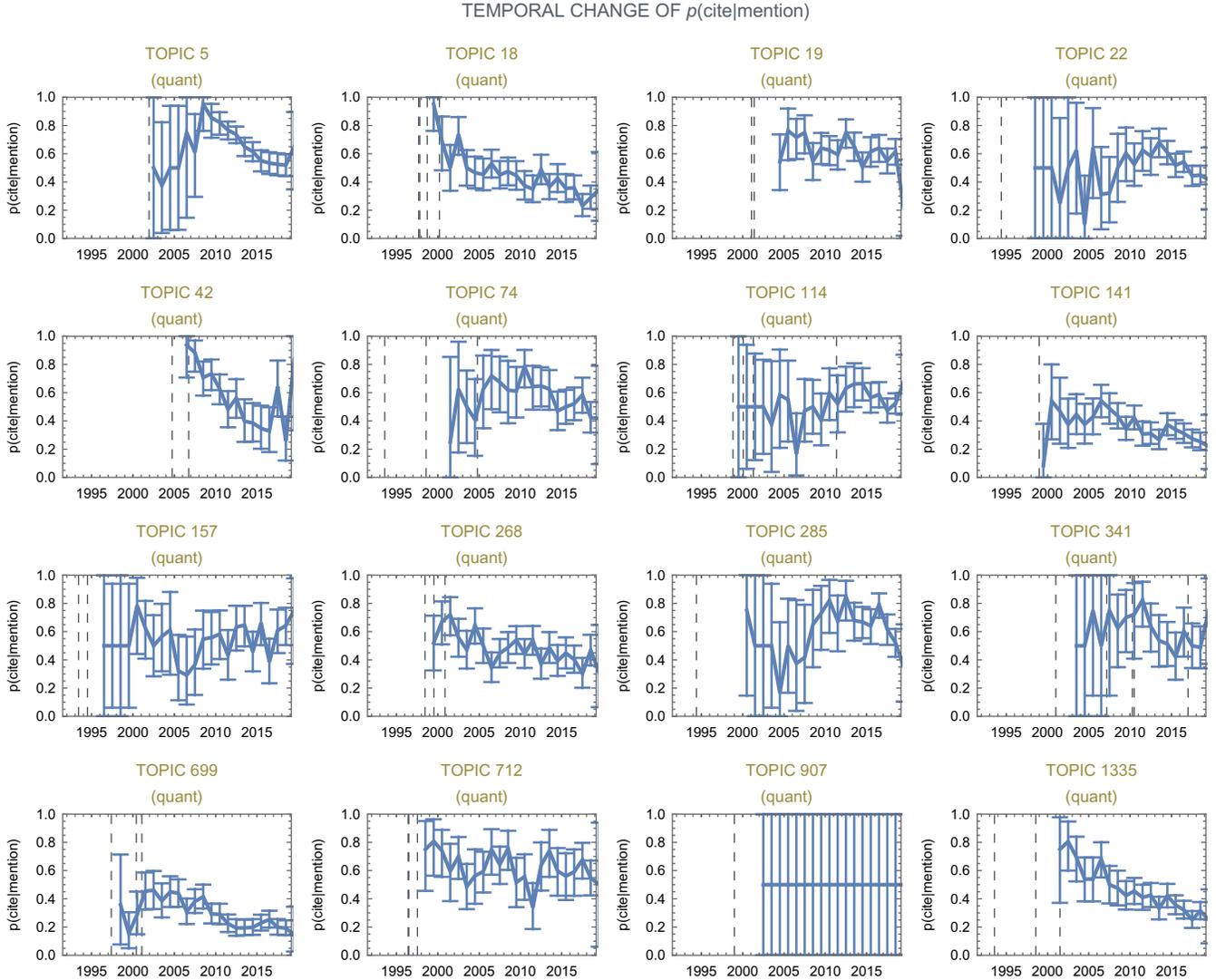

**Fig. S5. Temporal change of $p(\text{cite}|\text{mention})$ for 16 examples from the "quant" category.** Error bars represent 95% confidence intervals.

In Fig. S5, we show $p(\text{cite}|\text{mention})$ for 16 topics from the "quant" category. The solid line connects the expectation value $\langle p(\text{cite}|\text{mention})\rangle$ [Eq. (2)] for each window, while the error bars denote the 2.5% quantiles of the Beta distribution for both the upper and the lower. The error bars are asymmetric since the Beta distribution is not symmetric around $\langle p(\text{cite}|\text{mention})\rangle$.

The change of $p(\text{cite}|\text{mention})$ over time is captured by

$$\delta_t = \ln\left[p_t(\text{cite}|\text{mention})\right] - \ln\left[p_{t-1}(\text{cite}|\text{mention})\right], \qquad [3]$$

which is the logarithmic change of $p_t(\text{cite}|\text{mention})$ over $t$. Note that $\delta_t \approx (p_t - p_{t-1})/p_{t-1}$. Here, we use $t = 1, 2, \cdots$ to label each time window from the beginning per topic. The distribution of $\delta_t$ calculated by the compound probability rule on $p_t$ and $p_{t-1}$ (which follow the Beta distribution) is nontrivial. Denote with $\delta_t \sim \mathcal{X}(N_{c\cap m}^t, N_m^t, N_{c\cap m}^{t-1}, N_m^{t-1})$, where, by the compound

probability rule, the probability distribution function (PDF) of $\mathcal{X}$ is given by

$$\begin{cases} C^{-1} e^{-\left(\frac{1}{2}+N_{c\cap m}^{t-1}\right)\delta} B(1+N_{c\cap m}^{t-1}+N_{c\cap m}^{t}, \frac{1}{2}+N_m^t-N_{c\cap m}^t) \cdot \\ \quad {}_2F_1(\frac{1}{2}-N_m^{t-1}+N_{c\cap m}^{t-1}, 1+N_{c\cap m}^{t-1}+N_{c\cap m}^t, \frac{3}{2}+N_m^t+N_{c\cap m}^{t-1}, e^{-\delta}), & \delta \geq 0, \\ C^{-1} e^{\left(\frac{1}{2}+N_{c\cap m}^{t}\right)\delta} B(1+N_{c\cap m}^{t-1}+N_{c\cap m}^{t}, \frac{1}{2}+N_m^{t-1}-N_{c\cap m}^{t-1}) \cdot \\ \quad {}_2F_1(\frac{1}{2}-N_m^t+N_{c\cap m}^t, 1+N_{c\cap m}^{t-1}+N_{c\cap m}^t, \frac{3}{2}+N_m^{t-1}+N_{c\cap m}^t, e^{\delta}), & \delta < 0, \end{cases} \quad [4]$$

where $C = B(\frac{1}{2}+N_{c\cap m}^{t-1}, \frac{1}{2}+N_m^{t-1}-N_{c\cap m}^{t-1}) B(\frac{1}{2}+N_{c\cap m}^{t}, \frac{1}{2}+N_m^{t}-N_{c\cap m}^{t})$. Here, $B(x,y)$ is the beta function (not to be confused with the Beta distribution, $\mathcal{B}$), and ${}_2F_1(a,b,c;z)$ is the hypergeometric function. The expectation $\langle \delta_t \rangle$ has to be numerically calculated (integrated) from the PDF of $\mathcal{X}$ [Eq. (4)], as there is no analytical solution, and one should be aware that $\langle \delta_t \rangle \neq \ln(N_{c\cap m}^t/N_m^t) - \ln(N_{c\cap m}^{t-1}/N_m^{t-1})$ which would be the naïve frequentist estimation. Similarly, the variance $\langle \delta_t^2 \rangle - \langle \delta_t \rangle^2$ can also be calculated numerically.

We first calculate the temporal change of $p(\text{cite}|\text{mention})$ for each individual topic; then, at each window $t$, we calculate $\langle \delta_t \rangle \pm \sqrt{\langle \delta_t^2 \rangle - \langle \delta_t \rangle^2}$ for each topic; finally, at each window $t$, we take the average of $\langle \delta_t \rangle \pm \sqrt{\langle \delta_t^2 \rangle - \langle \delta_t \rangle^2}$ for all topics. The average is to be weighted by the uncertainty of each $\langle \delta_t \rangle$. The consequent weighted average over all topics (plus-minus uncertainty) will simply be denoted by $\mu_t \pm \sigma_t$. This line of calculation is a reflection of the belief that topics from the same category should have identical temporal change of $p(\text{cite}|\text{mention})$ at each time window $t$, and this identical temporal change is estimated to be $\mu_t$, with uncertainty $\sigma_t$. Finally, we accumulate and exponentiate $\mu_t \pm \sigma_t$ along $t = 1, 2, \cdots$, giving an accumulated time series: $\exp[\mu_1 \pm \sigma_1] - 1$, $\exp\left[(\mu_1 + \mu_2) \pm \sqrt{\sigma_1^2 + \sigma_2^2}\right] - 1$, $\exp\left[(\mu_1 + \mu_2 + \mu_3) \pm \sqrt{\sigma_1^2 + \sigma_2^2 + \sigma_3^2}\right] - 1, \cdots$. This time series is the final result of the (accumulated) *relative* temporal change of $p(\text{cite}|\text{mention})$ per category.

In practice, we cannot use all 353 topics to calculate the temporal change of $p(\text{cite}|\text{mention})$ for the following reasons:

1. For some topics, the earliest foundational paper was published earlier than 1991, the first year of records in arXiv. Therefore, we cannot track the full time evolution of $p(\text{cite}|\text{mention})$ for these topics, since there are no arXiv records available right after the publication of the earliest foundational paper.

2. For some topics, the earliest foundational paper was still noticeably later ($\sim$ more than one year) than the first mention of the catchphrases. This is due to the limitation of the LDA model, which could miss very early foundational papers if the topics are already too general nowadays or the early papers themselves are missing from the dataset. We cannot track the full time evolution of $p(\text{cite}|\text{mention})$ for these topics either, since the earliest of the possible foundational papers was incorrectly determined.

Eliminating these topics, we are left with 127 topics, including 55 topics in "hep," 35 topics in "cond," 17 topics in "quant," 10 topics in "astro," and 10 topics in "other," of which the time evolution of $p(\text{cite}|\text{mention})$ can be fully tracked, shown in our paper.

## 10. Indirect Citations versus Implicit Citations

To distinguish *indirect citations* (where catchphrases do not cite any foundational papers but cite papers that in turn cite foundational papers) versus *implicit citations* (where catchphrases do not cite anything at all), we repeated the temporal analysis but this time also including indirect citations as part of the citation count (cite + indirectly cite). We remain conservative and only consider indirect citations as hidden citations that have a citation path length of at most 2 to the foundational papers (i.e., papers that cite a foundational paper). We find that in this case the temporal drop of the probability of foundational papers being cited disappeared. Indeed, now $p\,[(\text{cite} + \text{indirectly cite})|\text{mention}]$ increases with time (Fig. S6). This is since, while the number of foundational papers remains constant, the number of papers that indirect citations cited—namely, papers that have a citation path length of 1 to the foundational papers—increases over time. Therefore, $p\,(\text{indirectly cite}|\text{mention})$, the chance of citing a paper that cited the foundational papers, rises rapidly, offsetting the temporal decrease of $p\,(\text{cite}|\text{mention})$, resulting in a net temporal increase of $p\,[(\text{cite} + \text{indirectly cite})|\text{mention}]$.

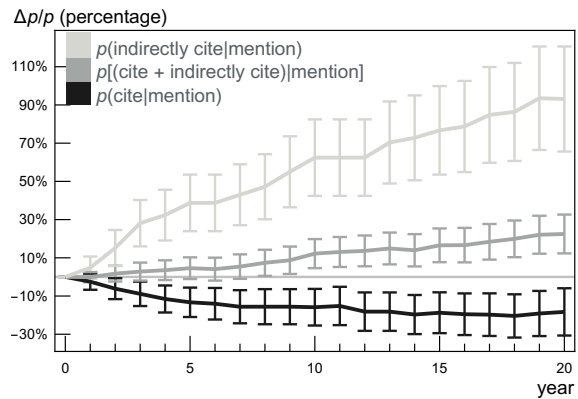

**Fig. S6.** **Comparison of the temporal changes of** $p(\textbf{cite}|\textbf{mention})$**,** $p\,(\textbf{indirectly cite}|\textbf{mention})$**, and** $p\,[(\textbf{cite} + \textbf{indirectly cite})|\textbf{mention}]$**.** Error bars represent 95% confidence intervals.

## 11. Cocitation Network of Foundational Papers

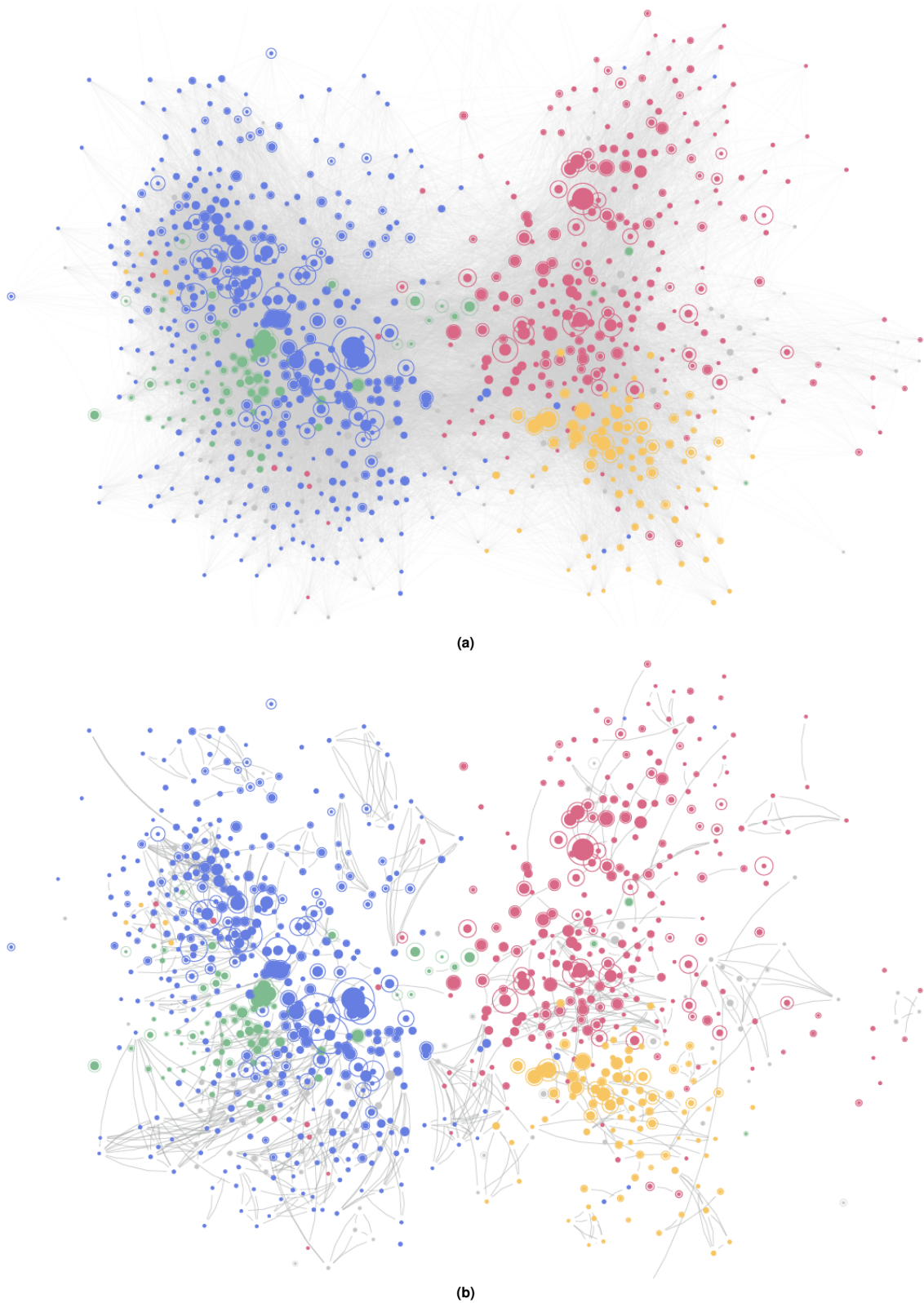

**Fig. S7. Cocitation network between** $880$ **foundational papers in physics.** Each node represents a foundational paper with both explicit citations (small core in the figure) and hidden citations (larger circle). Two nodes are linked if their in-degree cosine similarity $d(i, j) > c$. **(a)** $c = 0$. **(b)** $c = 0.2$.

**Xiangyi Meng, Onur Varol, and Albert-László Barabási**

We reallocate the hidden citations of each topic to its *individual* foundational paper(s). Assuming that each follower of the topic only needs to cite one of the foundational paper(s), we divided and assigned the papers that mentioned but not cited the topic to each foundational paper $d$, in proportion to its $P(d|z)$ (Fig. S3), since $P(d|z)$ can be considered as the representativeness of each foundational paper. For example, we find that for the "AdS/CFT" topic, the ratio between the $P(d|z)$ of the two foundational papers (2166248051) (7) and (2039609754) (17) is approximately $69.1\% : 30.9\%$. Therefore, given $13,447$ hidden citations of the topic "AdS/CFT" in total, we assigned $9,298.09$ to (2166248051) (7), and $4,148.91$ to (2039609754) (17).

After assigning the hidden citations to each foundational paper, we collected all 880 foundational papers, and their citations and hidden citations, allowing us to build a cocitation network (6) (Fig. S7). Explicit citations (small core in the figure) and hidden citations (larger circle) come from arXiv papers. Two foundational papers $i$ and $j$ are connected by a cocitation link based on the in-degree cosine similarity $d(i,j)$ between nodes $i$ and $j$ in the joint arXiv-MAG citation network (Section 1). The in-degree cosine similarity measures the number of common in-degree neighbors (i.e., the number of papers that cite both $i$ and $j$) divided by the geometric mean of their in-degrees (i.e., the square root of the product between the number of papers citing $i$ and the number of papers citing $j$). Two foundational papers $i$ and $j$ are connected if and only if $d(i,j) > c$. Here, we show the results with $c = 0$ (two foundational papers are linked if they are cocited by at least *one* arXiv paper) [Fig. S7(a)] and $c = 0.2$ [Fig. S7(b)].

The layout of nodes was automatically generated using the "GravityEmbedding" method of the "GraphPlot" function of Mathematica, applied to the cocitation network with $c = 0$. Nodes are closer if they share more links among each other. The butterfly shape of the network confirms the division between high energy physics and condensed matter physics and indicates that quantum physics is cocited with condensed-matter papers, and that astrophysics is largely indistinguishable from high energy physics in terms of cocitation patterns.

We also detected 394 communities in the $c = 0.2$ cocitation network [Fig. S7(b)] using a modularity maximization method provided by the function "FindGraphCommunities" of Mathematica. We observe that the community structure [Fig. S8(a)], where the matrix entry $(i,j)$ is 1 when foundational papers $i$ and $j$ belong to the same community, is almost identical to the topical structure as identified by LDA [Fig. S8(b)], where $(i,j)$ is 1 when $i$ and $j$ come from the same topic, and different from a random one [Fig. S8(c)], where the labels of nodes are randomly reordered. We find that $74.4\%^{+2.7\%}_{-2.8\%}$ of the 966 links and $89.2\%^{+2.9\%}_{-3.2\%}$ of the 394 communities connect foundational papers that belong to the same topic. This strong similarity between the community structure, a network attribute, and the topical structure defined by natural language processing cross-validates our LDA results.

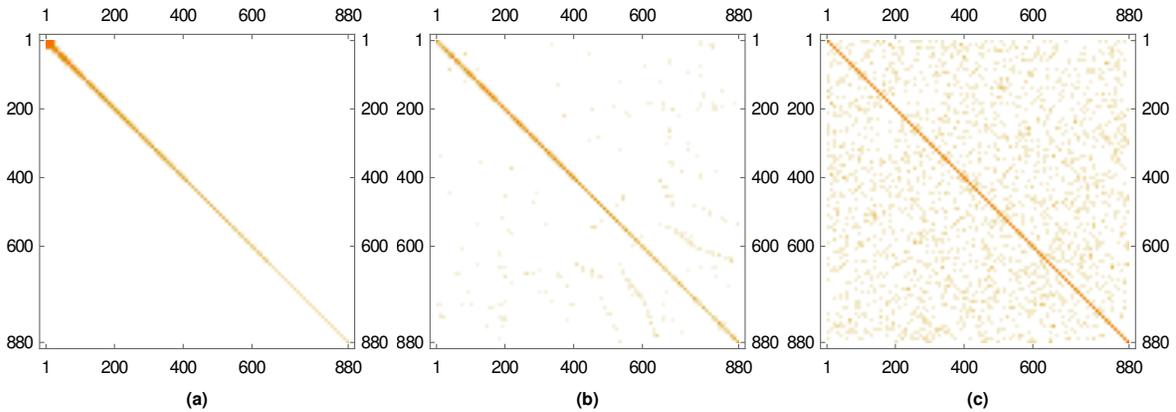

**Fig. S8. Community structure versus topical structure of the cocitation network. (a)** Community structure. Each matrix block represents a community [Fig. S8(a)]. **(b)** Topical structure. **(c)** Topical structure with randomly reordered nodes.

We also calculated the changes in citation-based ranks between foundational papers. Figure S9 shows the rank changes in different categories. We see that the majority of papers experienced changes in their ranks, irrespective of whether they possessed a high number of explicit citations or a relatively lower count.

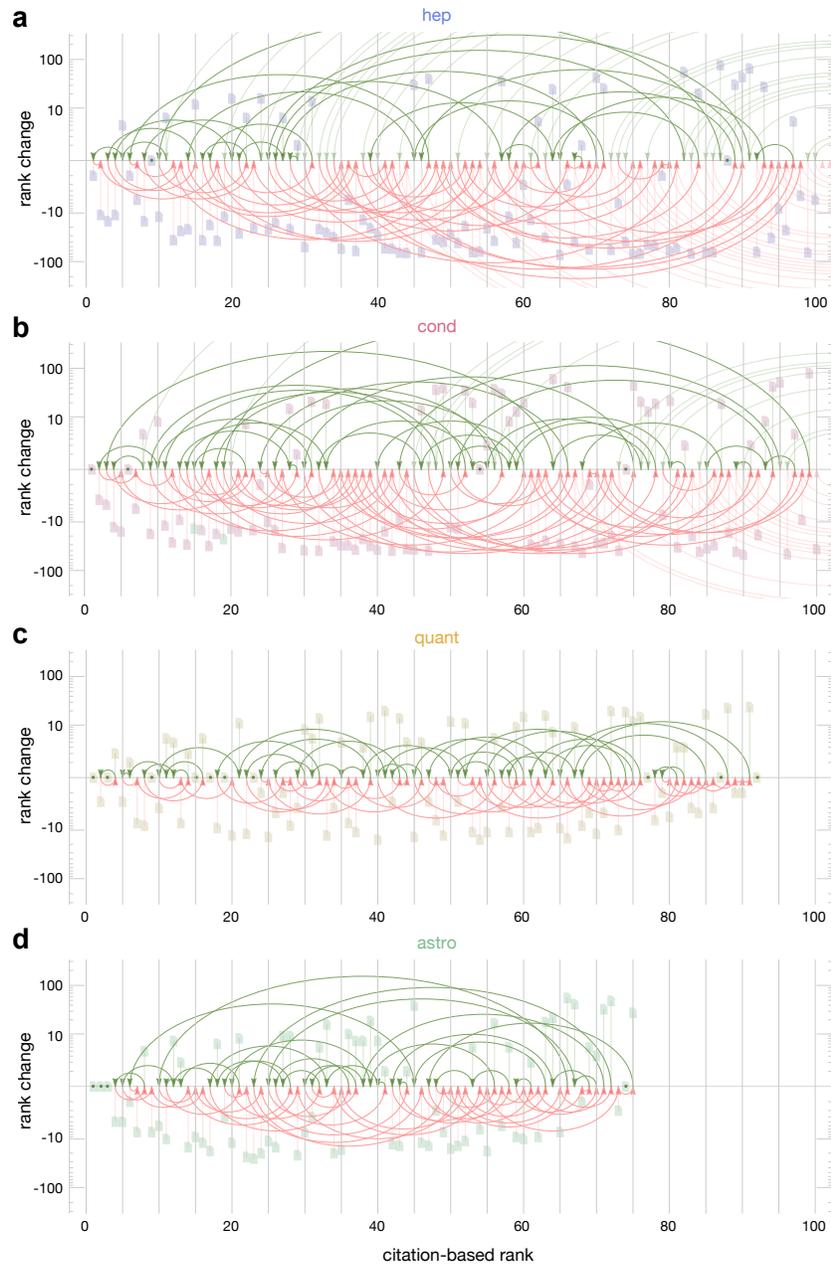

**Fig. S9. Citation-based ranks in different categories.** Changes in the citation rank of the top-ranked foundational papers after taking hidden citations into account, shown by arrows from the old explicit-citation-based rank to the new explicit-plus-hidden-citation-based rank (green: rank rise; red: rank drop). **(a)** hep. **(b)** cond. **(c)** quant. **(d)** astro.

## 12. Socialdemographic Aspects of Foundational Papers

We investigated the potential socialdemographic inequalities across foundational papers. To be specific, we successfully identified for 581 papers the author genders (of all 880 foundational papers). We find that the ratio between female-author-majority papers (28 papers) and male-author-majority papers (553 papers) is $5.1\%^{+2.8\%}_{-1.7\%}$, statistically indistinguishable from the gender ratio of $4.2\%^{+0.5\%}_{-0.5\%}$ for all papers in the corpus. Hence, while science is known to be characterized by a strong gender bias (33), this bias is not more (or less) notable for foundational papers. We also find no significant difference for the country of origins of papers that acquire hidden citations, relative to the publication volume of their country.

We also measured the prestige of the working units where papers were produced. For this, we used the Microsoft Academic Graph *institute saliency* metric (4). Specifically, we compared the institute saliency of the original institutions of published papers, finding that the average institute saliency for papers that acquired hidden citations was $77.5\% \pm 2.0\%$, compared to an average of $77.2\% \pm 0.6\%$ for all papers, again no significant difference. These results suggest that foundational papers can emerge in any institution, regardless of its level of prestige. Hidden citations also lack any specific gender or country of origin bias apart from the overall bias.

## 13. Relationship between Foundational Papers and Catchphrases

We are interested in the relationship between foundational papers and catchphrases. Since LDA is a generative model that automatically produces the joint probability $P(w, z, d)$ (Section 3), for each $n$-gram $w$ and paper $d$, we can directly calculate the conditional probabilities $P(w|d)$, of seeing a paper $d$ and linking it to a specific $n$-gram $w$, and $P(d|w)$, of seeing an $n$-gram $w$ and linking it to a specific paper $d$. For instance, we selected 12 Nobel-prize papers (34) that are identified by our model as foundational papers (Table S5). We find that the $n$-grams with the highest $P(w|d)$ for each foundational paper is textually very relevant to the scope of the work (Fig. S10). We also identified the corresponding catchphrases for the topics of these foundational papers, finding that each foundational paper also appears among the papers with the highest $P(d|w)$ for each catchphrase (Fig. S11). However, these foundational papers usually do not take the top spot in $P(d|w)$ but only the 2nd or 3rd, since the paper occupying the top spot is usually a review paper or a book (which were eliminated later in our analysis, see Section 3).

**Table S5. Nobel-prize foundational papers.**

| MAG id | Title (journal, year) |
|---|---|
| (2006294391) (35) | two dimensional magnetotransport in the extreme quantum limit (physical review letters, 1982) |
| (1996413343) (36) | broken symmetries and the masses of gauge bosons (physical review letters, 1964) |
| (2014437371) (37) | theory of superconductivity (physical review, 1957) |
| (1963826472) (38) | anomalous quantum hall effect an incompressible quantum fluid with fractionally charged excitations (physical review letters, 1983) |
| (2134803267) (39) | evidence for oscillation of atmospheric neutrinos (physical review letters, 1998) |
| (2055032269) (40) | reliable perturbative results for strong interactions (physical review letters, 1973) |
| (2030976617) (41) | inhomogeneous electron gas (physical review, 1964) |
| (1983987483) (42) | direct evidence for neutrino flavor transformation from neutral current interactions in the sudbury neutrino observatory (physical review letters, 2002) |
| (2058122340) (43) | electric field effect in atomically thin carbon films (science, 2004) |
| (1965245884) (44) | giant magnetoresistance of 001 fe 001 cr magnetic superlattices (physical review letters, 1988) |
| (2068718327) (45) | observation of a neutrino burst from the supernova sn1987a (physical review letters, 1987) |
| (2169090745) (46) | cp violation in the renormalizable theory of weak interaction (progress of theoretical physics, 1973) |

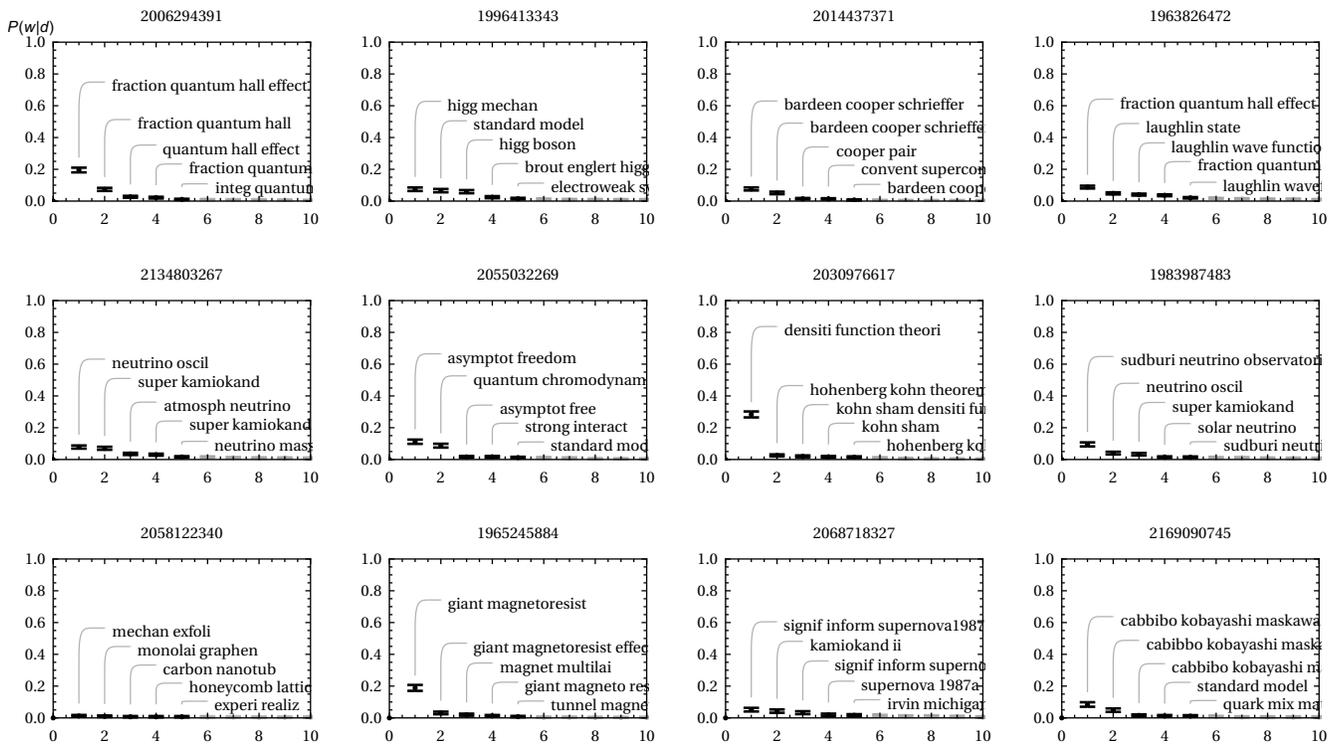

**Fig. S10. Conditional probability $P(w|d)$ for Nobel-prize foundational paper $d$.** Error bars represent $95\%$ confidence intervals.

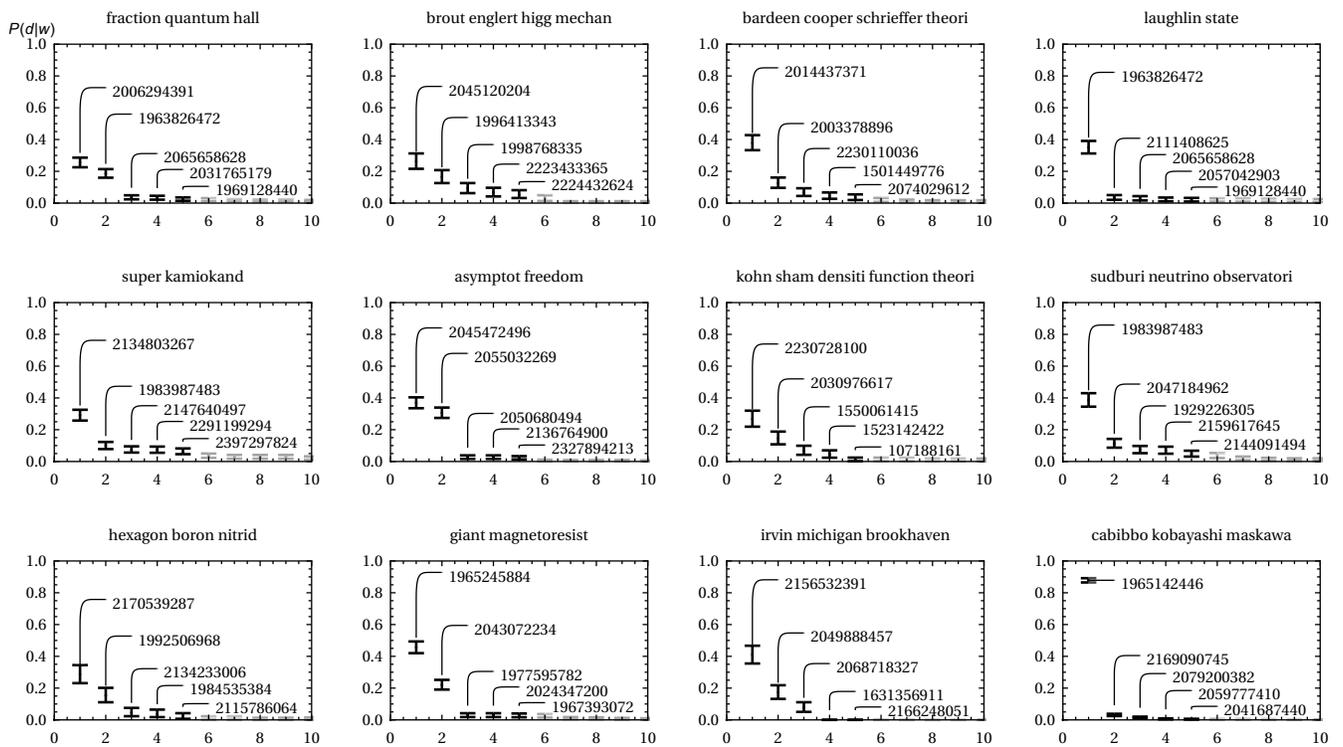

**Fig. S11. Conditional probability** $P(d|w)$ **for Nobel-prize catchphrases** $w$. Error bars represent $95\%$ confidence intervals.

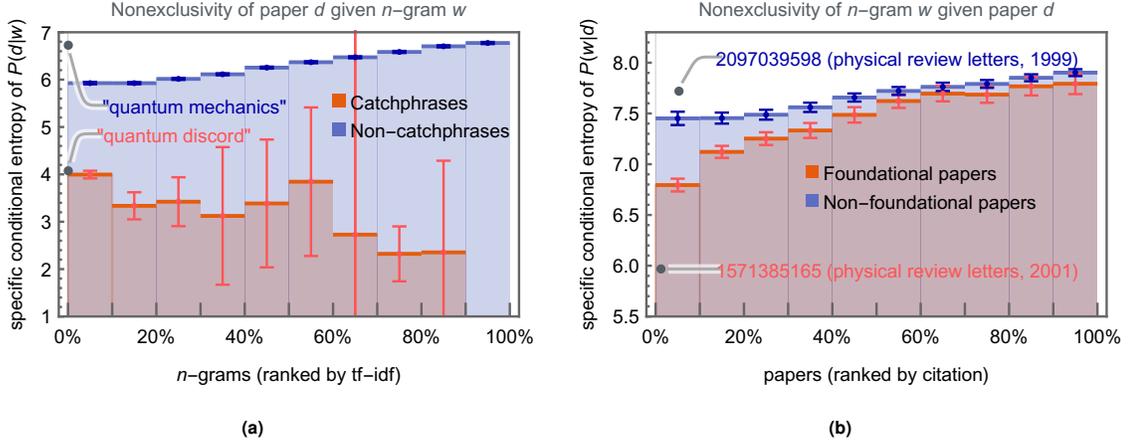

**Fig. S12. Specific conditional entropy** $S(d|w)$ **and** $S(w|d)$**. (a)** $S(d|w)$ measures the nonexclusivity of seeing an $n$-gram $w$ and linking it to a specific paper $d$. **(b)** $S(w|d)$ measures the nonexclusivity of seeing a paper $d$ and linking it to a specific $n$-gram $w$. Error bars represent 95% confidence intervals.

Our inspection suggests that foundational papers and their corresponding catchphrases are closely related. We proceeded by calculating the *specific conditional entropies*, $S(d|w)$ (given a specific $n$-gram $w$), as well as $S(w|d)$ (given a specific paper $d$), defined as

$$S(d|w) = -\sum_d P(d|w) \ln P(d|w),$$
$$S(w|d) = -\sum_w P(w|d) \ln P(w|d). \quad [5]$$

The specific conditional entropies directly measure the *nonexclusivity*: a lower $S(d|w)$ indicates that, seeing the $n$-gram $w$, one can link it more exclusively (i.e., with less uncertainty) to some paper $d$; and the same for a lower $S(w|d)$, that seeing the paper $d$, one can link it more exclusively to some $n$-gram $w$. We calculated $S(d|w)$ for each of the $100,000$ $n$-grams and $S(w|d)$ for each of the $9,999$ papers as used in our LDA model, separating the results for catchphrases vs. non-catchphrases as well as foundational papers vs. non-foundational papers.

By definition (Section 3), we expect that $S(d|w)$ should be lower for catchphrases than non-catchphrases, which is confirmed by our results [Fig. S12(a)]. This indicates that a catchphrase almost always points to one or a few well-defined papers compared to non-catchphrases. Conversely, we find that $S(w|d)$ is also lower for foundational paper than non-foundational paper [Fig. S12(b)], indicating that a foundational paper almost always points to one or a few well-defined phrases as well. Thus, the (catchphrase)–(foundational paper) pairs must acquire *mutual* exclusivity, a necessary condition for hidden citations to be developed.

## 14. Origins of Catchphrases

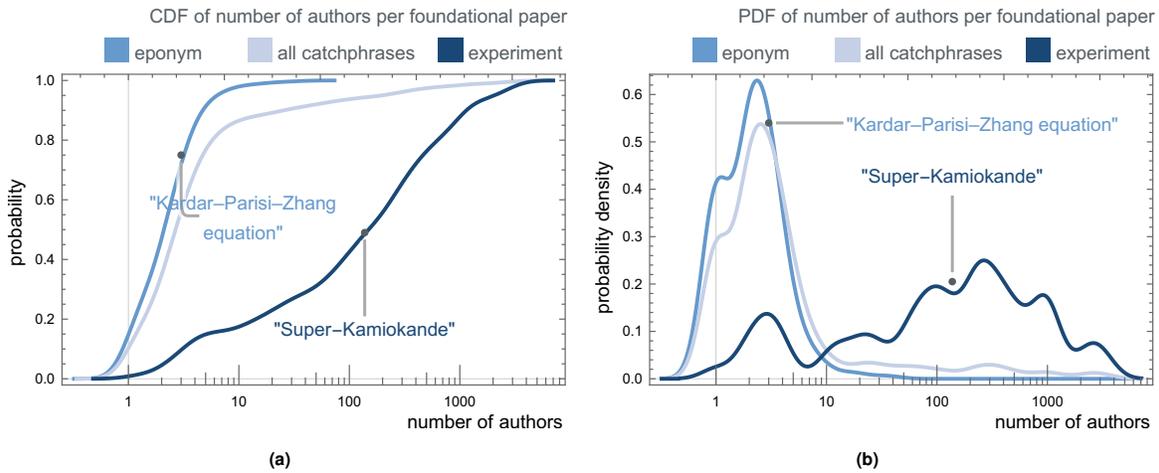

**Fig. S13. Probability distribution of the number of authors per foundational paper.** **(a)** Cumulative distribution function (CDF). **(b)** Probability distribution function (PDF). Typically, eponym-related foundational papers have fewer authors, while experiment-related foundational papers have more authors.

Next, we checked whether the catchphrases of a foundational paper have appeared in the foundational paper itself. Their absence suggests that these catchphrases were not proposed by the authors of the foundational papers, but were assigned later by the community. We find that 694 foundational papers (among all 880 foundational papers) that do not contain *any* of the corresponding catchphrases in the titles or abstracts, a proportion of 78.8% as discussed in the paper. As Microsoft Academic Graph (MAG) does not provide full texts for foundational papers, we successfully identified the arXiv preprint version for 435 foundational papers. Accounting for their full texts, we find 191 do not containing any catchphrase in their titles/abstracts/full-texts, by extrapolation suggesting that roughly half of all foundational papers do not contain the catchphrase in their titles/abstracts/full-texts. The catchphrases of these papers must come from the community.

We also find that many catchphrases are *eponyms*, i.e., carrying the initials of the names of authors. We find 229 foundational papers that have at least one eponym as catchphrases, amounting to $26.0\%^{+2.9\%}_{-2.8\%}$ of all foundational papers. In addition, catchphrases can also carry the names of *experimental projects* (catchphrases that contain grams like "experi," "observatori," "xenon," "collid," "detector," "survei," "satellit," "telescop," and "kamiokand"). We find 62 foundational papers that have at least one experiment name as catchphrases, amounting to $7.1\%^{+1.8\%}_{-1.6\%}$ of all foundational papers. As for each individual foundational paper, we counted the number of authors of the foundational papers that belong to eponym-related vs. experiment-related vs. all topics and show their histograms in the main paper. Here, we also show the normalized (and smoothed) cumulative distributions and probability density distributions of the number of authors (Fig. S13). Our result suggests that the community tends to assign eponyms as catchphrases to foundational papers with 1-3 authors and assign project names for foundational papers with hundreds of authors.

Our results justify our machine-learning methodology for identifying hidden citations: although LDA is a *gram-level* natural language processing model that can hardly be generalized to *sentence-level*, we find that for most well-acknowledged foundational papers, there will almost surely be a catchphrase that is not assigned by the authors themselves but the community (e.g., an eponym). This implies that the community is capable to (and tends to) distill the concept of the foundational paper as a catchphrase, not a full sentence.

**SI Dataset S1 (data_suppl.pdf)**

(Catchphrase)–(foundational paper) pairs within the field of physics, extracted using the latent Dirichlet allocation (LDA) technique from arXiv.

**Data S1: (catchphrase)-(founding paper) pairs.**

| topic id | catchphrase - ranked by tf-idf | founding paper (% of representativeness) - ranked by real citations | arxiv category |
|---|---|---|---|
| 1 | anti de sitter conform field theori correspond, holograph correspond | (2166248051) the large n limit of superconformal field theories and supergravity (1999) (60%), (2039609754) gauge theory correlators from non critical string theory (1998) (23.2%), (1889142700) anti de sitter space and holography (1998) (16.8%) | hep |
| 2 | densiti matrix renorm group, densiti matrix renorm group algorithm | (2037768897) density matrix formulation for quantum renormalization groups (1992) (76.5%), (2016407890) density matrix algorithms for quantum renormalization groups (1993) (23.5%) | cond |
| 3 | payload antimatt matter explor light nuclei astrophi, alpha magnet spectromet, advanc thin ioniz calorimet | (1770713873) (75.1%), (1968443673) pamela results on the cosmic ray antiproton flux from 60 mev to 180 gev in kinetic energy (2010) (24.9%) | astro |
| 4 | soft collinear effect theori | (1975024736) an effective field theory for collinear and soft gluons heavy to light decays (2001) (46.3%), (2122791239) soft collinear factorization in effective field theory (2002) (39.5%), (1966640291) invariant operators in collinear effective theory (2001) (14.2%) | hep |
| 5 | quantum discord | (1571385165) quantum discord a measure of the quantumness of correlations (2001) (100%) | quant |
| 7 | eigenst thermal hypothesi, eigenst thermali hypothesi | (2034251480) chaos and quantum thermalization (1994) (38.8%), (2086612234) quantum statistical mechanics in a closed system (1991) (37.5%), (1986168108) thermalization and its mechanism for generic isolated quantum systems (2008) (23.7%) | cond |
| 8 | gener gradient approxim, gradient approxim perdew burk ernzerhof | (1981368803) generalized gradient approximation made simple (1996) (89.1%), (1998866028) atoms molecules solids and surfaces applications of the generalized gradient approximation for exchange and correlation (1992) (10.9%) | cond |
| 9 | perdew burk ernzerhof | (1981368803) generalized gradient approximation made simple (1996) (100%) | cond |
| 10 | time depend densiti function theori | (2058398316) density functional theory for time dependent systems (1984) (65.4%), (2799719062) (24.3%), (2070235865) excitation energies from time dependent density functional theory (1996) (10.3%) | cond |
| 11 | grand unifi theori | (2096742001) unity of all elementary particle forces (1974) (100%) | hep |

| | | | |
|---|---|---|---|
| 12 | multi scale entangl renorm ansatz | (1999488180) entanglement renormalization and holography (2012) (14.6%), (2093774709) class of quantum many body states that can be efficiently simulated (2008) (37.4%), (1934222401) holographic quantum error correcting codes toy models for the bulk boundary correspondence (2015) (12.4%), (2036463035) classical simulation of quantum many body systems with a tree tensor network (2006) (10.6%), (2224517465) holographic duality from random tensor networks (2016) (11.3%), (2055670190) algorithms for entanglement renormalization (2009) (13.7%) | cond |
| 13 | bardeen cooper schrieffer bose einstein conden crossov | (2000530877) observation of resonance condensation of fermionic atom pairs (2004) (100%) | cond |
| 14 | standard model exten, lorentz violat | (2489872803) the standard model (1981) (49.9%), (2148545031) limits on a lorentz and parity violating modification of electrodynamics (1990) (21.8%), (2163842679) gravity lorentz violation and the standard model (2004) (28.2%) | hep |
| 16 | laser interferomet gravit wave observatori | (2252795400) observation of gravitational waves from a binary black hole merger (2016) (60.6%), (2437571239) gw151226 observation of gravitational waves from a 22 solar mass binary black hole coalescence (2016) (23%), (2765081049) gw170817 observation of gravitational waves from a binary neutron star inspiral (2017) (16.5%) | astro |
| 17 | gradient approxim | (1981368803) generalized gradient approximation made simple (1996) (74.2%), (2049079467) accurate and simple analytic representation of the electron gas correlation energy (1992) (12.4%), (1998866028) atoms molecules solids and surfaces applications of the generalized gradient approximation for exchange and correlation (1992) (13.4%) | cond |
| 18 | decoh free subspac, subsystem code, noiseless subsystem | (2017964195) decoherence free subspaces for quantum computation (1998) (37.1%), (2139625432) noiseless quantum codes (1997) (31.7%), (2087256203) theory of quantum error correction for general noise (2000) (17.6%), (2007536474) preserving coherence in quantum computation by pairing quantum bits (1997) (13.7%) | quant |
| 19 | measur base quantum comput | (2035993353) a one way quantum computer (2001) (66.8%), (1997837296) persistent entanglement in arrays of interacting particles (2001) (13.5%), (2761960687) (19.7%) | quant |
| 20 | compact muon solenoid, relev kinemat variabl | (2151512268) the cms experiment at the cern lhc (2008) (91.1%), (2079755804) observation of a new boson with mass near 125 gev in pp collisions at sqrt s 7 and 8 tev (2013) (8.94%) | hep |
| 21 | project entangl pair state, infinit project entangl pair state, matrix product oper | (2036604884) efficient classical simulation of slightly entangled quantum computations (2003) (29.8%), (2122816100) the density matrix renormalization group in the age of matrix product states (2010) (30.9%), (1570334424) renormalization algorithms for quantum many body systems in two and higher dimensions (2004) (39.3%) | cond |
| 22 | quantum fisher inform | (2083423624) statistical distance and the geometry of quantum states (1994) (100%) | quant |

| | | | |
|---|---|---|---|
| 24 | projector augment wave | (1970127494) projector augmented wave method (1994) (44.1%), (1979544533) from ultrasoft pseudopotentials to the projector augmented wave method (1999) (55.9%) | cond |
| 25 | dzyaloshinskii moriya interact, dzyaloshinskii moriya | (1963551001) spin current and magnetoelectric effect in noncollinear magnets (2005) (48.3%), (2735806552) anisotropic superexchange interaction and weak ferromagnetism (1960) (51.7%) | cond |
| 27 | larg synopt survei telescop, dark energi survei | (1956442392) euclid definition study report (2011) (21.9%), (2159714423) lsst from science drivers to reference design and anticipated data products (2019) (36.5%), (1918432732) lsst science book version 2 0 (2009) (31.8%), (2027101160) wide field infrared survey telescope astrophysics focused telescope assets wfirst afta 2015 report (2015) (9.75%) | astro |
| 28 | gener uncertainti principl | (2142397325) hilbert space representation of the minimal length uncertainty relation (1995) (62.5%), (2086439234) universality of quantum gravity corrections (2008) (20.7%), (2032174421) a generalized uncertainty principle in quantum gravity (1993) (16.8%) | hep |
| 31 | continu time quantum mont carlo | (2068539538) continuous time solver for quantum impurity models (2006) (62.5%), (2038941806) continuous time quantum monte carlo method for fermions (2005) (37.5%) | cond |
| 32 | quantum teleport, quantum den code, den code, quantum commun protocol, super den code | (1978553093) teleporting an unknown quantum state via dual classical and einstein podolsky rosen channels (1993) (51.9%), (2051051926) quantum cryptography based on bell s theorem (1991) (20.3%), (2020725861) communication via one and two particle operators on einstein podolsky rosen states (1992) (27.8%) | quant |
| 34 | composit fermion | (1995680910) composite fermion approach for the fractional quantum hall effect (1989) (56.5%), (2174568001) theory of the half filled landau level (1993) (43.5%) | cond |
| 35 | spin transfer torqu | (2014987330) emission of spin waves by a magnetic multilayer traversed by a current (1996) (100%) | cond |
| 36 | baryon oscil spectroscop survei | (2162447676) detection of the baryon acoustic peak in the large scale correlation function of sdss luminous red galaxies (2005) (37.7%), (2121951791) the 6df galaxy survey baryon acoustic oscillations and the local hubble constant (2011) (38%), (2113869105) baryon acoustic oscillations in the sloan digital sky survey data release 7 galaxy sample (2010) (24.3%) | astro |
| 37 | self interact dark matter | (2083034637) observational evidence for self interacting cold dark matter (2000) (100%) | astro |
| 40 | mode coupl theori, dynam heterogen, supercool liquid | (1965234283) testing mode coupling theory for a supercooled binary lennard jones mixture i the van hove correlation function (1995) (100%) | cond |

| # | Topic | Papers | Field |
|---|---|---|---|
| 41 | sachdev ye kitaev | (2007833526) gapless spin fluid ground state in a random quantum heisenberg magnet (1993) (36.1%), (2252917783) the spectrum in the sachdev ye kitaev model (2016) (16.2%), (2415090955) conformal symmetry and its breaking in two dimensional nearly anti de sitter space (2016) (9.95%), (2344612405) remarks on the sachdev ye kitaev model (2016) (15.2%), (1597956866) bekenstein hawking entropy and strange metals (2015) (12.3%), (2310057123) bi local holography in the syk model (2016) (10.1%) | cond |
| 42 | entangl sudden death | (2025028812) finite time disentanglement via spontaneous emission (2004) (74.2%), (2011359321) quantum open system theory bipartite aspects (2006) (25.8%) | quant |
| 43 | south pole telescop, planck collabor, wilkinson microwav anisotropi probe satellit, planck mission | (1909359706) planck 2015 results xiii cosmological parameters (2016) (40.3%), (2139937287) planck 2013 results xvi cosmological parameters (2014) (35.3%), (2147875900) seven year wilkinson microwave anisotropy probe wmap observations cosmological interpretation (2011) (24.4%) | astro |
| 44 | entangl spectrum | (2082257352) entanglement spectrum as a generalization of entanglement entropy identification of topological order in non abelian fractional quantum hall effect states (2008) (100%) | cond |
| 45 | maxim local wannier function | (2164805481) maximally localized generalized wannier functions for composite energy bands (1997) (62.5%), (2121448251) maximally localized wannier functions for entangled energy bands (2001) (25.8%), (2121624572) wannier90 a tool for obtaining maximally localised wannier functions (2008) (11.6%) | cond |
| 47 | potenti non relativist quantum chromodynam | (2008745657) potential nrqcd an effective theory for heavy quarkonium (2000) (50.4%), (2121797390) effective field theory for ultrasoft momenta in nrqcd and nrqed (1998) (49.6%) | hep |
| 50 | densiti function perturb theori | (2067328332) dynamical matrices born effective charges dielectric permittivity tensors and interatomic force constants from density functional perturbation theory (1997) (53.9%), (1976008838) green s function approach to linear response in solids (1987) (46.1%) | cond |
| 51 | greenberg horn zeil state, greenberg horn zeil | (1974928788) three qubits can be entangled in two inequivalent ways (2000) (100%) | quant |
| 52 | discret light cone quantiz | (2117709364) m theory as a matrix model a conjecture (1997) (50.9%), (2015548884) a large n reduced model as superstring (1997) (14.3%), (1616828411) another conjecture about m atrix theory (1997) (14.3%), (2127522773) solving field theory in one space and one time dimension (1985) (20.4%) | hep |
| 54 | einstein podolski rosen, einstein podolski rosen paradox | (1986407511) can quantum mechanical description of physical reality be considered complete (1935) (73.8%), (2055350506) realization of the einstein podolsky rosen paradox for continuous variables (1992) (13.5%), (2019335204) demonstration of the einstein podolsky rosen paradox using nondegenerate parametric amplification (1989) (12.6%) | quant |

| # | Topic | References | Field |
|---|---|---|---|
| 56 | balitski fadin kuraev lipatov | (2151327750) the pomeranchuk singularity in quantum chromodynamics (1978) (32.7%), (1858338496) the pomeranchuk singularity in nonabelian gauge theories (1977) (29.6%), (2112742282) bfkl pomeron in the next to leading approximation (1998) (10.1%), (1679675592) multi reggeon processes in the yang mills theory (1976) (18.8%), (2287414693) reggeization of the vector meson and the vacuum singularity in nonabelian gauge theories (1976) (8.81%) | hep |
| 57 | clauser horn shimoni holt inequ, local realism | (2028815089) proposed experiment to test local hidden variable theories (1969) (100%) | quant |
| 59 | decoi state method, decoi state | (1865778220) quantum key distribution (2008) (25.3%), (2055617162) quantum key distribution with high loss toward global secure communication (2003) (22.8%), (1716995571) beating the photon number splitting attack in practical quantum cryptography (2005) (19%), (2100257068) practical decoy state for quantum key distribution (2005) (15.5%), (2081059083) device independent quantum key distribution (2012) (17.4%) | quant |
| 60 | solar dynam observatori | (1971455763) effective hamiltonian for the superconducting cu oxides (1988) (17.1%), (2142234851) the atmospheric imaging assembly aia on the solar dynamics observatory sdo (2012) (12.2%), (2094483427) the solar dynamics observatory sdo (2012) (38.7%), (2111078819) the helioseismic and magnetic imager hmi investigation for the solar dynamics observatory sdo (2012) (32%) | astro |
| 63 | topolog massiv graviti | (2095543708) three dimensional massive gauge theories (1982) (58.7%), (2072028848) topologically massive gauge theories (1982) (41.3%) | hep |
| 65 | time evolv block decim | (2036604884) efficient classical simulation of slightly entangled quantum computations (2003) (36.7%), (2004203093) efficient simulation of one dimensional quantum many body systems (2004) (34.8%), (2021712328) real time evolution using the density matrix renormalization group (2004) (18%), (1550516781) matrix product density operators simulation of finite temperature and dissipative systems (2004) (10.4%) | cond |
| 67 | rydberg state | (2042458092) dipole blockade and quantum information processing in mesoscopic atomic ensembles (2001) (100%) | physics |
| 71 | standard model like higg boson | (1858542512) observation of a new particle in the search for the standard model higgs boson with the atlas detector at the lhc (2012) (54.2%), (2167727518) observation of a new boson at a mass of 125 gev with the cms experiment at the lhc (2012) (45.8%) | hep |
| 72 | dark energi spectroscop instrument | (1956442392) euclid definition study report (2011) (17.8%), (2530454890) cmb s4 science book first edition (2016) (14.8%), (1576070759) the desi experiment a whitepaper for snowmass 2013 (2013) (35.6%), (2547895327) the desi experiment part i science targeting and survey design (2016) (31.8%) | astro |

| 74 | discret time quantum walk | (1966092221) quantum random walks (1993) (31.6%), (1975528636) quantum computation and decision trees (1998) (24.9%), (1561344775) quantum speed up of markov chain based algorithms (2004) (18.6%), (92553147)  (24.9%) | quant |
| --- | --- | --- | --- |
| 76 | modifi newtonian dynam | (2763417190)  (23%), (1990148897) a modification of the newtonian dynamics as a possible alternative to the hidden mass hypothesis (1983) (22.5%), (2162982196) relativistic gravitation theory for the modified newtonian dynamics paradigm (2004) (9.94%), (2161525754) modified newtonian dynamics mond observational phenomenology and relativistic extensions (2012) (10.2%), (2132565565) convergence of probability measures (1970) (26.1%), (2115399036) modified newtonian dynamics as an alternative to dark matter (2002) (8.3%) | astro |
| 77 | minim flavour violat | (2150001361) minimal flavour violation an effective field theory approach (2002) (40.3%), (2146582577) weak scale effective supersymmetry (1990) (18.5%), (2020985823) universal unitarity triangle and physics beyond the standard model (2001) (22.2%), (2141022417) general minimal flavor violation (2009) (8.59%), (2159463196) composite technicolor standard model (1987) (10.5%) | hep |
| 78 | fuld ferrel larkin ovchinnikov | (2746484546)  (29.6%), (2104255960) crystalline color superconductivity (2001) (14.2%), (1976175889) superconductivity in a strong spin exchange field (1964) (56.2%) | cond |
| 80 | intern linear collid, compact linear collid, futur circular collid | (1998119675) the international linear collider technical design report volume 2 physics (2013) (28.7%), (1482677613) tesla technical design report part iii physics at an e e linear collider (2001) (15%), (2114816908) first look at the physics case of tlep (2014) (20%), (1828854676) ilc higgs white paper (2013) (11.3%), (1591258193) physics and detectors at clic clic conceptual design report (2012) (14%), (1952620202) ilc reference design report volume 1 executive summary (2007) (11.1%) | hep |
| 81 | cabibbo kobayashi maskawa, cabbibo kobayashi maskawa, quark mix matrix | (1965142446) unitary symmetry and leptonic decays (1963) (83.6%), (2079200382) parametrization of the kobayashi maskawa matrix (1983) (6.65%), (2169090745) cp violation in the renormalizable theory of weak interaction (1973) (9.79%) | hep |
| 82 | constrain random phase approxim | (1972871080) frequency dependent local interactions and low energy effective models from electronic structure calculations (2004) (49.2%), (2059315794) density functional calculation of effective coulomb interactions in metals (1991) (11.2%), (2023093804) density functional calculation of the parameters in the anderson model application to mn in cdte (1989) (17.3%), (2000596750) screened coulomb interaction in the maximally localized wannier basis (2008) (12.3%), (2034007959) ground states of constrained systems application to cerium impurities (1984) (10%) | cond |

| # | Topic | Papers | Field |
|---|---|---|---|
| 84 | navarro frenk white | (1532082951) the structure of cold dark matter halos (1996) (50.2%), (2060003036) a universal density profile from hierarchical clustering (1997) (49.8%) | astro |
| 87 | vacuum string field theori | (2166462696) string field theory around the tachyon vacuum (2001) (40.5%), (1574441858) classical solutions in string field theory around the tachyon vacuum (2001) (25.4%), (1509348833) ghost structure and closed strings in vacuum string field theory (2002) (34%) | hep |
| 89 | van der waal densiti function | (2003193361) van der waals density functional for general geometries (2004) (37.6%), (2059885388) accurate molecular van der waals interactions from ground state electron density and free atom reference data (2009) (29.1%), (2000536749) van der waals density functionals applied to solids (2011) (10.1%), (2043535587) van der waals density functional self consistent potential and the nature of the van der waals bond (2007) (12.8%), (2020072134) accurate and efficient method for many body van der waals interactions (2012) (10.4%) | cond |
| 90 | weak graviti conjectur | (2136697405) the string landscape black holes and gravity as the weakest force (2007) (62.3%), (2144658339) causality analyticity and an ir obstruction to uv completion (2006) (14.3%), (2059177657) on the geometry of the string landscape and the swampland (2007) (9.84%), (1840111988) fencing in the swampland quantum gravity constraints on large field inflation (2015) (13.6%) | hep |
| 91 | color flavor lock, color superconduct | (2096475520) cosmic separation of phases (1984) (20.1%), (2137222925) diquark bose condensates in high density matter and instantons (1998) (34.2%), (2168490876) the condensed matter physics of qcd (2000) (22.6%), (2075227120) color flavor locking and chiral symmetry breaking in high density qcd (1999) (23.1%) | hep |
| 93 | seesaw mechan, left right | (2166252135) neutrino mass and spontaneous parity nonconservation (1980) (55.3%), (2143468856) complex spinors and unified theories (1979) (24.6%), (2101073879) neutrino masses and mixings in gauge models with spontaneous parity violation (1981) (20.1%) | hep |
| 94 | sudburi neutrino observatori, solar neutrino experi | (2134803267) evidence for oscillation of atmospheric neutrinos (1998) (37.2%), (1983987483) direct evidence for neutrino flavor transformation from neutral current interactions in the sudbury neutrino observatory (2002) (39.9%), (2159617645) first results from kamland evidence for reactor antineutrino disappearance (2003) (22.8%) | hep |
| 95 | microwav induc resist oscil | (2020147709) type iib green schwarz superstring in plane wave ramond ramond background (2002) (42.7%), (2016949519) a new maximally supersymmetric background of iib superstring theory (2002) (25%), (2007471302) shubnikov de haas like oscillations in millimeterwave photoconductivity in a high mobility two dimensional electron gas (2001) (17.3%), (2127713223) exactly solvable model of superstring in plane wave ramond ramond background (2002) (15.1%) | cond |

| # | Topic | References | Field |
|---|---|---|---|
| 96 | clauser horn shimoni holt, bell clauser horn shimoni holt inequ | (2028815089) proposed experiment to test local hidden variable theories (1969) (100%) | quant |
| 97 | bern carrasco johansson relat | (2160023088) new relations for gauge theory amplitudes (2008) (27.7%), (2056432625) minimal basis for gauge theory amplitudes (2009) (23.5%), (1949904826) open closed vs pure open string disk amplitudes (2009) (19%), (2046528357) free energy of a nonuniform system i interfacial free energy (1958) (18.8%), (1969310996) the momentum kernel of gauge and gravity theories (2011) (11%) | hep |
| 99 | kardar parisi zhang equat | (2087496472) dynamic scaling of growing interfaces (1986) (100%) | cond |
| 101 | larg underground xenon | (2042260934) first results from the lux dark matter experiment at the sanford underground research facility (2014) (46.2%), (2097820102) dark matter results from 225 live days of xenon100 data (2012) (23.6%), (2534308951) results from a search for dark matter in the complete lux exposure (2017) (30.2%) | hep |
| 103 | dvali gabadadz porrati, dvali gabadadz porrati model | (1488720593) 4d gravity on a brane in 5d minkowski space (2000) (100%) | hep |
| 105 | super kamiokand | (2134803267) evidence for oscillation of atmospheric neutrinos (1998) (69.8%), (1983987483) direct evidence for neutrino flavor transformation from neutral current interactions in the sudbury neutrino observatory (2002) (15.7%), (2291199294) the super kamiokande collaboration (1999) (14.5%) | hep |
| 107 | kardar parisi zhang | (2087496472) dynamic scaling of growing interfaces (1986) (87.9%), (2038758925) large distance and long time properties of a randomly stirred fluid (1977) (12.1%) | cond |
| 109 | nuclear parton distribut function | (2044460160) eps09 a new generation of nlo and lo nuclear parton distribution functions (2009) (28.3%), (1967142479) models for the polarized parton distributions of the nucleon (2001) (13.3%), (2041287808) nuclear parton distributions at next to leading order (2004) (27.7%), (2149860191) determination of nuclear parton distribution functions and their uncertainties at next to leading order (2007) (15.2%), (2024624849) global analysis of nuclear parton distributions (2012) (15.4%) | hep |
| 111 | adam optim | (1522301498) adam a method for stochastic optimization (2014) (100%) | physics |
| 113 | lieb robinson bound | (2011290825) the finite group velocity of quantum spin systems (1972) (32.9%), (1975451401) lieb schultz mattis in higher dimensions (2004) (21.6%), (1977658351) lieb robinson bounds and the generation of correlations and topological quantum order (2006) (16.1%), (2080580319) spectral gap and exponential decay of correlations (2006) (16.5%), (1978774199) lieb robinson bounds and the exponential clustering theorem (2006) (12.8%) | cond |

| # | Topic | Papers | Field |
|---|---|---|---|
| 114 | quantum anneal | (1983624953) quantum annealing in the transverse ising model (1998) (33%), (1508413598) quantum computation by adiabatic evolution (2000) (30%), (2105614108) a quantum adiabatic evolution algorithm applied to random instances of an np complete problem (2001) (25.3%), (2024860775) quantum annealing with manufactured spins (2011) (11.7%) | quant |
| 115 | partial transpo, partial transpo criterion | (2011208902) entanglement of formation of an arbitrary state of two qubits (1998) (41.8%), (2034920898) separability criterion for density matrices (1996) (58.2%) | quant |
| 116 | massiv graviti, boulwar deser ghost, de rham gabadadz tollei, massiv graviti theori | (2097968935) resummation of massive gravity (2011) (37.7%), (2114832931) generalization of the fierz pauli action (2010) (29.6%), (2177175366) can gravitation have a finite range (1972) (17.8%), (2089960419) bimetric gravity from ghost free massive gravity (2012) (14.9%) | hep |
| 118 | new massiv graviti | (2009437393) massive gravity in three dimensions (2009) (57.9%), (2048156229) chiral gravity in three dimensions (2008) (10%), (2102742538) more on massive 3d gravity (2009) (32.1%) | hep |
| 119 | giant graviton | (2015533781) bubbling ads space and 1 2 bps geometries (2004) (22.3%), (1988959111) invasion of the giant gravitons from anti de sitter space (2000) (21.7%), (1658912092) exact correlators of giant gravitons from dual n 4 sym (2001) (20.4%), (1966028577) a toy model for the ads cft correspondence (2004) (13.5%), (2094446552) susy and goliath (2000) (11.3%), (2047366750) large branes in ads and their field theory dual (2000) (10.8%) | hep |
| 123 | topolog entangl entropi | (2046439097) topological entanglement entropy (2006) (41.8%), (2019024022) detecting topological order in a ground state wave function (2006) (35.3%), (2069840277) entanglement entropy and quantum field theory (2004) (22.9%) | cond |
| 124 | gross pitaevskii equat, gross pitaevskii | (2562972486) bose einstein condensation (1997) (100%) | cond |
| 126 | randal sundrum model, randal sundrum, randal sundrum scenario | (2095721433) an alternative to compactification (1999) (52.4%), (2064777467) large mass hierarchy from a small extra dimension (1999) (47.6%) | hep |
| 130 | inert doublet model, larg electron positron collid | (2173403199) search for the standard model higgs boson at lep (2003) (55.3%), (2180262730) pattern of symmetry breaking with two higgs doublets (1978) (17.3%), (2136821211) improved naturalness with a heavy higgs an alternative road to lhc physics (2006) (27.4%) | hep |
| 131 | asymmetr dark matter | (2031026718) secluded wimp dark matter (2008) (17.5%), (2023608666) asymmetric dark matter (2009) (38.9%), (2114165334) asymmetric dark matter theories signatures and constraints (2013) (24.1%), (2069880592) single explanation for both baryon and dark matter densities (1992) (19.5%) | hep |

| # | Topic | References | Field |
|---|---|---|---|
| 133 | atmosph imag assembl | (2132856166) a measurement of the damping tail of the cosmic microwave background power spectrum with the south pole telescope (2011) (14.4%), (2078533319) the atacama cosmology telescope cosmological parameters from three seasons of data (2013) (21.5%), (2142234851) the atmospheric imaging assembly aia on the solar dynamics observatory sdo (2012) (26%), (2146810479) a measurement of the cosmic microwave background damping tail from the 2500 square degree spt sz survey (2013) (10.5%), (2050602733) the atacama cosmology telescope temperature and gravitational lensing power spectrum measurements from three seasons of data (2014) (18.2%), (2062234728) detection of the power spectrum of cosmic microwave background lensing by the atacama cosmology telescope (2011) (9.41%) | astro |
| 134 | zero resist state | (2021079052) evidence for a new dissipationless effect in 2d electronic transport (2003) (40.2%), (2007471302) shubnikov de haas like oscillations in millimeterwave photoconductivity in a high mobility two dimensional electron gas (2001) (14.6%), (2107781596) radiation induced magnetoresistance oscillations in a 2d electron gas (2003) (13.2%), (2063351833) dynamical symmetry breaking as the origin of the zero dc resistance state in an ac driven system (2003) (21%), (1860678796) theory of microwave induced oscillations in the magnetoconductivity of a two dimensional electron gas (2005) (11%) | cond |
| 135 | except point | (2140726841) complex extension of quantum mechanics (2002) (54.6%), (2067123857) real spectra in non hermitian hamiltonians having pt symmetry (1998) (23.7%), (1999750768) spectral singularities of complex scattering potentials and infinite reflection and transmission coefficients at real energies (2009) (21.7%) | math |
| 136 | next to next to lead order, lead order cross section, amont carlo, madgraph madev, us madgraph5 | (2110999179) the automated computation of tree level and next to leading order differential cross sections and their matching to parton shower simulations (2014) (100%) | hep |
| 137 | dokshitz gribov lipatov altarelli parisi | (1989494434) asymptotic freedom in parton language (1977) (67%), (1543695929) deep inelastic e p scattering in perturbation theory (1972) (33%) | hep |
| 139 | fluctuat theorem, jarzynski equal, jarzynski relat | (2149801992) nonequilibrium equality for free energy differences (1997) (39.7%), (2021820673) entropy production fluctuation theorem and the nonequilibrium work relation for free energy differences (1999) (25%), (2056080528) dynamical ensembles in nonequilibrium statistical mechanics (1995) (18.8%), (1988388205) probability of second law violations in shearing steady states (1993) (16.5%) | cond |
| 141 | quantum repeat, long distanc quantum commun | (2092105345) quantum repeaters the role of imperfect local operations in quantum communication (1998) (100%) | quant |

| 143 | tensor renorm group | (1531177993) tensor renormalization group approach to two dimensional classical lattice models (2007) (29.7%), (2074814947) accurate determination of tensor network state of quantum lattice models in two dimensions (2008) (17.2%), (1987034448) renormalization of tensor network states (2010) (25%), (1887381766) tensor entanglement renormalization group approach as a unified method for symmetry breaking and topological phase transitions (2008) (11.1%), (2088160021) coarse graining renormalization by higher order singular value decomposition (2012) (17%) | cond |
|---|---|---|---|
| 145 | supersymmetri le houch accord | (2008813150) susy les houches accord 2 (2009) (49.3%), (2114510786) susy les houches accord interfacing susy spectrum calculators decay packages and event generators (2004) (50.7%) | hep |
| 148 | britto cachazo feng witten recur relat | (2165400029) perturbative gauge theory as a string theory in twistor space (2004) (33.7%), (2065937626) one loop n point gauge theory amplitudes unitarity and collinear limits (1994) (34.5%), (2046938033) new recursion relations for tree amplitudes of gluons (2005) (31.8%) | hep |
| 149 | hexagon boron nitrid | (2058122340) electric field effect in atomically thin carbon films (2004) (15%), (2134233006) two dimensional atomic crystals (2005) (15%), (2170539287) boron nitride substrates for high quality graphene electronics (2010) (34.5%), (2096421416) van der waals heterostructures (2013) (16.7%), (1992506968) one dimensional electrical contact to a two dimensional material (2013) (18.8%) | cond |
| 152 | einstein aether theori | (2162957442) gravity with a dynamical preferred frame (2001) (53.2%), (2141523732) einstein aether gravity a status report (2008) (30.4%), (2034156815) extended horava gravity and einstein aether theory (2010) (16.4%) | gr |
| 156 | fermi larg area telescop data, larg area telescop | (2167277890) the large area telescope on the fermi gamma ray space telescope mission (2009) (100%) | astro |
| 157 | spin squeez state | (2057883617) squeezed spin states (1993) (59.2%), (2059235845) squeezed atomic states and projection noise in spectroscopy (1994) (40.8%) | quant |
| 158 | siver function | (2137549325) time reversal odd distribution functions in leptoproduction (1998) (55.9%), (2174029682) single spin production asymmetries from the hard scattering of pointlike constituents (1990) (44.1%) | hep |
| 160 | poisson sigma model | (2106817500) a path integral approach to the kontsevich quantization formula (2000) (23.1%), (2169107602) the geometry of the master equation and topological quantum field theory (1997) (19%), (2104391478) poisson structure induced topological field theories (1994) (32.3%), (2023992287) two dimensional gravity and nonlinear gauge theory (1994) (25.5%) | math |

| # | Topic | Papers | Category |
|---|---|---|---|
| 161 | topolog crystallin insul | (1538825449) topological semimetal and fermi arc surface states in the electronic structure of pyrochlore iridates (2011) (100%) | cond |
| 166 | extend dynam mean field theori | (2074896825) low dimensional bose liquids beyond the gross pitaevskii approximation (2000) (18.1%), (2071461200) kosterlitz thouless transition and short range spatial correlations in an extended hubbard model (1996) (81.9%) | cond |
| 167 | andreev reflect | (1989116198) transition from metallic to tunneling regimes in superconducting microconstrictions excess current charge imbalance and supercurrent conversion (1982) (100%) | cond |
| 168 | continu spontan local | (2127669635) feynhiggs a program for the calculation of the masses of the neutral cp even higgs bosons in the mssm (2000) (16.2%), (2168001822) towards high precision predictions for the mssm higgs sector (2003) (19.9%), (2058509714) markov processes in hilbert space and continuous spontaneous localization of systems of identical particles (1990) (21.3%), (2030711967) the higgs boson masses and mixings of the complex mssm in the feynman diagrammatic approach (2007) (16.3%), (2161859157) the masses of the neutral cp even higgs bosons in the mssm accurate analysis at the two loop level (1999) (13.9%), (1994358697) combining stochastic dynamical state vector reduction with spontaneous localization (1989) (12.3%) | quant |
| 170 | higg inflat | (2115754103) the standard model higgs boson as the inflaton (2008) (53%), (2145954660) higgs mass and vacuum stability in the standard model at nnlo (2012) (26%), (2152418277) investigating the near criticality of the higgs boson (2013) (21%) | hep |
| 175 | martin siggia rose | (1990056747) statistical dynamics of classical systems (1973) (41.7%), (2087489306) bond operator representation of quantum spins mean field theory of frustrated quantum heisenberg antiferromagnets (1990) (17%), (2131883893) search for neutral mssm higgs bosons at lep (2006) (30.4%), (2130616914) dynamics as a substitute for replicas in systems with quenched random impurities (1978) (10.8%) | cond |
| 176 | linear muffin tin orbit | (2121431463) linear methods in band theory (1975) (80.7%), (2041383172) explicit first principles tight binding theory (1984) (19.3%) | cond |
| 177 | kerr conform field theori correspond | (2156366140) the kerr cft correspondence (2009) (61.1%), (2132082295) hidden conformal symmetry of the kerr black hole (2010) (17%), (2145531879) cft duals for extreme black holes (2009) (10.1%), (2035877031) black hole superradiance from kerr cft (2010) (11.8%) | hep |

| | | | |
|---|---|---|---|
| 178 | deep underground neutrino experi | (2094219754) universal noninteger ground state degeneracy in critical quantum systems (1991) (24.8%), (2341520670) the long baseline neutrino experiment exploring fundamental symmetries of the universe (2013) (38%), (2199785003) long baseline neutrino facility lbnf and deep underground neutrino experiment dune conceptual design report volume 2 the physics program for dune at lbnf (2015) (37.2%) | hep |
| 183 | hard thermal loop | (2056604715) scattering amplitudes in hot gauge theories (1989) (61.2%), (1975507567) production of soft dileptons in the quark gluon plasma (1990) (38.8%) | hep |
| 184 | laughlin state | (1963826472) anomalous quantum hall effect an incompressible quantum fluid with fractionally charged excitations (1983) (46.3%), (2111408625) fractional quantization of the hall effect a hierarchy of incompressible quantum fluid states (1983) (31.5%), (1969128440) beyond paired quantum hall states parafermions and incompressible states in the first excited landau level (1999) (22.2%) | cond |
| 186 | random energi model | (2083715726) random energy model an exactly solvable model of disordered systems (1981) (59%), (2001226941) random energy model limit of a family of disordered models (1980) (41%) | cond |
| 189 | inver spin hall effect, spin pump | (2123618453) enhanced gilbert damping in thin ferromagnetic films (2002) (74.9%), (2089480386) finite element theory of transport in ferromagnet normal metal systems (2000) (25.1%) | cond |
| 192 | iter perturb theori | (1983596827) hubbard model in infinite dimensions (1992) (32.4%), (2029256954) exact diagonalization approach to correlated fermions in infinite dimensions mott transition and superconductivity (1994) (15.1%), (2071432767) hubbard model in infinite dimensions a quantum monte carlo study (1992) (15.7%), (2001466065) mott hubbard transition in infinite dimensions (1992) (15.7%), (2165959915) new iterative perturbation scheme for lattice models with arbitrary filling (1996) (21.1%) | cond |
| 196 | giant magnetoresist, tunnel magnetoresist | (1965245884) giant magnetoresistance of 001 fe 001 cr magnetic superlattices (1988) (47.6%), (1977595782) theory of the perpendicular magnetoresistance in magnetic multilayers (1993) (12.3%), (2043072234) enhanced magnetoresistance in layered magnetic structures with antiferromagnetic interlayer exchange (1989) (21.7%), (1967393072) large magnetoresistance at room temperature in ferromagnetic thin film tunnel junctions (1995) (18.4%) | cond |
| 197 | daya bai, neutrino oscil experi, oscil paramet, reactor experi | (2107918562) observation of electron antineutrino disappearance at daya bay (2012) (100%) | hep |
| 201 | shear visco, fluid graviti correspond | (2097909025) viscosity in strongly interacting quantum field theories from black hole physics (2005) (47.4%), (1992872919) nonlinear fluid dynamics from gravity (2008) (52.6%) | hep |

| | | | |
|---|---|---|---|
| 203 | su schrieffer heeger | (2072703702) solitons in polyacetylene (1979) (72.3%), (2062089197) soliton excitations in polyacetylene (1980) (27.7%) | cond |
| 204 | de sitter conform field theori correspond | (2120830691) cosmological event horizons thermodynamics and particle creation (1977) (24%), (2135779038) the ds cft correspondence (2001) (39%), (2127210737) vacuum states in de sitter space (1985) (19.6%), (1641975541) quantum gravity in de sitter space (2001) (17.4%) | hep |
| 206 | calderbank shor stean code, calderbank shor stean | (2042474962) error correcting codes in quantum theory (1996) (29.6%), (2763302490) theory of quantum error correcting codes (1997) (19.3%), (2069711632) class of quantum error correcting codes saturating the quantum hamming bound (1996) (15.2%), (1017660896) error correcting codes (1988) (21%), (2131690084) multiple particle interference and quantum error correction (1996) (14.9%) | quant |
| 207 | anomali mediat supersymmetri break, anomali mediat | (1998924215) out of this world supersymmetry breaking (1999) (62.6%), (2131715088) gaugino mass without singlets (1998) (37.4%) | hep |
| 208 | bayesian inform criterion | (2168175751) estimating the dimension of a model (1978) (100%) | astro |
| 210 | gregori laflamm instabl, gregori laflamm | (2019649342) black strings and p branes are unstable (1993) (53.8%), (2087745624) the instability of charged black strings and p branes (1994) (18.3%), (1997075450) static axisymmetric vacuum solutions and non uniform black strings (2003) (15.5%), (2040122207) on non uniform black branes (2002) (12.4%) | hep |
| 211 | vienna ab initio simul packag, vienna ab initio simul packag code | (1970127494) projector augmented wave method (1994) (17.3%), (1979544533) from ultrasoft pseudopotentials to the projector augmented wave method (1999) (36.7%), (2007395042) efficiency of ab initio total energy calculations for metals and semiconductors using a plane wave basis set (1996) (21%), (2083222334) efficient iterative schemes for ab initio total energy calculations using a plane wave basis set (1996) (18.1%), (2079105963) ab initio molecular dynamics for liquid metals (1993) (6.92%) | cond |
| 214 | ginsparg wilson relat, domain wall fermion, overlap dirac oper, ginsparg wilson | (2135985420) a remnant of chiral symmetry on the lattice (1982) (30.7%), (2000898529) exactly massless quarks on the lattice (1998) (18.1%), (1546518322) a method for simulating chiral fermions on the lattice (1992) (12.8%), (2078463625) exact chiral symmetry on the lattice and the ginsparg wilson relation (1998) (11.2%), (2216314481) chiral fermions from lattice boundaries (1993) (9.85%), (2046698055) axial symmetries in lattice qcd with kaplan fermions (1995) (8.1%), (1979948273) more about exactly massless quarks on the lattice (1998) (9.21%) | hep |
| 222 | logarithm neg | (1984004232) computable measure of entanglement (2002) (75.1%), (1979369481) logarithmic negativity a full entanglement monotone that is not convex (2005) (24.9%) | quant |

| | | | |
|---|---|---|---|
| 225 | confin induc reson | (2014049935) atomic scattering in the presence of an external confinement and a gas of impenetrable bosons (1998) (100%) | cond |
| 226 | akaik inform criterion | (2142635246) a new look at the statistical model identification (1974) (66.7%), (2168175751) estimating the dimension of a model (1978) (33.3%) | astro |
| 229 | gottfri sum rule | (1506552701)  (46.7%), (73041511)  (27.9%), (2114929791) sum rule for high energy electron proton scattering (1967) (25.5%) | hep |
| 231 | finit differ time domain | (2142063750) numerical solution of initial boundary value problems involving maxwell s equations in isotropic media (1966) (100%) | physics |
| 232 | seiberg witten map | (1968388660) string theory and noncommutative geometry (1999) (77.1%), (2020600887) construction of non abelian gauge theories on noncommutative spaces (2001) (12.5%), (2062347297) enveloping algebra valued gauge transformations for non abelian gauge groups on non commutative spaces (2000) (10.3%) | hep |
| 238 | color glass conden, balitski kovchegov, mclerran venugopalan model | (2157155801) computing quark and gluon distribution functions for very large nuclei (1994) (34.6%), (2161737270) operator expansion for high energy scattering (1996) (25.2%), (2142299360) the colour glass condensate (2010) (15.9%), (1667268877) the color glass condensate and high energy scattering in qcd (2003) (12.3%), (1999089379) gluon distribution functions for very large nuclei at small transverse momentum (1994) (11.9%) | hep |
| 241 | werner state | (2062292260) quantum states with einstein podolsky rosen correlations admitting a hidden variable model (1989) (100%) | quant |
| 247 | bardeen cooper schrieffer theori | (2014437371) theory of superconductivity (1957) (100%) | cond |
| 252 | dynam mode decomposit | (2014356541) dynamic mode decomposition of numerical and experimental data (2010) (100%) | physics |
| 256 | minim composit higg model | (2152418277) investigating the near criticality of the higgs boson (2013) (47.9%), (2133022420) two loop renormalization group equations for soft supersymmetry breaking couplings (1994) (52.1%) | hep |
| 258 | eight vertex model | (2135745986) partition function of the eight vertex lattice model (1972) (100%) | math |
| 259 | littl higg model, littlest higg model, littl higg, composit higg | (2018548893) the minimal composite higgs model (2005) (25.1%), (1978052395) electroweak symmetry breaking from dimensional deconstruction (2001) (31%), (2077552005) the littlest higgs (2002) (43.9%) | hep |
| 262 | norm conserv pseudopotenti | (2045596260) efficient pseudopotentials for plane wave calculations (1991) (59.8%), (2087585288) soft self consistent pseudopotentials in a generalized eigenvalue formalism (1990) (9.54%), (2055640329) efficacious form for model pseudopotentials (1982) (21.2%), (2076858582) norm conserving pseudopotentials (1979) (9.41%) | cond |

| # | Topic | Papers | Category |
|---|---|---|---|
| 267 | nonrelativist quantum chromodynam | (2133238133) rigorous qcd analysis of inclusive annihilation and production of heavy quarkonium (1995) (87%), (2112702755) improved nonrelativistic qcd for heavy quark physics (1992) (13%) | hep |
| 268 | stochast local oper classic commun, bound entangl | (1974928788) three qubits can be entangled in two inequivalent ways (2000) (46.4%), (2003989474) mixed state entanglement and distillation is there a bound entanglement in nature (1998) (30%), (2141140927) conditions for a class of entanglement transformations (1999) (23.6%) | quant |
| 269 | small world network, suscept infect suscept, barabasi albert | (2008620264) emergence of scaling in random networks (1999) (39.2%), (2038195874) epidemic spreading in scale free networks (2001) (23.7%), (2112090702) collective dynamics of small world networks (1998) (37.2%) | cond |
| 271 | bagger lambert gustavsson theori, maldacena theori, maldacena model | (2171335381) n=6 superconformal chern simons matter theories m2 branes and their gravity duals (2008) (66.8%), (2166592280) gauge symmetry and supersymmetry of multiple m2 branes (2008) (14.5%), (2011546725) fractional m2 branes (2008) (18.7%) | hep |
| 279 | see saw mechan, type seesaw mechan | (2166252135) neutrino mass and spontaneous parity nonconservation (1980) (72.5%), (2143468856) complex spinors and unified theories (1979) (27.5%) | hep |
| 281 | coulomb blockad | (2103282498) quantum computation with quantum dots (1998) (100%) | cond |
| 285 | quantum cramer rao bound | (2083423624) statistical distance and the geometry of quantum states (1994) (100%) | quant |
| 287 | sakai sugimoto model, soft wall model | (2159787624) qcd and a holographic model of hadrons (2005) (32.1%), (2097479977) linear confinement and ads qcd (2006) (37%), (2091079419) low energy hadron physics in holographic qcd (2005) (30.9%) | hep |
| 288 | bodi local, bodi local transit | (2073848376) localization of interacting fermions at high temperature (2007) (35.9%), (2079602089) metal insulator transition in a weakly interacting many electron system with localized single particle states (2006) (34.7%), (2137084802) many body localization and thermalization in quantum statistical mechanics (2015) (29.4%) | cond |
| 290 | fuzzi sphere | (1968388660) string theory and noncommutative geometry (1999) (50.5%), (2142239943) noncommutative geometry and matrix theory compactification on tori (1998) (21.3%), (2094667859) the fuzzy sphere (1992) (28.2%) | hep |
| 291 | larg volum scenario | (2020332468) systematics of moduli stabilisation in calabi yau flux compactifications (2005) (68.1%), (2019920269) large volume flux compactifications moduli spectrum and d3 d7 soft supersymmetry breaking (2005) (31.9%) | hep |
| 293 | dirac born infeld inflat | (2035433892) non gaussian features of primordial fluctuations in single field inflationary models (2003) (15.4%), (2128595492) scalar speed limits and cosmology acceleration from d cceleration (2004) (38%), (2144653019) observational signatures and non gaussianities of general single field inflation (2007) (23.7%), (2131261154) dbi in the sky (2004) (22.9%) | hep |

| # | Topic | References | Field |
|---|---|---|---|
| 305 | nekrasov partit function | (2598139253) seiberg witten prepotential from instanton counting (2003) (100%) | hep |
| 307 | asymptot freedom | (2045472496) ultraviolet behavior of non abelian gauge theories (1973) (53.1%), (2055032269) reliable perturbative results for strong interactions (1973) (46.9%) | hep |
| 311 | phase field crystal | (2057808716) modeling elasticity in crystal growth (2002) (53.9%), (1979546093) modeling elastic and plastic deformations in nonequilibrium processing using phase field crystals (2004) (46.1%) | cond |
| 312 | decihertz interferomet gravit wave observatori | (2121634014) possibility of direct measurement of the acceleration of the universe using 0 1 hz band laser interferometer gravitational wave antenna in space (2001) (100%) | astro |
| 315 | irvin michigan brookhaven | (2068718327) observation of a neutrino burst from the supernova sn1987a (1987) (34.9%), (2156532391) observation of a neutrino burst in coincidence with supernova 1987a in the large magellanic cloud (1987) (53.4%), (2049888457) measurement of atmospheric neutrino composition with the imb 3 detector (1991) (11.7%) | hep |
| 319 | random graph, configur model | (2798319823)  (57%), (2169015768) random graphs with arbitrary degree distributions and their applications (2001) (43%) | cond |
| 321 | determin quantum mont carlo | (2172568165) monte carlo calculations of coupled boson fermion systems i (1981) (26.6%), (2017474131) computational complexity and fundamental limitations to fermionic quantum monte carlo simulations (2005) (26.7%), (2092114168) two dimensional hubbard model numerical simulation study (1985) (12.4%), (2029960903) numerical study of the two dimensional hubbard model (1989) (14.1%), (2061579329) sign problem in the numerical simulation of many electron systems (1990) (20.3%) | cond |
| 322 | perdew burk ernzerhof function revi solid | (2016168218) restoring the density gradient expansion for exchange in solids and surfaces (2008) (55.3%), (2168442976) the snowmass points and slopes benchmarks for susy searches (2002) (13.6%), (2066713443) collective cooling and self organization of atoms in a cavity (2002) (18.2%), (2085890526) observation of collective friction forces due to spatial self organization of atoms from rayleigh to bragg scattering (2003) (12.9%) | cond |
| 324 | unruh effect | (2118148695) notes on black hole evaporation (1976) (100%) | hep |
| 330 | detrend fluctuat analysi | (2017821362) mosaic organization of dna nucleotides (1994) (100%) | physics |
| 331 | compact muon solenoid detector | (2151512268) the cms experiment at the cern lhc (2008) (100%) | hep |
| 333 | lumin red galaxi | (2153811626) cosmological parameters from sdss and wmap (2004) (100%) | astro |
| 337 | random benchmark | (2170620327) randomized benchmarking of quantum gates (2008) (41.8%), (1848366941) exact and approximate unitary 2 designs and their application to fidelity estimation (2009) (13.9%), (1532834996) scalable and robust randomized benchmarking of quantum processes (2011) (33.3%), (2090368878) direct fidelity estimation from few pauli measurements (2011) (11.1%) | quant |

| | | | |
|---|---|---|---|
| 341 | gener probabilist theori | (2040870895) information processing in generalized probabilistic theories (2007) (33.9%), (2148098913) quantum theory from five reasonable axioms (2001) (23.8%), (2113235349) probabilistic theories with purification (2010) (21.4%), (2585503045) derivation of quantum theory (2017) (9.97%), (2161316267) a derivation of quantum theory from physical requirements (2010) (10.9%) | quant |
| 352 | swendsen wang | (1969758109) collective monte carlo updating for spin systems (1989) (41.9%), (2037139490) nonuniversal critical dynamics in monte carlo simulations (1987) (45.5%), (1987792759) conserving approximations for strongly correlated electron systems bethe salpeter equation and dynamics for the two dimensional hubbard model (1989) (12.7%) | cond |
| 356 | effect bodi | (2148994792) effective one body approach to general relativistic two body dynamics (1999) (43.6%), (2116108818) transition from inspiral to plunge in binary black hole coalescences (2000) (33.2%), (2155802808) coalescence of two spinning black holes an effective one body approach (2001) (23.2%) | gr |
| 357 | kawai lewellen tye relat | (2160023088) new relations for gauge theory amplitudes (2008) (24.4%), (2005261731) perturbative quantum gravity as a double copy of gauge theory (2010) (30%), (1988998898) a relation between tree amplitudes of closed and open strings (1986) (31.9%), (1964396907) gravity as the square of gauge theory (2010) (13.7%) | hep |
| 360 | tokai kamioka | (2115891487) indication of electron neutrino appearance from an accelerator produced off axis muon neutrino beam (2011) (41.8%), (2116793086) the jhf kamioka neutrino project (2001) (29.9%), (2027957671) observation of electron neutrino appearance in a muon neutrino beam (2014) (14.8%), (1490074796) nova proposal to build a 30 kiloton off axis detector to study neutrino oscillations in the fermilab numi beamline (2005) (13.5%) | hep |
| 366 | hopf algebra | (2126692863) renormalization in quantum field theory and the riemann hilbert problem i the hopf algebra structure of graphs and the main theorem (2000) (100%) | math |
| 367 | supersymmetr phenomenolog packag | (2132653891) generating feynman diagrams and amplitudes with feynarts 3 (2001) (26.1%), (2124410456) spheno a program for calculating supersymmetric spectra susy particle decays and susy particle production at e e colliders (2003) (37.9%), (1965311100) spheno 3 1 extensions including flavour cp phases and models beyond the mssm (2012) (36%) | hep |
| 375 | kolmogorov smirnov | (2313307644) (55.4%), (2171074980) a simplex method for function minimization (1965) (44.6%) | astro |
| 378 | kohn sham, kohn sham densiti function theori | (2230728100) self consistent equations including exchange and correlation effects (1965) (53%), (2030976617) inhomogeneous electron gas (1964) (47%) | cond |
| 379 | dual superconform symmetri | (1973074703) dual superconformal symmetry of scattering amplitudes in n 4 super yang mills theory (2010) (100%) | hep |

| | | | |
|---|---|---|---|
| 381 | stochast thermodynam | (2149801992) nonequilibrium equality for free energy differences (1997) (39.8%), (2021820673) entropy production fluctuation theorem and the nonequilibrium work relation for free energy differences (1999) (26.5%), (2086845400) entropy production along a stochastic trajectory and an integral fluctuation theorem (2005) (33.7%) | cond |
| 385 | floquet topolog insul | (1996350481) topological characterization of periodically driven quantum systems (2010) (56.7%), (2055924724) anomalous edge states and the bulk edge correspondence for periodically driven two dimensional systems (2012) (43.3%) | cond |
| 393 | quantum capac | (2001619599) sending classical information via noisy quantum channels (1997) (56.8%), (2141412822) the private classical capacity and quantum capacity of a quantum channel (2005) (43.2%) | cond |
| 405 | obliqu paramet | (1993681677) new constraint on a strongly interacting higgs sector (1990) (52.9%), (2111808546) estimation of oblique electroweak corrections (1992) (47.1%) | hep |
| 413 | breitenlohn freedman | (2151271622) breaking an abelian gauge symmetry near a black hole horizon (2008) (12.6%), (1971615006) ads cft correspondence and symmetry breaking (1999) (19.1%), (1989132853) stability in gauged extended supergravity (1982) (39.1%), (1972290465) positive energy in anti de sitter backgrounds and gauged extended supergravity (1982) (29.2%) | hep |
| 414 | polyakov nambu jona lasinio model, polyakov nambu jona lasinio | (2160215843) phases of qcd lattice thermodynamics and a field theoretical model (2006) (42.6%), (2592326221) polyakov loop diquarks and the two flavor phase diagram (2007) (15.3%), (2162510881) phase diagrams in the three flavor nambu jona lasinio model with the polyakov loop (2008) (17.9%), (2112319901) phase structure of the polyakov quark meson model (2007) (11.9%), (1982433853) chiral effective model with the polyakov loop (2004) (12.3%) | hep |
| 416 | self organ critic | (1966151890) punctuated equilibrium and criticality in a simple model of evolution (1993) (17.5%), (2008879046) self organized criticality (1988) (36.7%), (240942792) (22.2%), (2037190943) self organized critical state of sandpile automaton models (1990) (23.6%) | cond |
| 420 | sideband cool | (2077473036) quantum theory of cavity assisted sideband cooling of mechanical motion (2007) (32.2%), (2051421285) theory of ground state cooling of a mechanical oscillator using dynamical backaction (2007) (26.2%), (2074383155) laser cooling of a nanomechanical oscillator into its quantum ground state (2011) (23.2%), (1982991836) sideband cooling of micromechanical motion to the quantum ground state (2011) (18.4%) | quant |
| 421 | cryogen dark matter search | (2006441137) first results from the xenon10 dark matter experiment at the gran sasso national laboratory (2008) (36.1%), (2031831197) results from a low energy analysis of the cdms ii germanium data (2011) (63.9%) | astro |

| | | | |
|---|---|---|---|
| 427 | larg hadron colliderb detector | (2109896109) the lhcb detector at the lhc (2008) (70.3%), (2112358548) current matrix elements from a relativistic quark model (1971) (29.7%) | hep |
| 430 | ghost conden | (2004840376) ghost condensation and a consistent infrared modification of gravity (2004) (100%) | hep |
| 434 | electron ion collid | (1902565119) gluons and the quark sea at high energies distributions polarization tomography (2011) (38.4%), (1796954972) electron ion collider the next qcd frontier understanding the glue that binds us all (2016) (51.7%), (2166235937) a large hadron electron collider at cern (2012) (9.83%) | hep |
| 435 | quantum anom hall, haldan model | (2101893110) model for a quantum hall effect without landau levels condensed matter realization of the parity anomaly (1988) (100%) | cond |
| 439 | anti de sitter conform field theori, gaug graviti dualiti, anti de sitter conform field theori dictionari, gaug graviti correspond, maldacena conjectur | (2166248051) the large n limit of superconformal field theories and supergravity (1999) (69.1%), (2039609754) gauge theory correlators from non critical string theory (1998) (30.9%) | hep |
| 440 | fraction quantum hall, laughlin wave function | (1963826472) anomalous quantum hall effect an incompressible quantum fluid with fractionally charged excitations (1983) (56.5%), (2006294391) two dimensional magnetotransport in the extreme quantum limit (1982) (43.5%) | cond |
| 453 | real network, degr correl | (2040956707) assortative mixing in networks (2002) (100%) | physics |
| 464 | local techniqu, supersymmetr local | (2062544124) localization of gauge theory on a four sphere and supersymmetric wilson loops (2012) (48.5%), (2024239048) exact results for wilson loops in superconformal chern simons theories with matter (2010) (51.5%) | hep |
| 478 | geant4 toolkit | (2128158076) geant4 a simulation toolkit (2003) (36%), (2128881154) geant4 developments and applications (2006) (39.5%), (2088532709) the lhcb simulation application gauss design evolution and experience (2011) (24.5%) | hep |
| 479 | bose glass | (1967336145) boson localization and the superfluid insulator transition (1989) (100%) | cond |
| 481 | thermal inflat | (1966834271) thermal inflation and the moduli problem (1996) (64.3%), (1534863451) cosmology with a tev mass higgs field breaking the grand unified theory gauge symmetry (1995) (35.7%) | hep |
| 482 | tachikawa correspond | (2107702031) liouville correlation functions from four dimensional gauge theories (2010) (66.3%), (2081349667) security of quantum key distribution (2008) (10.5%), (2164253138) the operational meaning of min and max entropy (2009) (11.1%), (2044028691) structural relaxation made simple (2006) (12.1%) | hep |

| | | | |
|---|---|---|---|
| 484 | singl field inflat, quasi singl field inflat | (2035433892) non gaussian features of primordial fluctuations in single field inflationary models (2003) (100%) | hep |
| 488 | planck result | (1909359706) planck 2015 results xiii cosmological parameters (2016) (39.6%), (2139937287) planck 2013 results xvi cosmological parameters (2014) (36.1%), (2147875900) seven year wilkinson microwave anisotropy probe wmap observations cosmological interpretation (2011) (13%), (2111420418) planck 2013 results xxii constraints on inflation (2014) (11.3%) | astro |
| 495 | low frequenc arrai | (2147626801) lofar the low frequency array (2013) (100%) | astro |
| 508 | gener parton distribut, deepli virtual compton scatter, deepli virtual meson product, doubl distribut | (2066711625) exclusive processes in perturbative quantum chromodynamics (1980) (17%), (1993982013) gauge invariant decomposition of nucleon spin (1997) (27.4%), (2096088858) factorization for hard exclusive electroproduction of mesons in qcd (1997) (14%), (2128251357) nonforward parton distributions (1997) (24.2%), (2158050233) deeply virtual compton scattering (1997) (17.4%) | hep |
| 509 | britto cachazo feng witten | (2165400029) perturbative gauge theory as a string theory in twistor space (2004) (25.8%), (2046938033) new recursion relations for tree amplitudes of gluons (2005) (51.6%), (1811565114) direct proof of the tree level scattering amplitude recursion relation in yang mills theory (2005) (22.7%) | hep |
| 511 | anti algorithm, fastjet packag | (2165037242) fastjet user manual (2012) (77.1%), (2020410627) pileup subtraction using jet areas (2008) (13.4%), (2107816296) the catchment area of jets (2008) (9.5%) | hep |
| 513 | relativist heavi ion collid | (2181986281) formation of dense partonic matter in relativistic nucleus nucleus collisions at rhic experimental evaluation by the phenix collaboration (2005) (31.6%), (2169165406) experimental and theoretical challenges in the search for the quark gluon plasma the star collaboration s critical assessment of the evidence from rhic collisions (2005) (30.4%), (1986221934) the star collaboration (2009) (38%) | hep |
| 515 | cambridg aachen | (1999557321) successive combination jet algorithm for hadron collisions (1993) (21.9%), (1996262512) better jet clustering algorithms (1997) (35.7%), (2898699751) hadronization corrections to jet cross sections in deep inelastic scattering (1998) (33.2%), (2150097517) a practical seedless infrared safe cone jet algorithm (2007) (9.32%) | hep |
| 525 | scale free network, power law degr distribut, preferenti attach, barabasi albert model, network scienc, preferenti attach model | (2008620264) emergence of scaling in random networks (1999) (72.5%), (2112090702) collective dynamics of small world networks (1998) (27.5%) | cond |

| # | Topic | Papers | Field |
|---|---|---|---|
| 526 | brane tile | (2090467627) gauge theories from toric geometry and brane tilings (2006) (20%), (1990823599) brane dimers and quiver gauge theories (2005) (27.1%), (1601290775) dimer models and toric diagrams (2005) (23%), (2161141719) dimer models from mirror symmetry and quivering amoebae (2008) (14.3%), (2067947224) quivers tilings branes and rhombi (2007) (15.7%) | hep |
| 529 | d brane | (2084722249) dirichlet branes and ramond ramond charges (1995) (100%) | hep |
| 530 | quantum kei distribut, devic independ quantum kei distribut, superden code | (1978553093) teleporting an unknown quantum state via dual classical and einstein podolsky rosen channels (1993) (22.7%), (2051051926) quantum cryptography based on bell s theorem (1991) (55.1%), (2020725861) communication via one and two particle operators on einstein podolsky rosen states (1992) (22.3%) | quant |
| 533 | chiral kinet theori | (2019119165) hydrodynamics with triangle anomalies (2009) (18.8%), (2223420584) axial vector vertex in spinor electrodynamics (1969) (14.9%), (2046474690) berry curvature triangle anomalies and the chiral magnetic effect in fermi liquids (2012) (21.9%), (2060440937) chiral kinetic theory (2012) (26%), (2108823852) kinetic theory with berry curvature from quantum field theories (2013) (18.4%) | hep |
| 546 | holomorph anomali equat | (1996204343) kodaira spencer theory of gravity and exact results for quantum string amplitudes (1994) (49.5%), (2169889297) matrix models topological strings and supersymmetric gauge theories (2002) (23.4%), (2006442667) invariants of algebraic curves and topological expansion (2007) (27.2%) | hep |
| 548 | 5post newtonian order, effect on bodi | (2148994792) effective one body approach to general relativistic two body dynamics (1999) (49.5%), (2153135464) gravitational radiation from post newtonian sources and inspiralling compact binaries (2002) (50.5%) | gr |
| 554 | generali parton distribut | (1993982013) gauge invariant decomposition of nucleon spin (1997) (83.8%), (2128251357) nonforward parton distributions (1997) (16.2%) | hep |
| 556 | bern carrasco johansson | (2160023088) new relations for gauge theory amplitudes (2008) (78.7%), (2007148718) new color decompositions for gauge amplitudes at tree and loop level (2000) (21.3%) | hep |
| 557 | hamiltonian mean field | (2048147273) clustering and relaxation in hamiltonian long range dynamics (1995) (59.1%), (2052646173) inequivalence of ensembles in a system with long range interactions (2001) (19.4%), (1985860461) non gaussian equilibrium in a long range hamiltonian system (2001) (21.5%) | cond |
| 566 | laser assist tunnel | (2063327258) realization of the hofstadter hamiltonian with ultracold atoms in optical lattices (2013) (38.2%), (2143622822) realizing the harper hamiltonian with laser assisted tunneling in optical lattices (2013) (37.2%), (2093214152) tunable gauge potential for neutral and spinless particles in driven optical lattices (2012) (24.6%) | cond |

| # | Topic | Papers | Field |
|---|---|---|---|
| 567 | cresst ii, dama libra | (2159751525) results from a search for light mass dark matter with a p type point contact germanium detector (2011) (31.1%), (2133224752) results from 730 kg days of the cresst ii dark matter search (2012) (38.2%), (2098422406) silicon detector dark matter results from the final exposure of cdms ii (2013) (30.7%) | hep |
| 568 | horava lifshitz | (2158117921) quantum gravity at a lifshitz point (2009) (81%), (2101593131) solutions to horava gravity (2009) (19%) | hep |
| 586 | atla detector | (2098406858) the atlas experiment at the cern large hadron collider (2020) (49.5%), (2144690984) determination of jet energy calibration and transverse momentum resolution in cms (2011) (23.7%), (2115613123) performance of cms muon reconstruction in pp collision events at sqrt s 7 tev (2012) (26.8%) | hep |
| 589 | replica symmetri break, sherrington kirkpatrick model, sherrington kirkpatrick | (2158843074) spin glass theory (1987) (43%), (2022820481) solvable model of a spin glass (1975) (57%) | cond |
| 591 | stochast seri expan | (1974033108) quantum monte carlo with directed loops (2002) (26.5%), (2066428616) path integral computation of superfluid densities (1987) (18.7%), (2004352277) finite size scaling of the ground state parameters of the two dimensional heisenberg model (1997) (17.4%), (1990455780) quantum monte carlo simulation method for spin systems (1991) (37.4%) | cond |
| 597 | spontan parametr conver, spontan parametr down conver, parametr down conver, hong ou mandel | (2062584998) measurement of subpicosecond time intervals between two photons by interference (1987) (70.3%), (2045140905) new high intensity source of polarization entangled photon pairs (1995) (29.7%) | quant |
| 602 | brown rho scale | (1971702030) scaling effective lagrangians in a dense medium (1991) (100%) | hep |
| 604 | newman penro formal | (2153344483) an approach to gravitational radiation by a method of spin coefficients (1962) (61%), (2006682002) classification of the weyl tensor in higher dimensions (2004) (39%) | gr |
| 610 | advanc laser interferomet gravit wave observatori, gravit wave150914, gravit wave151226 | (2252795400) observation of gravitational waves from a binary black hole merger (2016) (49.4%), (2437571239) gw151226 observation of gravitational waves from a 22 solar mass binary black hole coalescence (2016) (22.4%), (2621266390) gw170104 observation of a 50 solar mass binary black hole coalescence at redshift 0 2 (2017) (12.4%), (2423333910) binary black hole mergers in the first advanced ligo observing run (2016) (15.8%) | astro |
| 614 | data process inequ, asymptot equipartit properti | (1995875735) a mathematical theory of communication (1948) (100%) | quant |
| 615 | superconduct qubit, charg qubit, flux qubit, transmon qubit | (2060904747) cavity quantum electrodynamics for superconducting electrical circuits an architecture for quantum computation (2004) (52.8%), (1963734567) charge insensitive qubit design derived from the cooper pair box (2007) (47.2%) | cond |

| | | | |
|---|---|---|---|
| 617 | tolman oppenheim volkoff | (183999792) (100%) | astro |
| 624 | algebra beth ansatz, six vertex model | (2047240241) absence of mott transition in an exact solution of the short range one band model in one dimension (1968) (100%) | math |
| 629 | evolutionari game theori | (1997418576) scale free networks provide a unifying framework for the emergence of cooperation (2005) (41.1%), (2025490132) evolutionary games and spatial chaos (1992) (28.3%), (2167062553) five rules for the evolution of cooperation (2006) (30.5%) | physics |
| 658 | cheval polarski linder | (2057645607) exploring the expansion history of the universe (2003) (65.7%), (2117797710) accelerating universes with scaling dark matter (2001) (34.3%) | astro |
| 669 | minim sensit | (2158631147) optimized perturbation theory (1981) (86.6%), (2138618377) a joint analysis of planck and bicep2 b modes including dust polarization uncertainty (2014) (13.4%) | hep |
| 672 | louvain method | (2095293504) finding and evaluating community structure in networks (2004) (15.2%), (2127048411) community detection in graphs (2009) (22.7%), (1971421925) community structure in social and biological networks (2002) (10.3%), (2131681506) fast unfolding of communities in large networks (2008) (22.4%), (2047940964) finding community structure in very large networks (2004) (8.83%), (2151936673) modularity and community structure in networks (2006) (10.6%), (2164928285) uncovering the overlapping community structure of complex networks in nature and society (2005) (9.9%) | physics |
| 676 | ghirardi rimini weber | (2082739523) unified dynamics for microscopic and macroscopic systems (1986) (64.3%), (2058509714) markov processes in hilbert space and continuous spontaneous localization of systems of identical particles (1990) (21.1%), (2065788435) models for universal reduction of macroscopic quantum fluctuations (1989) (14.6%) | quant |
| 691 | diffu limit aggreg model | (1985997303) diffusion limited aggregation a kinetic critical phenomenon (1981) (76.9%), (2159146375) internal structure of black holes (1990) (11.3%), (2034389272) inner structure of a charged black hole an exact mass inflation solution (1991) (11.8%) | cond |
| 694 | asymmetr simpl exclu process | (2076340951) boundary induced phase transitions in driven diffusive systems (1991) (17.7%), (2060101754) exact solution of a 1d asymmetric exclusion model using a matrix formulation (1993) (38.9%), (1977046089) nonequilibrium steady states of matrix product form a solver s guide (2007) (23.5%), (2133135948) nonequilibrium statistical mechanics of the zero range process and related models (2005) (20%) | cond |
| 697 | nose hoover thermostat | (106507613) (49.8%), (1991794210) canonical dynamics equilibrium phase space distributions (1985) (50.2%) | cond |
| 699 | electromagnet induc transpar | (1998325718) dark state polaritons in electromagnetically induced transparency (2000) (41.5%), (2056174226) storage of light in atomic vapor (2001) (25.6%), (2150262561) electromagnetically induced transparency (1997) (32.9%) | quant |

| | | | |
|---|---|---|---|
| 700 | transver momentum depend | (2103895497) universality of soft and collinear factors in hard scattering factorization (2004) (27.9%), (2149238093) qcd factorization for semi inclusive deep inelastic scattering at low transverse momentum (2005) (18.4%), (2093602712) leading twist single transverse spin asymmetries drell yan and deep inelastic scattering (2002) (32.2%), (1964553457) final state interactions and gauge invariant parton distributions (2003) (21.5%) | hep |
| 706 | particl flow | (2020410627) pileup subtraction using jet areas (2008) (20.6%), (2624881227) particle flow reconstruction and global event description with the cms detector (2017) (79.4%) | hep |
| 712 | grover search algorithm | (2002372750) quantum mechanics helps in searching for a needle in a haystack (1997) (43.7%), (2084652510) a fast quantum mechanical algorithm for database search (1996) (39.4%), (2020918964) tight bounds on quantum searching (1996) (16.9%) | quant |
| 722 | pyrochlor irid | (1538825449) topological semimetal and fermi arc surface states in the electronic structure of pyrochlore iridates (2011) (72.7%), (2046004949) weyl semimetal in a topological insulator multilayer (2011) (27.3%) | cond |
| 729 | low densiti pariti check | (2128765501) low density parity check codes (1963) (100%) | quant |
| 756 | f theori, f theori compactif, higg bundl | (2028727458) evidence for f theory (1996) (31.8%), (2121857565) guts and exceptional branes in f theory i (2009) (36.9%), (1965395651) model building with f theory (2011) (31.3%) | hep |
| 761 | axion like particl | (14341150) (100%) | hep |
| 762 | semiconductor nanowir | (2055424520) majorana fermions and a topological phase transition in semiconductor superconductor heterostructures (2010) (39.5%), (2067650803) helical liquids and majorana bound states in quantum wires (2010) (37.4%), (1965741196) generic new platform for topological quantum computation using semiconductor heterostructures (2010) (23.1%) | cond |
| 763 | network parton distribut functions2 | (2138291582) parton distributions with lhc data (2013) (81.1%), (1984125696) encoding a qubit in an oscillator (2001) (18.9%) | |
| 767 | traci widom distribut | (2499952289) level spacing distributions and the airy kernel (1994) (53.5%), (1993405207) shape fluctuations and random matrices (2000) (16.2%), (2068830353) limits on the majorana neutrino mass in the 0 1 ev range (1999) (13.2%), (2002912011) on orthogonal and symplectic matrix ensembles (1996) (17.1%) | math |
| 769 | linear augment planewav | (2068134289) ground state properties of lanthanum treatment of extended core states (1991) (100%) | cond |
| 771 | posit weight lead order match scheme | (2156247879) matching nlo qcd computations with parton shower simulations the powheg method (2007) (48.7%), (1991302676) a new method for combining nlo qcd with shower monte carlo algorithms (2004) (38.8%), (1768021241) observation of single top quark production (2009) (12.5%) | hep |

| | | | |
|---|---|---|---|
| 785 | neural networkparton distribut functions3 | (2151165908) parton distributions for the lhc (2009) (26%), (2082138120) new generation of parton distributions with uncertainties from global qcd analysis (2002) (27.4%), (2136044884) new parton distributions for collider physics (2010) (18.7%), (2133761278) implications of cteq global analysis for collider observables (2008) (14.2%), (2148539882) parton distributions for the lhc run ii (2015) (13.6%) | hep |
| 795 | univ feynrul output | (2029454374) ufo the universal feynrules output (2012) (67.2%), (2130277268) non abelian discrete symmetries in particle physics (2010) (32.8%) | hep |
| 797 | metropoli hast algorithm, metropoli hast | (2153208487) cosmological parameters from cmb and other data a monte carlo approach (2002) (21.5%), (2056760934) equation of state calculations by fast computing machines (1953) (30%), (1525315946) monte carlo methods (2013) (17.6%), (2138309709) monte carlo sampling methods using markov chains and their applications (1970) (30.9%) | astro |
| 799 | chameleon mechan | (2152565162) chameleon fields awaiting surprises for tests of gravity in space (2004) (78.4%), (2171317216) models of f r cosmic acceleration that evade solar system tests (2007) (21.6%) | astro |
| 822 | hard wall model | (2159787624) qcd and a holographic model of hadrons (2005) (39.9%), (2097479977) linear confinement and ads qcd (2006) (23.1%), (2000417761) hard scattering and gauge string duality (2002) (20.4%), (2058757509) chiral symmetry breaking from five dimensional spaces (2005) (16.6%) | hep |
| 842 | monkhorst pack | (2036113194) special points for brillouin zone integrations (1976) (100%) | cond |
| 850 | quantum hall ferromagnet | (2013841375) skyrmions and the crossover from the integer to fractional quantum hall effect at small zeeman energies (1993) (46.6%), (2060724751) quantum hall ferromagnetism in graphene (2006) (18.2%), (1975307432) (35.1%) | cond |
| 857 | doubl field theori | (2095752533) double field theory (2009) (33.1%), (2163909615) superspace duality in low energy superstrings (1993) (23%), (2098956145) generalized metric formulation of double field theory (2010) (24.6%), (2121037413) background independent action for double field theory (2010) (19.3%) | hep |
| 858 | ademollo gatto theorem | (2076541770) nonrenormalization theorem for the strangeness violating vector currents (1964) (100%) | hep |

| 861 | wilkinson microwav anisotropi probe, wilkinson microwav anisotropi probe data | (2139937287) planck 2013 results xvi cosmological parameters (2014) (10.2%), (2147875900) seven year wilkinson microwave anisotropy probe wmap observations cosmological interpretation (2011) (37.3%), (2115185242) nine year wilkinson microwave anisotropy probe wmap observations cosmological parameter results (2013) (16.6%), (2118265220) first year wilkinson microwave anisotropy probe wmap observations determination of cosmological parameters (2003) (15.7%), (2001369144) nine year wilkinson microwave anisotropy probe wmap observations final maps and results (2013) (9.56%), (2075884494) first year wilkinson microwave anisotropy probe wmap observations preliminary maps and basic results (2003) (10.7%) | astro |
| 865 | string ga cosmolog, pre big bang | (2029143135) ekpyrotic universe colliding branes and the origin of the hot big bang (2001) (51.9%), (1979287894) cosmic evolution in a cyclic universe (2002) (23.9%), (2068380679) pre big bang in string cosmology (1993) (24.2%) | hep |
| 881 | parton shower, powheg box | (2258584306) pythia 6 4 physics and manual (2006) (48.8%), (2157014333) matching nlo qcd computations and parton shower simulations (2002) (51.2%) | hep |
| 885 | dynam mean field theori, dynam cluster approxim | (1983596827) hubbard model in infinite dimensions (1992) (49.8%), (2009742424) cellular dynamical mean field approach to strongly correlated systems (2001) (50.2%) | cond |
| 886 | larg underground xenon experi | (2042260934) first results from the lux dark matter experiment at the sanford underground research facility (2014) (65%), (2534308951) results from a search for dark matter in the complete lux exposure (2017) (20.9%), (2340376607) improved limits on scattering of weakly interacting massive particles from reanalysis of 2013 lux data (2016) (14.1%) | hep |
| 892 | dual conform symmetri | (2030665456) gluon scattering amplitudes at strong coupling (2007) (27.7%), (2068283118) conformal properties of four gluon planar amplitudes and wilson loops (2008) (18.7%), (1973074703) dual superconformal symmetry of scattering amplitudes in n 4 super yang mills theory (2010) (21.1%), (2076329426) conformal ward identities for wilson loops and a test of the duality with gluon amplitudes (2010) (16%), (1970198451) magic identities for conformal four point integrals (2007) (16.6%) | hep |
| 895 | brout englert higg mechan | (1996413343) broken symmetries and the masses of gauge bosons (1964) (50.4%), (2045120204) broken symmetry and the mass of gauge vector mesons (1964) (49.6%) | hep |
| 907 | non determinist polynomi problem complet | (2891212941) computers and intractability a guide to the theory of np completeness (1999) (100%) | quant |
| 920 | batalin vilkoviski formal, batalin vilkoviski | (2078426698) gauge algebra and quantization (1981) (54.4%), (2156682843) quantization of gauge theories with linearly dependent generators (1983) (45.6%) | hep |
| 921 | margin fermi liquid | (1988371505) phenomenology of the normal state of cu o high temperature superconductors (1989) (100%) | cond |

| | | | |
|---|---|---|---|
| 923 | holograph entangl entropi | (2157692768) a covariant holographic entanglement entropy proposal (2007) (100%) | hep |
| 934 | potenti local orbit | (1974378834) full potential nonorthogonal local orbital minimum basis band structure scheme (1999) (100%) | cond |
| 936 | slow roll inflat, e fold, inflationari cosmolog, slow roll, flat problem, cosmic inflat | (2134251287) inflationary universe a possible solution to the horizon and flatness problems (1981) (100%) | hep |
| 952 | quantum einstein graviti, asymptot safeti scenario | (2592434037) nonperturbative evolution equation for quantum gravity (1998) (35%), (2000661521) fixed points of quantum gravity (2004) (12.5%), (2030583794) investigating the ultraviolet properties of gravity with a wilsonian renormalization group equation (2009) (15.7%), (2148908350) ultraviolet fixed point and generalized flow equation of quantum gravity (2001) (16.8%), (2099174206) renormalization group flow of quantum gravity in the einstein hilbert truncation (2002) (20%) | hep |
| 957 | glashow iliopoulo maiani | (2117836026) weak interactions with lepton hadron symmetry (1970) (100%) | hep |
| 962 | singular theorem | (2052045615) gravitational collapse and space time singularities (1965) (100%) | gr |
| 964 | beck lee yang parr | (2023271753) development of the colle salvetti correlation energy formula into a functional of the electron density (1988) (63.6%), (2086957099) density functional exchange energy approximation with correct asymptotic behavior (1988) (36.4%) | cond |
| 970 | camassa holm equat | (2026031976) an integrable shallow water equation with peaked solitons (1993) (100%) | math |
| 971 | code anisotropi microwav background | (1944901286) efficient computation of cmb anisotropies in closed frw models (2000) (100%) | astro |
| 975 | hoeffd inequ | (2103012681) probability inequalities for sums of bounded random variables (1963) (100%) | quant |
| 979 | usadel equat | (2058903656) generalized diffusion equation for superconducting alloys (1970) (82.2%), (1983230862) josephson junction through a thin ferromagnetic layer negative coupling (2002) (17.8%) | cond |
| 981 | mclerran venugopalan | (2157155801) computing quark and gluon distribution functions for very large nuclei (1994) (65.3%), (1999089379) gluon distribution functions for very large nuclei at small transverse momentum (1994) (34.7%) | hep |
| 994 | collid detector fermilab | (2596708472)  (29.1%), (2118143002) nonperturbative strong field effects in tensor scalar theories of gravitation (1993) (10.1%), (2157878177) evidence for a mass dependent forward backward asymmetry in top quark pair production (2011) (22.5%), (2162860066) evidence for an anomalous like sign dimuon charge asymmetry (2010) (11.8%), (2119952623) forward backward asymmetry in top quark antiquark production (2011) (17%), (2605950437)  (9.49%) | hep |

| | | | |
|---|---|---|---|
| 1000 | polyakov loop extend nambu jona lasinio | (2160215843) phases of qcd lattice thermodynamics and a field theoretical model (2006) (16.4%), (2592326221) polyakov loop diquarks and the two flavor phase diagram (2007) (22.8%), (2112319901) phase structure of the polyakov quark meson model (2007) (26%), (1973150779) susceptibilities and the phase structure of a chiral model with polyakov loops (2007) (13.5%), (2011685017) polyakov loop in chiral quark models at finite temperature (2006) (21.3%) | hep |
| 1006 | kachru kallosh lind trivedi | (2125835296) de sitter vacua in string theory (2003) (50.6%), (2173328249) hierarchies from fluxes in string compactifications (2002) (49.4%) | hep |
| 1014 | strong disord renorm group | (2065552177) random antiferromagnetic quantum spin chains (1994) (27.2%), (1989792517) random transverse field ising spin chains (1992) (25.5%), (2083024011) critical behavior of random transverse field ising spin chains (1995) (18%), (2004482355) random antiferromagnetic chain (1979) (10.3%), (2016751437) low temperature properties of the random heisenberg antiferromagnetic chain (1980) (10.5%), (1989444358) infinite randomness quantum ising critical fixed points (2000) (8.54%) | cond |
| 1018 | dimen deconstruct | (2027981434) de constructing dimensions (2001) (74.5%), (2140164202) gauge invariant effective lagrangian for kaluza klein modes (2001) (25.5%) | hep |
| 1028 | pontryagin maximum principl | (2147733222) time optimal control in spin systems (2001) (12%), (2079336704) teaching lasers to control molecules (1992) (12.2%), (2083494341) optimal control of quantum mechanical systems existence numerical approximation and applications (1988) (17.4%), (320263972) theory of optimal processes (1961) (58.4%) | math |
| 1033 | bragg spectroscopi | (2138628761) bragg spectroscopy of a bose einstein condensate (1999) (100%) | cond |
| 1035 | laporta algorithm | (2339590713) a 2 4 determination of the local value of the hubble constant (2016) (14.2%), (2045719640) constraints on the redshift dependence of the dark energy potential (2005) (34.4%), (2020646255) algorithm fire feynman integral reduction (2008) (19.8%), (2099740576) fire5 a c implementation of feynman integral reduction (2015) (14.4%), (2140092737) high precision calculation of multiloop feynman integrals by difference equations (2000) (17.2%) | hep |
| 1038 | self propel particl, vicsek model | (2015410655) novel type of phase transition in a system of self driven particles (1995) (100%) | cond |
| 1040 | diffu mont carlo | (2026907619) ground state of the electron gas by a stochastic method (1980) (100%) | cond |
| 1043 | dodelson widrow | (2138874905) sterile neutrinos as dark matter (1994) (60.9%), (2007184284) new dark matter candidate nonthermal sterile neutrinos (1999) (19.6%), (2020336568) sterile neutrino hot warm and cold dark matter (2001) (19.5%) | hep |

| 1053 | jam transit | (2035178137) jamming at zero temperature and zero applied stress the epitome of disorder (2003) (68.6%), (2031715663) random packings of frictionless particles (2002) (31.4%) | cond |
| --- | --- | --- | --- |
| 1056 | master stabil function | (1965928854) complete characterization of a quantum process the two bit quantum gate (1997) (30.3%), (2078439793) prescription for experimental determination of the dynamics of a quantum black box (1997) (19.4%), (1984964553) ancilla assisted quantum process tomography (2003) (9.47%), (2083787309) master stability functions for synchronized coupled systems (1998) (40.8%) | cond |
| 1062 | weyl semimet, weyl node | (1538825449) topological semimetal and fermi arc surface states in the electronic structure of pyrochlore iridates (2011) (64.5%), (2046004949) weyl semimetal in a topological insulator multilayer (2011) (35.5%) | cond |
| 1095 | inver scatter transform | (2026031976) an integrable shallow water equation with peaked solitons (1993) (100%) | math |
| 1098 | klebanov strassler | (2145530412) supergravity and a confining gauge theory duality cascades and ??sb resolution of naked singularities (2000) (60.9%), (2023378071) superconformal field theory on threebranes at a calabi yau singularity (1998) (16%), (2053126470) comments on conifolds (1990) (13.3%), (2084282513) the baryonic branch of klebanov strassler solution a supersymmetric family of su 3 structure backgrounds (2005) (9.76%) | hep |
| 1138 | brodski lepag mackenzi | (2116659902) on the elimination of scale ambiguities in perturbative quantum chromodynamics (1983) (100%) | hep |
| 1178 | singl second gener potenti model | (2057910081) equation of state for dense nucleon matter (1988) (14.4%), (2050773679) accurate nucleon nucleon potential with charge independence breaking (1995) (85.6%) | |
| 1197 | camassa holm | (1990332384) surface barrier in type ii superconductors (1964) (10.3%), (2026031976) an integrable shallow water equation with peaked solitons (1993) (89.7%) | math |
| 1201 | gillespi algorithm | (2155418451) exact stochastic simulation of coupled chemical reactions (1977) (89.1%), (2037014352) adaptive competition market efficiency and phase transitions (1999) (10.9%) | cond |
| 1203 | cachazo yuan | (2076890194) scattering of massless particles scalars gluons and gravitons (2014) (34.9%), (1640497314) scattering equations and matrices from einstein to yang mills dbi and nlsm (2015) (33%), (2106416540) ambitwistor strings and the scattering equations (2014) (17.4%), (2009867235) einstein yang mills scattering amplitudes from scattering equations (2015) (14.7%) | hep |
| 1226 | lieb theorem | (2066009637) two theorems on the hubbard model (1989) (76%), (1996133706) ferromagnetism in the hubbard models with degenerate single electron ground states (1992) (9.96%), (1999692357) flux phase of the half filled band (1994) (14%) | cond |

| | | | |
|---|---|---|---|
| 1231 | symmetri protect topolog | (2003095492) symmetry protected topological orders and the group cohomology of their symmetry group (2013) (47.3%), (2168101857) tensor entanglement filtering renormalization approach and symmetry protected topological order (2009) (17.4%), (2105077826) classification of gapped symmetric phases in one dimensional spin systems (2011) (17.9%), (1816505947) theory and classification of interacting integer topological phases in two dimensions a chern simons approach (2012) (17.3%) | cond |
| 1233 | lut liquid | (2101936997) effective harmonic fluid approach to low energy properties of one dimensional quantum fluids (1981) (100%) | cond |
| 1238 | axion monodromi inflat, axion monodromi | (2132522347) monodromy in the cmb gravity waves and string inflation (2008) (53.3%), (2142786715) gravity waves and linear inflation from axion monodromy (2010) (46.7%) | hep |
| 1287 | nonequilibrium green function, kadanoff baym equat, keldysh formal | (2044369461) (100%) | cond |
| 1295 | k essenc, quintess model | (2049393017) cosmological imprint of an energy component with general equation of state (1998) (24%), (2137919480) cosmological consequences of a rolling homogeneous scalar field (1988) (27.2%), (2032901690) quintessence cosmic coincidence and the cosmological constant (1999) (23.5%), (2036482104) dynamical solution to the problem of a small cosmological constant and late time cosmic acceleration (2000) (25.3%) | astro |
| 1301 | self interact correct | (2052637367) self interaction correction to density functional approximations for many electron systems (1981) (46.4%), (2049079467) accurate and simple analytic representation of the electron gas correlation energy (1992) (24.9%), (2026907619) ground state of the electron gas by a stochastic method (1980) (28.7%) | cond |
| 1305 | caldeira leggett model | (2038293483) influence of dissipation on quantum tunneling in macroscopic systems (1981) (46.4%), (2052199078) quantum brownian motion in a general environment exact master equation with nonlocal dissipation and colored noise (1992) (53.6%) | cond |
| 1324 | griffith phase | (24079269) (100%) | cond |
| 1335 | quantum walk, continu time quantum walk | (1966092221) quantum random walks (1993) (36.2%), (1975528636) quantum computation and decision trees (1998) (24.9%), (2054665888) quantum walks on graphs (2001) (19.8%), (2101697221) one dimensional quantum walks (2001) (19.1%) | quant |
| 1340 | perfect len | (2122765624) negative refraction makes a perfect lens (2000) (56.3%), (1995796806) controlling electromagnetic fields (2006) (26.1%), (2048377078) optical conformal mapping (2006) (17.7%) | physics |

| | | | |
|---|---|---|---|
| 1375 | pierr auger observatori | (2479485000) the pierre auger collaboration (2013) (78.4%), (2594331859) extension of the cosmic ray energy spectrum beyond the predicted greisen zatsepin kuz min cutoff (1998) (21.6%) | astro |
| 1387 | ashtekar variabl | (2098989214) new variables for classical and quantum gravity (1986) (100%) | gr |
| 1399 | non determinist polynomi hard | (2891212941) computers and intractability a guide to the theory of np completeness (1999) (100%) | cond |
| 1411 | casimir forc | (1968717374) measurement of the casimir force between parallel metallic surfaces (2002) (47.6%), (1974106262) casimir forces between arbitrary compact objects (2007) (52.4%) | cond |
| 1449 | chiral magnet effect, chiral vortic effect | (2019119165) hydrodynamics with triangle anomalies (2009) (19.6%), (1958894153) chiral magnetic effect (2008) (38.7%), (2025712550) the effects of topological charge change in heavy ion collisions event by event p and cp violation (2008) (24.7%), (1995175048) equilibrium parity violating current in a magnetic field (1980) (17%) | hep |
| 1451 | schrieffer heeger | (2072703702) solitons in polyacetylene (1979) (81%), (2029044648) natural susy endures (2012) (19%) | cond |
| 1481 | holograph renorm | (2099204364) a stress tensor for anti de sitter gravity (1999) (35.3%), (2146203925) the holographic weyl anomaly (1998) (21.8%), (2035873690) holographic reconstruction of spacetime and renormalization in the ads cft correspondence (2001) (25.5%), (2153274301) lecture notes on holographic renormalization (2002) (17.3%) | hep |
| 1489 | parametr reson | (2007443706) reheating after inflation (1994) (37.4%), (2009173309) modulus stabilization with bulk fields (1999) (21.1%), (2158442860) towards the theory of reheating after inflation (1997) (26.7%), (2112197499) particle production during out of equilibrium phase transitions (1990) (14.9%) | hep |
| 1496 | surfac code, stabil code, stabil formal | (1839785972) stabilizer codes and quantum error correction (1997) (24.4%), (2135273380) fault tolerant quantum computation by anyons (2003) (17.7%), (2030867075) surface codes towards practical large scale quantum computation (2012) (21.4%), (2088636071) topological quantum memory (2001) (19.6%), (2013754283) topological quantum distillation (2006) (17%) | quant |
| 1501 | banado teitelboim zanelli black hole, banado teitelboim zanelli | (1763748250) black hole in three dimensional spacetime (1992) (67.4%), (2166574392) geometry of the 2 1 black hole (1993) (32.6%) | gr |

| | | | |
|---|---|---|---|
| 1503 | emb tensor formal | (2090873441) on lagrangians and gaugings of maximal supergravities (2003) (21.3%), (2091528229) magnetic charges in local field theory (2005) (22.7%), (1985154855) maximal gauged supergravity in three dimensions (2001) (14.8%), (2061201154) lectures on gauged supergravity and flux compactifications (2008) (10.1%), (2035903567) gauged supergravities tensor hierarchies and m theory (2008) (12.7%), (1981433114) the maximal d 4 supergravities (2007) (18.5%) | hep |
| 1536 | voter model | (2150041461) voter model on heterogeneous graphs (2005) (18.8%), (2083689991) mixing beliefs among interacting agents (2000) (32.5%), (2241340194) dynamics of majority rule in two state interacting spin systems (2003) (21.7%), (2113096089) opinion dynamics and bounded confidence models analysis and simulation (2002) (27.1%) | cond |
| 1561 | photon band gap | (2075689277) inhibited spontaneous emission in solid state physics and electronics (1987) (100%) | cond |
| 1593 | doubl special rel | (2087945706) lorentz invariance with an invariant energy scale (2002) (70.2%), (1526184262) (12.1%), (2130314416) egret observations of the extragalactic gamma ray emission (1998) (17.7%) | hep |
| 1607 | wang landau | (2003582562) efficient multiple range random walk algorithm to calculate the density of states (2001) (74.2%), (1970897635) determining the density of states for classical statistical models a random walk algorithm to produce a flat histogram (2001) (25.8%) | cond |
| 1608 | entangl wedg | (2098838878) bulk locality and quantum error correction in ads cft (2014) (11.8%), (1973553511) maximin surfaces and the strong subadditivity of the covariant holographic entanglement entropy (2014) (17%), (2008982490) holographic representation of local bulk operators (2006) (10.3%), (2067737180) the gravity dual of a density matrix (2012) (18.1%), (2115667196) causality holographic entanglement entropy (2014) (17.4%), (2074712751) causal holographic information (2012) (12.1%), (2470729034) reconstruction of bulk operators within the entanglement wedge in gauge gravity duality (2016) (13.4%) | hep |
| 1626 | move punctur approach | (1989170879) evolution of binary black hole spacetimes (2005) (22.3%), (1997367763) accurate evolutions of orbiting black hole binaries without excision (2006) (34.8%), (1971068715) gravitational wave extraction from an inspiraling configuration of merging black holes (2006) (29.3%), (2179614604) evolution of three dimensional gravitational waves harmonic slicing case (1995) (13.6%) | astro |
| 1644 | macdonald polynomi | (2611234537) (100%) | math |
| 1654 | kovtun son starinet bound | (2097909025) viscosity in strongly interacting quantum field theories from black hole physics (2005) (100%) | hep |

| | | | |
|---|---|---|---|
| 1666 | heavi flavor averag group | (2163389141) averages of b hadron c hadron and tau lepton properties as of early 2012 (2017) (33.9%), (1523380628) averages of b hadron c hadron and tau lepton properties (2010) (21.7%), (1615451475) averages of b hadron and c hadron properties at the end of 2007 (2008) (22.4%), (2806596999) (22%) | hep |
| 1684 | loop quantum cosmolog | (2149289292) quantum nature of the big bang improved dynamics (2006) (27.9%), (2060198182) quantum nature of the big bang (2006) (11.8%), (2109902906) loop quantum cosmology (2005) (11.8%), (2064559964) loop quantum cosmology a status report (2011) (13.6%), (2071554951) quantum nature of the big bang an analytical and numerical investigation (2006) (13.7%), (2101065856) mathematical structure of loop quantum cosmology (2003) (11.5%), (1975600559) robustness of key features of loop quantum cosmology (2008) (9.71%) | gr |
| 1694 | numer algorithm | (2313307644) (100%) | cond |
| 1703 | seiberg witten curv, instanton partit function | (2120340707) n 2 dualities (2012) (100%) | hep |
| 1715 | cosmomont carlo | (2153208487) cosmological parameters from cmb and other data a monte carlo approach (2002) (77.6%), (2092194717) multinest an efficient and robust bayesian inference tool for cosmology and particle physics (2009) (22.4%) | astro |
| 1726 | bubbl colli | (1998067752) gravitational radiation from first order phase transitions (1994) (28.7%), (2014058866) gravitational radiation from colliding vacuum bubbles (1992) (14.7%), (2277474101) science with the space based interferometer elisa ii gravitational waves from cosmological phase transitions (2016) (13.8%), (2060583654) gravitational waves from first order cosmological phase transitions (1992) (14.3%), (2053135586) gravitational wave production by collisions more bubbles (2008) (14.9%), (2117249822) gravitational radiation from colliding vacuum bubbles envelope approximation to many bubble collisions (1993) (13.5%) | hep |
| 1734 | coleman de luccia | (2159119089) gravitational effects on and of vacuum decay (1980) (43.6%), (2085793749) fate of the false vacuum semiclassical theory (1977) (36.7%), (1629285415) fate of the false vacuum ii first quantum corrections (1977) (19.8%) | hep |
| 1761 | univ set | (2102008187) elementary gates for quantum computation (1995) (100%) | quant |
| 1777 | moduli stabil | (2125835296) de sitter vacua in string theory (2003) (65.6%), (2173328249) hierarchies from fluxes in string compactifications (2002) (34.4%) | hep |
| 1793 | coleman weinberg mechan, coleman weinberg | (2160306805) radiative corrections as the origin of spontaneous symmetry breaking (1973) (100%) | hep |

| | | | |
|---|---|---|---|
| 1794 | bagger lambert gustavsson model, bagger lambert gustavsson | (2171335381) n=6 superconformal chern simons matter theories m2 branes and their gravity duals (2008) (11.2%), (2131934658) algebraic structures on parallel m2 branes (2009) (33.4%), (2166592280) gauge symmetry and supersymmetry of multiple m2 branes (2008) (38%), (2075672245) comments on multiple m2 branes (2008) (17.4%) | hep |
| 1799 | still weber | (2037782625) computer simulation of local order in condensed phases of silicon (1985) (46.1%), (2023616054) empirical potential for hydrocarbons for use in simulating the chemical vapor deposition of diamond films (1990) (24.5%), (2082144555) modeling solid state chemistry interatomic potentials for multicomponent systems (1989) (29.5%) | cond |
| 1844 | blonder tinkham klapwijk | (1989116198) transition from metallic to tunneling regimes in superconducting microconstrictions excess current charge imbalance and supercurrent conversion (1982) (100%) | cond |
| 1845 | elitzur theorem | (2104549471) impossibility of spontaneously breaking local symmetries (1975) (100%) | hep |
| 1850 | weak valu | (2004360248) ultrasensitive beam deflection measurement via interferometric weak value amplification (2009) (32.5%), (1984118521) properties of a quantum system during the time interval between two measurements (1990) (24.7%), (2037253643) realization of a measurement of a weak value (1991) (22.2%), (1968356682) how the result of a measurement of a component of the spin of a spin 1 2 particle can turn out to be 100 (1988) (20.6%) | quant |
| 1875 | binder cumul | (2032367804) scaling theory for finite size effects in the critical region (1972) (49.3%), (2062078129) critical properties from monte carlo coarse graining and renormalization (1981) (50.7%) | cond |
| 1885 | 3d topolog insul, three dimen topolog insul | (1514707913) topological field theory of time reversal invariant insulators (2008) (49.5%), (2031028572) topological insulators with inversion symmetry (2007) (50.5%) | cond |
| 1902 | kibbl zurek mechan | (2142768579) topology of cosmic domains and strings (1976) (15.1%), (2161655651) dynamics of a quantum phase transition (2005) (21.7%), (2090911654) dynamics of a quantum phase transition exact solution of the quantum ising model (2005) (15.3%), (1586349125) universal adiabatic dynamics in the vicinity of a quantum critical point (2005) (14.2%), (2065469285) cosmological experiments in superfluid helium (1985) (19.4%), (2027742874) the simplest quantum model supporting the kibble zurek mechanism of topological defect production landau zener transitions from a new perspective (2005) (14.3%) | cond |
| 1904 | projector augment | (1970127494) projector augmented wave method (1994) (100%) | cond |

| | | | |
|---|---|---|---|
| 1907 | horndeski theori | (2062912482) second order scalar tensor field equations in a four dimensional space (1974) (22.3%), (2115120324) from k essence to generalized galileons (2011) (23.9%), (2171957754) generalized g inflation inflation with the most general second order field equations (2011) (20.2%), (2790775510) healthy theories beyond horndeski (2016) (20.3%), (2152850103) exploring gravitational theories beyond horndeski (2014) (13.3%) | gr |
| 1920 | mikheev smirnov wolfenstein | (2115896628) neutrino oscillations in matter (1978) (86.1%), (2118199673) 2hdmc two higgs doublet model calculator (2010) (13.9%) | hep |
| 1998 | anderson orthogon catastroph | (2061206493) infrared catastrophe in fermi gases with local scattering potentials (1967) (100%) | cond |



\end{document}